\RequirePackage[l2tabu,orthodox]{nag}
\documentclass[copyright,a4paper,10pt]{eptcs}
\newcommand\publicationstatus{}

\pdfoutput=1

\usepackage[T1]{fontenc}
\usepackage{lmodern}
\usepackage{fixltx2e}
\usepackage{microtype}
\usepackage{xspace}
\usepackage{enumitem}

\makeatletter
\newcommand\etc{etc\@ifnextchar.{}{.\@}\xspace}

\makeatother

\usepackage{graphicx}

\newcommand{\inlinegraphic}[2]{
  \dimendef\grafheight=255\dimendef\grafvshift=254
  \grafheight=#1
  \grafvshift=-0.5\grafheight
  \advance\grafvshift by 0.5ex
  \raisebox{\grafvshift}{\includegraphics[height=\grafheight]{images/#2}\xspace}
}

\newcommand{\ninlinegraphic}[2][1.0]{
  \dimendef\grafheight=255\dimendef\grafvshift=254
  \setbox0 = \hbox{\scalebox{#1}{\includegraphics{images/#2}}}
  \grafheight=\the\ht0
  \grafvshift=-0.5\grafheight
  \advance\grafvshift by 0.5ex
  \raisebox{\grafvshift}{\includegraphics[height=\grafheight]{images/#2}\xspace}
}

\usepackage{latexsym}
\usepackage{amssymb}
\usepackage{amsmath} 
\usepackage{stmaryrd}
\usepackage{mathtools}
\usepackage{stackrel}

\usepackage{amsthm}

\theoremstyle{definition}
\theoremstyle{definition}
\theoremstyle{definition}
\theoremstyle{definition}
\theoremstyle{definition}
\theoremstyle{definition}

\ifx\numericids\undefined

\else

\fi

\usepackage{tikz-cd}

\usepackage{dsfont}

\renewcommand{\int}[1]{\ensuremath{\mathrm{int}(#1)}}

\usepackage{macros/tikzfig}
\tikzstyle{H}=[-, style=dashed]
\tikzstyle{left aligned text}=[align=left]

\usepackage{subcaption}
\usepackage{csvsimple}
\usepackage{multirow}
\usepackage{makecell}

\usepackage{hyperref}

\usepackage{ccg-latex}

\usepackage[outline]{contour}

\usepackage[normalem]{ulem}
  {\gdef\scalefactor{#1}\begin{center}\proofSkipAmount \leavevmode}%
  {\scalebox{\scalefactor}{\DisplayProof}\proofSkipAmount \end{center} }
\def\defaultHypSeparation{\hskip 0.7em}
\def\ScoreOverhang{1pt}

\usepackage{rwd-drafting} 

\usepackage{bussproofs}
\usepackage{algorithm}
\usepackage{algorithmic}
\usepackage{caption}
\usepackage{float} 

\input{front-matter.tex}  

\begin{document}
\maketitle

\begin{abstract}
    There is a significant disconnect between linguistic theory and modern NLP practice, which relies heavily on inscrutable black-box architectures.
    DisCoCirc is a newly proposed model for meaning that aims to bridge this divide, by providing neuro-symbolic models that incorporate linguistic structure.
    DisCoCirc represents natural language text as a `circuit' that captures the core semantic information of the text.
    These circuits can then be interpreted as modular machine learning models.
    Additionally, DisCoCirc fulfils another major aim of providing an NLP model that can be implemented on near-term quantum computers.
    
    In this paper we describe a software pipeline\footnote{Code available at \url{https://github.com/CQCL/text\_to\_discocirc}} that converts English text to its DisCoCirc representation.
    The pipeline achieves coverage over a large fragment of the English language.
    It relies on Combinatory Categorial Grammar (CCG) parses of the input text as well as coreference resolution information.
    This semantic and syntactic information is used in several steps to convert the text into a simply-typed $\lambda$-calculus term, and then into a circuit diagram.
    This pipeline will enable the application of the DisCoCirc framework to NLP tasks, using both classical and quantum approaches.
  \end{abstract}

\tableofcontents


\section{Introduction}

Modern deep learning methods, especially those involving large transformer models~\cite{vaswani2017attention, openai2023gpt,touvron2023llama,chowdhery2022palm,thoppilan2022lamda}, have achieved remarkable results in the field of natural language processing (NLP)~\cite{radford2019language, brown2020language,bubeck2023sparks,choi2023chatgpt}.
However, these systems have many drawbacks -- the performance of large language models (LLMs) can be highly uneven~\cite{bian2023chatgpt,dziri2023faith,DBLP:journals/corr/abs-2302-06476,berglund2023reversal,koralus2023humans,DBLP:journals/corr/abs-2306-09479}, and the inscrutable nature of these big end-to-end networks means it is difficult to predict when they will fail, or understand how they arrive at their solutions~\cite{10.1145/3442188.3445922,rudin2019stop}.
Additionally, they offer little linguistic or theoretical insight into language.

In light of these issues, there has been significant interest in \textit{neuro-symbolic} approaches that bring together the neural network and symbolic traditions of artificial intelligence~\cite{sarker2021neuro, garcez2015neural,garcez2022neural}.
The neuro-symbolic methods have been shown to improve interpretability, out-of-distribution generalization, and learning from small data~\cite{chen2020compositional,mao2019neuro,demeter2020just,liang2018symbolic}.

DisCoCirc~\cite{wang2023distilling, coecke2021mathematics} is a recently proposed model from the lineage of `distributional compositional' models for meaning, which is a family of neuro-symbolic approaches.
Earlier work in the `DisCo' direction, mainly based around the DisCoCat framework~\cite{CoeckeBob2010MFfa}, aimed to synthesize vectorial semantics with syntax at the level of individual constituents and sentences.
The purpose of the syntax in these models is to mediate the flow of meaning and information.
While frameworks like DisCoCat were successful in tasks such as word sense disambiguation~\cite{grefenstette2011experimental, kartsaklis2013disambiguation}, paraphrase detection~\cite{Grefenstette2015Empirical}, and sentence classification~\cite{Meichanetzidis2023grammar},
they were crucially restricted to modelling phrases and sentences.

DisCoCirc extends the distributional compositional philosophy beyond the sentence level and models text consisting of multiple sentences, i.e. discourse.
It does this by representing text as a \textit{circuit}, allowing for the meaningful composition of the circuits corresponding to different sentences.
As such, they are often referred to as \textit{text circuits}.
DisCoCirc has been applied to problems such as solving logical puzzles~\cite{duneau2020puzzles, duneau2021parsing} and modelling conversational negation~\cite{rodatz2021conversational, shaikh2022conversational}. 
Moreover, whilst DisCoCirc is generally used as a model for natural language, it can also be applied to other modalities --
\cite{wangmascianica2021talking} applies it to model physical spatial relations, and~\cite{CoeckeBob2010MFfa} suggests applying it to visual modes.

Another feature of DisCoCirc, which partly motivated its development, is that by representing text in a circuit format it opens the door to implementation on near-term quantum computers.
Previous models like DisCoCat had been explored in the context of `quantum natural language processing', due to a close mathematical analogy between their mathematical structure and the mathematical structure of quantum processes~\cite{CoeckeBob2010MFfa,zeng2016quantum}.
The circuits of DisCoCirc can be much more naturally interpreted as a specification for a quantum circuit than the diagrams of DisCoCat.

The DisCoCirc approach of modelling semantics as compositional circuits aligns closely with the dynamic semantics perspective~\cite{Lewis2017DynamicSemantics}.
The shared philosophy is that pieces of text are viewed as updating an existing context with new information, the result of which is an updated context.
In our case, we have a context circuit representing the text up to a certain point, and reading additional pieces of text corresponds to sequentially composing the context circuit with additional circuits encoding the new information.
Among the theories of dynamic semantics, text circuits bear a particularly close resemblance to Discourse Representation Theory (DRT)~\cite{kamp2011discourse}, and shares some of its features.
Like DRT~\cite{abzianidze-etal-2017-parallel}, DisCoCirc exhibits a certain degree of \textit{language independence}, wherein different ways of `saying the same thing', be it within one language or across different languages, look the same when translated into the formalism~\cite{Waseem2022Language}.

While text circuits have the nice linguistic properties of semantic frameworks, unlike these traditional formalisms, text circuits can also naturally accommodate the power of machine learning for the purposes of NLP related tasks.
Practical applications of DRT for instance have mainly revolved around using reasoning tools to perform inference~\cite{bjerva-etal-2014-meaning}.
Text circuits however can be easily interpreted as a specification for a dynamically generated machine learning model, in which much of the hard work in interpreting the meaning of language can be offloaded to and learned by parametrized machine learning models.
We discuss the details of this a bit more in Section~\ref{sec:discussion}.

\subsection{An example circuit}

The key idea of the DisCoCirc formalism is that each \textit{discourse referent} introduced in the text introduces a \textit{wire} in the circuit.
For the purposes of this pipeline, we will consider a discourse referent to be a noun phrase that represents a unified, coherent entity.
Each wire should be thought of as carrying semantic information about that referent.
The information in the wires are initialized by \textit{states}.
The semantic content of a sentence is represented by \textit{gates} -- processes that perform an update to the wires of the relevant referents.
Thus, a text is understood to be a process that updates meanings.
An additional feature of DisCoCirc is that complicated gates may be broken down into simpler constituents -- i.e. complicated gates can be built up in a compositional way from \textit{simple gates} and \textit{frames} (higher-order processes that act on gates).

For instance, consider a simple text consisting of two sentences:
\begin{center}
	 \texttt{Alice really likes Bob. Claire dislikes Alice.}
\end{center}
We first obtain the text circuits for each of the sentences (read from top to bottom)
\begin{center}
	\begin{minipage}{0.49\linewidth}
	\includegraphics[width=\textwidth]{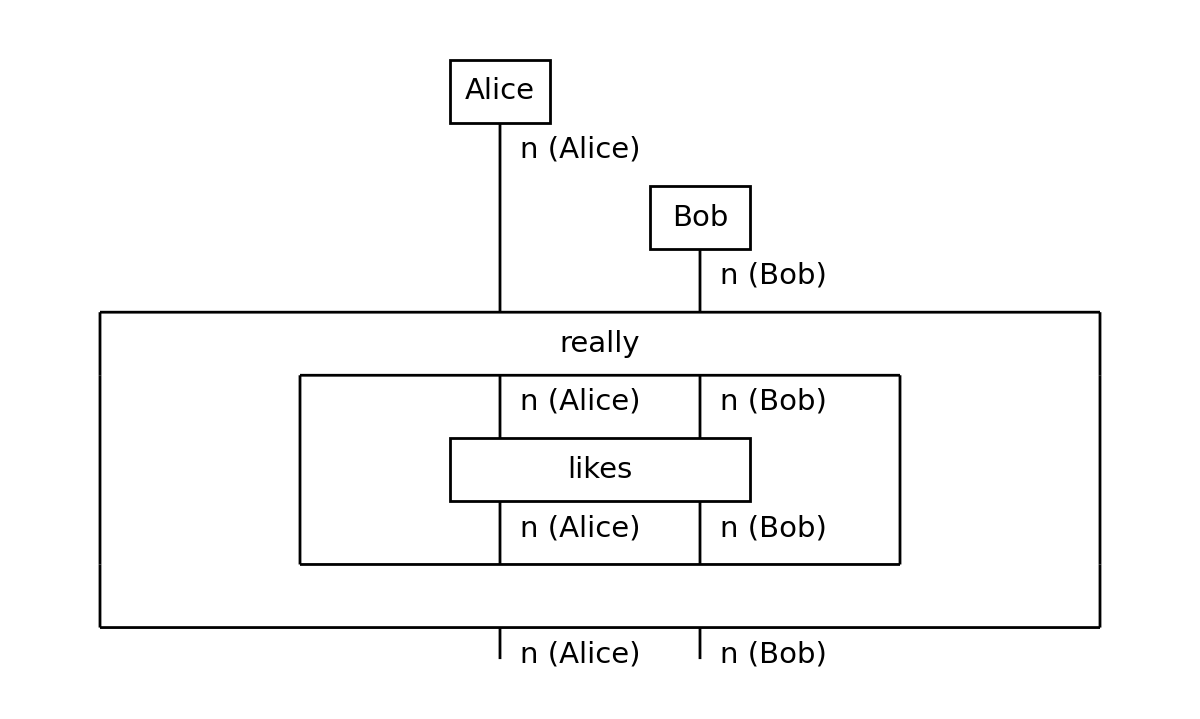}
	\begin{center} \texttt{Alice really likes Bob} \end{center}
	\end{minipage}
	\begin{minipage}{0.49\linewidth}
	\includegraphics[width=\textwidth]{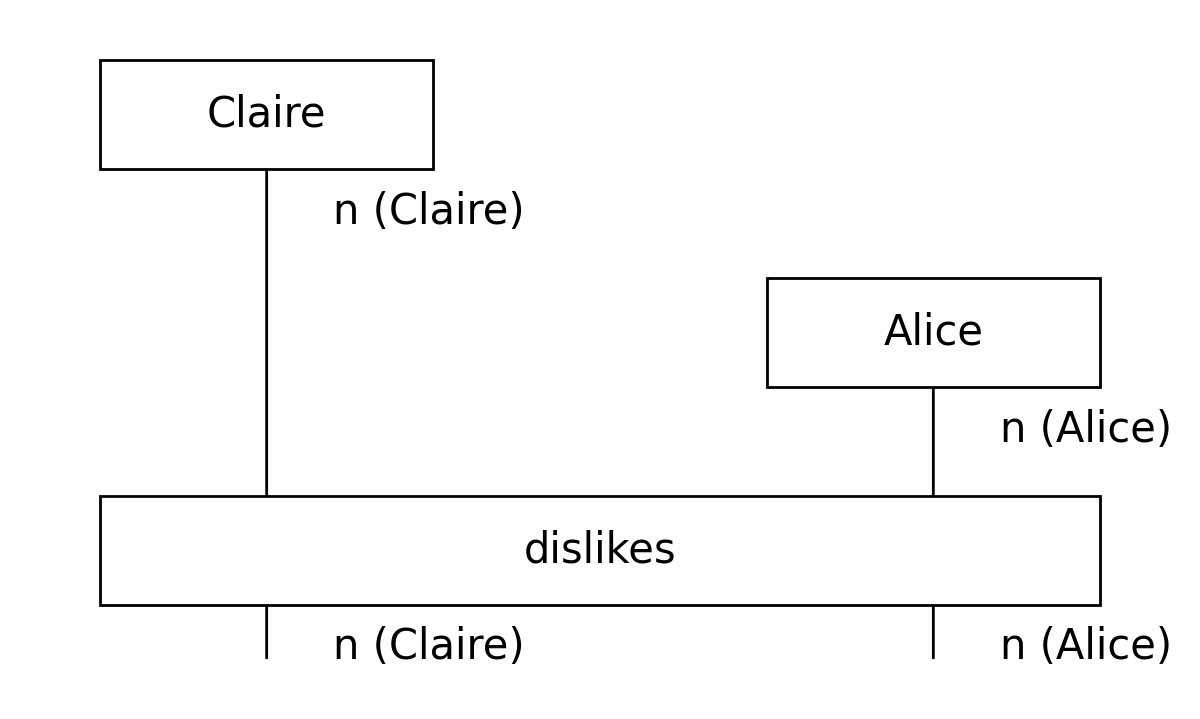}
	\begin{center} \texttt{Claire dislikes Alice} \end{center}
	\end{minipage}
\end{center}
Note that each referent (\texttt{Alice}, \texttt{Bob}, \texttt{Claire}) corresponds to a wire.
The content of the sentences is captured by the binary predicates \texttt{really likes} and \texttt{dislikes}, represented as gates acting on the wires.
In the case of the former, the \texttt{really likes} gate consists of a simple gate \texttt{likes} being modified by the frame \texttt{really}.

\begin{figure}[h]
	\caption{\texttt{Alice really likes Bob. Claire dislikes Alice.}}
	\centering
	\includegraphics[width=0.8\textwidth]{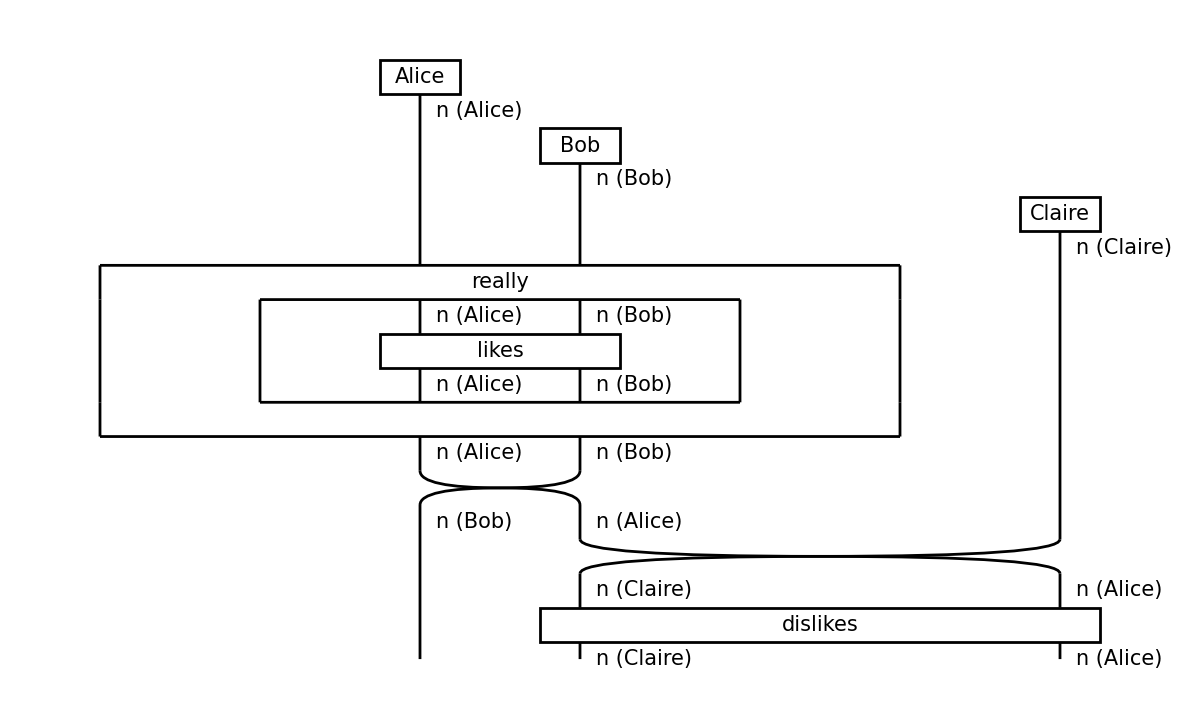}
	\label{fig:circuit_example}
\end{figure}

Having obtained these sentence circuits, we \textit{sequentially compose} them along matching referent wires to obtain the circuit for the overall text (Figure~\ref{fig:circuit_example}).
As these circuits are to be read from top to bottom, the earlier sentence goes at the top.
We call the crossed wires in the diagram a \textit{swap}, which represents a trivial passing around of information.

\subsection{Contributions}

\begin{figure*}
	\makebox[\textwidth][c]{\includegraphics[width=1.1\textwidth]{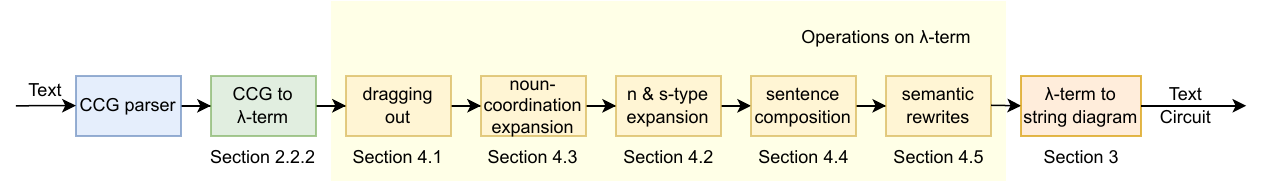}}%
	\caption{Our pipeline for generating the DisCoCirc representation of text.} 
	\label{fig:pipeline}
\end{figure*}

In this paper, we describe the first automated pipeline (Figure~\ref{fig:pipeline}) for parsing DisCoCirc structures from text.
That is, the pipeline accepts English natural language text and returns its text circuit representation.
The pipeline achieves good coverage over the fragment of English DisCoCirc discussed in~\cite{wang2023distilling}, dealing with  relative pronoun constructions, reflexive pronouns, and conjunctions of sentences.
The `sentential complement verbs' discussed in that paper are also cast as higher-order frames in our pipeline as desired.
Many semantic rewrites discussed therein are implemented as optional rewrites as part of this pipeline, including \texttt{is}-elimination, removal of passive voice, and rewrites involving possessive pronouns.

In addition, the development of the pipeline has led us to model new linguistic phenomena in the DisCoCirc formalism (e.g. phenomena not discussed in~\cite{wang2023distilling}).
These include coordinating conjunctions of arbitrary grammatical constituents, and specifically coordinating conjunctions of noun phrases which require special treatment.
Also, a new idea is the `expansion' of noun phrase wires, in addition to the existing notion of expanding sentence type wires. 

The pipeline can be categorised into four steps, as represented by the colour coding in Figure~\ref{fig:pipeline}. 

\begin{enumerate}
	\item To start, the input text is fed into the \textit{Bobcat} parser~\cite{clark2021something} from the \textit{lambeq} python package~\cite{kartsaklis2021lambeq}, to obtain combinatory categorial grammar (CCG) parses for each sentence. 
	\item Then the CCG parses are used to build a representation of the input as terms from the simply-typed $\lambda$-calculus, following a standard procedure explained in Section~\ref{ssec:ccg-to-lambda}.
	The majority of the pipeline works with these $\lambda$-terms as the underlying data structure.
	\item The next step (described in Section~\ref{sec:rewriting_diags_to_discocirc}) is the largest and constitutes the bulk of the original work involved in building this pipeline. 
	Given a $\lambda$-term corresponding to natural language text, we can already represent it in diagram form.
	However, the diagrams one gets from the $\lambda$-terms corresponding to unaltered CCG parses differ from our desired text circuits in various aspects. 
	Therefore, we have introduced five additional steps that modify the $\lambda$-terms to correct these differences.
	While the first three substeps depicted here --- dragging out (Section~\ref{ssec:dragging-out}), noun-coordination-expansion (Section~\ref{ssec:coord_expansion}) and type expansion (Section~\ref{ssec:type-expansion}) --- depend solely on the information given by the CCG parser, the step after this --- sentence composition (Section~\ref{sec:sentence-composition}) --- incorporates additional semantic information about the input text obtained from a coreference resolver.
	The last of these steps, applying \textit{semantic rewrites} (Section~\ref{sec:semantic_rewrites}), is optional and allows flexibility in tailoring the text circuit outputs to specific needs.
	\item After correcting the $\lambda$-terms accordingly, the final step in the pipeline converts the $\lambda$-terms into diagrams. 
	We describe this novel drawing algorithm in Section~\ref{sec:lambda-to-diagrams}.
	Note that this drawing method can be used at any point in the pipeline where we have $\lambda$-terms, and is used throughout the paper to showcase the impact of the steps on the $\lambda$-terms.
\end{enumerate}

\section{Preliminary background}
\label{sec:background}

\subsection{Simply-typed $\lambda$-calculus}
\label{ssec:st_lambda_calc}

We consider the simply-typed $\lambda$-calculus with \textit{function} $\to$ and \textit{product} $\times$ types.
That is, the set of types is recursively generated from some set of \textit{atomic} types, via
\begin{align*}
	T::= b \ |\ T\to T \ |\ T\times T
\end{align*}
where $b$ stands for any atomic type.
The set of $\lambda$-terms that we work with is recursively generated via \textit{application}, \textit{abstraction}, and \textit{lists}
\begin{align*}
	M ::= x \ |\ c \ |\ MM \ |\ \lambda x.M \ |\ (M,M).
\end{align*}
Here $x$ stands for a \textit{variable}, whereas $c$ stands for a \textit{constant}, which cannot be abstracted over.
Since we are using $\lambda$-terms to model natural language, concrete words (e.g. \texttt{Alice}, \texttt{likes}) will be considered constants rather than variables.
In practice however, our algorithms will deal with variables and constants similarly. Importantly, we restrict to linear $\lambda$-terms, meaning that each variable bound by an abstraction appears exactly once in the term. 

We will also be a little loose where it is clear how things should work. 
We allow list terms to be lists of arbitrary finite arity, and accordingly we allow product types to be of arbitrary finite arity.
We also allow lists of variables to be the bound variable in an abstraction.
This obviates the need for projection $\lambda$-terms.
So, for instance, $\lambda (x,y,z).(y,x,z)$ will be a valid term of type $X\times Y\times Z\to Y\times X\times Z$.

A simpler system of simply-typed $\lambda$-calculus, without product $\times$ types and $(M,M)$ list terms is sufficient to give semantics to CCG.
However, we will need to make use of products and lists for our pipeline.

$\lambda$-terms are often depicted in the form of a tree (a directed acyclic graph with an order on the edges leaving each vertex), the structure of which corresponds to the construction of the term.
The leaves of the tree are variables or constants, and the other nodes are labelled with either application ($@$), an abstracted bound variable (e.g. $\lambda x$), or a product ($\times$).
For instance, the typed $\lambda$-term 
\begin{center}
	$\lambda(x,y,z).(y,\; \texttt{dislikes}(x)(z))
	(\texttt{really}(\texttt{likes})(\texttt{Bob}) (\texttt{Alice}),\; \texttt{Claire})))$
\end{center}
which yields the diagram in Figure~\ref{fig:circuit_example} would be depicted as 
\[\begin{tikzpicture}
	\begin{pgfonlayer}{nodelayer}
		\node [style=none] (0) at (-4.5, 3.5) {$@$};
		\node [style=none] (2) at (-5.25, 4.125) {};
		\node [style=none] (3) at (-4.75, 3.625) {};
		\node [style=none] (4) at (-4.25, 3.625) {};
		\node [style=none] (5) at (-3.75, 4.125) {};
		\node [style=none] (7) at (-5.5, 4.75) {\texttt{dislikes}};
		\node [style=none] (8) at (-3.5, 4.75) {$x$};
		\node [style=none] (9) at (-3.5, 2.375) {$@$};
		\node [style=none] (10) at (-4.5, 1.25) {$\times$};
		\node [style=none] (11) at (-4.25, 3) {};
		\node [style=none] (12) at (-3.75, 2.5) {};
		\node [style=none] (13) at (-3.75, 1.875) {};
		\node [style=none] (14) at (-4.25, 1.375) {};
		\node [style=none] (15) at (-4.75, 1.375) {};
		\node [style=none] (16) at (-5.25, 1.875) {};
		\node [style=none] (17) at (-2.5, 3.5) {$z$};
		\node [style=none] (19) at (-2.75, 3) {};
		\node [style=none] (20) at (-3.25, 2.5) {};
		\node [style=none] (21) at (-5.5, 2.375) {$y$};
		\node [style=none] (23) at (-4.5, 0.75) {};
		\node [style=none] (24) at (-4.5, 0.25) {};
		\node [style=none] (25) at (-4.5, 0) {$\lambda (x,y,z)$};
		\node [style=none] (30) at (0, 4.625) {\texttt{really}};
		\node [style=none] (32) at (3.5, 4.625) {\texttt{likes}};
		\node [style=none] (34) at (1.75, 3.5) {};
		\node [style=none] (35) at (0.25, 4.125) {};
		\node [style=none] (36) at (2.25, 3.5) {};
		\node [style=none] (37) at (3.25, 4.125) {};
		\node [style=none] (38) at (2, 3.375) {$@$};
		\node [style=none] (39) at (3, 2.25) {$@$};
		\node [style=none] (40) at (4.5, 3.375) {\texttt{Bob}};
		\node [style=none] (42) at (4.25, 2.875) {};
		\node [style=none] (43) at (3.25, 2.375) {};
		\node [style=none] (44) at (2.25, 2.875) {};
		\node [style=none] (45) at (2.75, 2.375) {};
		\node [style=none] (46) at (5.5, 2.25) {\texttt{Alice}};
		\node [style=none] (48) at (5.25, 1.75) {};
		\node [style=none] (49) at (4, 1.125) {$@$};
		\node [style=none] (50) at (3.25, 1.75) {};
		\node [style=none] (51) at (3.75, 1.25) {};
		\node [style=none] (52) at (4.25, 1.25) {};
		\node [style=none] (53) at (5, 0) {$\times$};
		\node [style=none] (54) at (6.5, 1.125) {\texttt{Claire}};
		\node [style=none] (56) at (6.25, 0.625) {};
		\node [style=none] (57) at (5.25, 0.125) {};
		\node [style=none] (58) at (4.25, 0.625) {};
		\node [style=none] (59) at (4.75, 0.125) {};
		\node [style=none] (61) at (-4.25, -0.5) {};
		\node [style=none] (62) at (-0.25, -1.125) {};
		\node [style=none] (63) at (0.25, -1.125) {};
		\node [style=none] (64) at (4.75, -0.5) {};
		\node [style=none] (65) at (0, -1.25) {$@$};
		\node [style=none] (66) at (-5.5, 4.375) {$\scriptstyle n\to n\to n\times n$};
		\node [style=none] (67) at (-3.5, 4.375) {$\scriptstyle n$};
		\node [style=none] (68) at (-4.5, 3.125) {$\scriptstyle n\to n\times n$};
		\node [style=none] (69) at (-2.5, 3.125) {$\scriptstyle n$};
		\node [style=none] (70) at (-3.5, 2) {$\scriptstyle n\times n$};
		\node [style=none] (72) at (-5.5, 2) {$\scriptstyle n$};
		\node [style=none] (73) at (-4.5, 0.875) {$\scriptstyle n\times n\times n$};
		\node [style=none] (74) at (-4.5, -0.375) {$\scriptstyle n\times n\times n\to n\times n\times n$};
		\node [style=none] (75) at (0, 4.25) {$\scriptstyle (n\to n\to n\times n)\to n\to n\to n\times n$};
		\node [style=none] (76) at (3.5, 4.25) {$\scriptstyle n\to n\to n\times n$};
		\node [style=none] (77) at (2, 3) {$\scriptstyle n\to n\to n\times n$};
		\node [style=none] (78) at (4.5, 3) {$\scriptstyle n$};
		\node [style=none] (79) at (3, 1.875) {$\scriptstyle n\to n\times n$};
		\node [style=none] (80) at (5.5, 1.875) {$\scriptstyle n$};
		\node [style=none] (81) at (4, 0.75) {$\scriptstyle n\times n$};
		\node [style=none] (82) at (6.5, 0.75) {$\scriptstyle n$};
		\node [style=none] (83) at (5, -0.375) {$\scriptstyle n\times n\times n$};
		\node [style=none] (84) at (0, -1.625) {$\scriptstyle n\times n\times n$};
	\end{pgfonlayer}
	\begin{pgfonlayer}{edgelayer}
		\draw (2.center) to (3.center);
		\draw (4.center) to (5.center);
		\draw (11.center) to (12.center);
		\draw (13.center) to (14.center);
		\draw (15.center) to (16.center);
		\draw (20.center) to (19.center);
		\draw (23.center) to (24.center);
		\draw (34.center) to (35.center);
		\draw (36.center) to (37.center);
		\draw (43.center) to (42.center);
		\draw (45.center) to (44.center);
		\draw (50.center) to (51.center);
		\draw (52.center) to (48.center);
		\draw (57.center) to (56.center);
		\draw (58.center) to (59.center);
		\draw (63.center) to (64.center);
		\draw (62.center) to (61.center);
	\end{pgfonlayer}
\end{tikzpicture}\]
We will use the convention that in the case of application, the function term is on the left subtree and the argument term is on the right subtree.
We may also label each node with the type of the subterm corresponding to the subtree rooted at that node.
We will refer to these as $\lambda$-trees and use them interchangeably with $\lambda$-terms.

\subsection{Combinatory categorial grammar (CCG)}
\label{ssec:categorial-grammars}
Combinatory categorial grammar (CCG)~\cite{steedman2000syntactic} is a formal grammatical framework for natural language, derived from the broader family of categorial grammars~\cite{ajdukiewicz1978syntactic, bar1953quasi}.
A primary aim of the categorial grammar approach is to algorithmically distinguish
word strings that form grammatical sentences from non-grammatical word strings~\cite{lambek1958mathematics}.
To this end, a language in a categorial grammar is defined by first assigning each
word in the vocabulary a set of types -- this set of assignments is called the \textit{type lexicon}.
A string of words is in
the language (i.e. is considered to be a grammatically valid sentence) if it is possible to assign types to each word such that
the types can be combined using a language-independent universal set of rules
to give the sentence type $s$.

\subsubsection{Basic rules}

The set of types for a categorial grammar consists of a set of \textit{atomic types} and recursively generated \textit{functor types}.
Unlike the type system for $\lambda$-calculus, categorial grammars have two functor types
$$T::= t \ |\ T/T \ |\ T\backslash T.$$
The functor type $a\textbackslash b$ is given to a phrase that combines with a phrase of type $b$
on its left to give a phrase of type $a$,
and $a / b$ is given to a phrase that combines with a phrase of type $b$
on its right to give a phrase of type $a$ (see the `application' rule described below). For example, the version of CCG that we use contains the atomic types $s$ for sentences and $n$ for nouns. 
We use the \textit{Bobcat} parser~\cite{clark2021something} from the \textit{lambeq} python package~\cite{kartsaklis2021lambeq} to obtain parses which we then translate to this system of CCG.

The lexicon specifies the types that the words of a given natural language can take.
For instance, our lexicon for English would contain the following entries
\begin{align*}
	\texttt{Alice} &\vdash n
	\\
	\texttt{Bob} &\vdash n
	\\
	\texttt{likes} &\vdash (s\backslash n)/n
\end{align*}
Of course, it is possible for the same word (e.g. \texttt{who}) to be assigned several different types.

The type inference rules in categorial grammar come in pairs, with a forward and backward version.
In the Basic Categorial Grammar (BCG), there are only the forward and backward \textit{application} rules which combine an argument with an appropriate functor type.
\begin{align*}
	\hspace{1.8cm}\textbf{Application:}
	&&
	\AxiomC{$X/ Y$}
	\AxiomC{$Y$}
	\RightLabel{$>$}
	\BinaryInfC{$X$}
	\DisplayProof
	&&
	\AxiomC{$Y$}
	\AxiomC{$X\backslash Y$}
	\RightLabel{$<$}
	\BinaryInfC{$X$}
	\DisplayProof
\end{align*}
From a computational linguistics perspective, BCG is
weakly-equivalent to context-free grammars~\cite{bar1960categorial}, a degree of expressiveness that is known to be inadequate for natural language~\cite{bresnan1982cross, shieber1985evidence}.

In CCG, additional inference rules are introduced that increase its expressivity.
CCG is mildly context-sensitive, which allows us to capture most
linguistic features that appear in natural language~\cite{vijay1994equivalence}.
The most important additional rules are given below:
\begin{align*}
	\textbf{Composition:}
	&&
	\AxiomC{$X/ Y$}
	\AxiomC{$Y/ Z$}
	\RightLabel{$B_>$}
	\BinaryInfC{$X/ Z$}
	\DisplayProof
	&&
	\AxiomC{$Y\backslash Z$}
	\AxiomC{$X\backslash Y$}
	\RightLabel{$B_<$}
	\BinaryInfC{$X\backslash Z$}
	\DisplayProof
	\\
	\textbf{Type-raising:}
	&&
	\AxiomC{$X$}
	\RightLabel{$T_>$}
	\UnaryInfC{$Y/(Y\backslash X)$}
	\DisplayProof
	&&
	\AxiomC{$X$}
	\RightLabel{$T_<$}
	\UnaryInfC{$Y\backslash(Y/X)$}
	\DisplayProof
	\\
	\textbf{Crossed composition:}
	&&
	\AxiomC{$X/Y$}
	\AxiomC{$Y\backslash Z$}
	\RightLabel{$BX_>$}
	\BinaryInfC{$X\backslash Z$}
	\DisplayProof
	&&
	\AxiomC{$Y/ Z$}
	\AxiomC{$X\backslash Y$}
	\RightLabel{$BX_<$}
	\BinaryInfC{$X/Z$}
	\DisplayProof
	\\
	\textbf{Generalized composition:}&& &&	
\end{align*}
\begin{center}
	\AxiomC{$X/Y$}
	\AxiomC{$(\ldots((Y/Z_1)/Z_2)\ldots)/Z_n$}
	\RightLabel{$B_>^n$}
	\BinaryInfC{$(\ldots((X/Z_1)/Z_2)\ldots)/Z_n$}
	\DisplayProof
	\hskip 3.5em
	\AxiomC{$(\ldots((Y\backslash Z_1)\backslash Z_2)\ldots)\backslash Z_n$}
	\AxiomC{$X\backslash Y$}
	\RightLabel{$B_<^n$}
	\BinaryInfC{$(\ldots((X\backslash Z_1)\backslash Z_2)\ldots)\backslash Z_n$}
	\DisplayProof
	\\
	\
\end{center}
Note that generalized composition is a family of rules parametrized by natural numbers $n$.
In practice, only schemata up to some bounded $n$ are allowed (Steedman~\cite{steedman2000syntactic} assumes 4 for English). This prevents the formalism from being fully context-sensitive.
When $n=1$, the generalized composition reduces to the standard composition.
Sometimes the application rule is referred to as the $n=0$ case of generalized composition.

Here are two example derivations (using only the application rules) that witness the grammatical validity of the associated sentences.
\begin{center}
	\begin{minipage}{.3\textwidth}
		\small
		\cgex{3}{Alice & likes & Bob\\
			\cgul & \cgul & \cgul\\
			\cat{n} & \cat{(s\bs n)\fs n} & \cat{n}\\
			& \cgline{2}{\cgfa}\\
			& \cgres{2}{s\bs n}\\
			\cgline{3}{\cgba}\\
			\cgres{3}{s}\\
		}
	\end{minipage}
	\begin{minipage}{.3\textwidth}
		\small
		\cgex{5}{Colourless & green & ideas & sleep & furiously\\
			\cgul & \cgul & \cgul & \cgul & \cgul\\
			\cat{n\fs n} & \cat{n\fs n} & \cat{n} & \cat{s\bs n} & \cat{(s\bs n)\bs (s\bs n)}\\
			& \cgline{2}{\cgfa} & \cgline{2}{\cgba}\\
			& \cgres{2}{n} & \cgres{2}{s\bs n}\\
			\cgline{3}{\cgfa} & \\
			\cgres{3}{n} & \\
			\cgline{5}{\cgba} \\
			\cgres{5}{s} \\
		}
	\end{minipage}
\end{center}
Note that the derivation tree gives information about the constituency structure of the sentence.
For instance, in the second  example, we see that the phrase \texttt{sleep furiously} has the type of an intransitive verb, $s\backslash n$.

\subsubsection{Co-indexing of types}
\label{ssec:coindexing}
In practical implementations of CCG, an enriched version of the basic CCG types described above are often used.
In particular, one can introduce a \textit{co-indexing} mechanism, which helps to capture head word and long range dependency information, that in turn approximate the true semantic interpretation~\cite{Clark2002ccg,hockenmaier2005ccgbank}.
To illustrate how this works, consider the example of relative pronouns.

The CCG type for a relative pronoun is $(n\backslash n)/(s\backslash n)$ or $(n\backslash n)/(s/n)$. The way to understand this type is that the relative pronoun takes a
sentence missing a subject or an object to its right, and a noun phrase to its left.
The final result of this is also a noun phrase.
However, the noun phrase argument of the relative pronoun is also the missing object or subject of its sentential argument.
The noun phrase argument of the relative pronoun is in turn also the head of the resultant noun phrase.
This relationship is represented by co-indexing the $n$'s in the type, i.e.
$$(n\;(x)\backslash n\;(x))/(s\;(w)\backslash n\;(x)).$$

In the noun phrase \texttt{Bob who likes Alice}, the relative pronoun \texttt{who} has precisely this type.
Below is the enriched CCG parse for this sentence in which the types are equipped with co-indices, showing how the co-indexing on \texttt{who} helps to identify that the subject of \texttt{likes} is \texttt{Bob} and the object is \texttt{Alice}.
\begin{center}
	\small
	\cgex{4}{Bob & who & likes & Alice \\
		\cgul & \cgul & \cgul & \cgul\\[.35em]
		\cat{n\;(\texttt{Bob})} & 
		\cat{(n\;(x)\bs n\;(x))\fs (s\;(w)\bs n\;(x))} &
		\cat{(s\;(\texttt{likes})\bs n\;(y))\fs n\;(z)} & \cat{n\;(\texttt{Alice})}\\
		& & \cgline{2}{\cgfa}\\
		& &\cgres{2}{\cat{s\;(\texttt{likes})\bs n\;(y)}} \\
		& \cgline{2}{\cgfa}\\
		& \cgres{3}{\cat{n\;(x)\bs n\;(x)}}\\
		\cgline{2}{\cgba}\\[.35em]
		\cgres{2}{\cat{n\;(\texttt{Bob})}}
	}
\end{center}
In the forward application of \texttt{likes} to \texttt{Alice}, the unification procedure identifies $n\;(z)$ with $n\;(\texttt{Alice})$, i.e. a noun with the constant head \texttt{Bob}. In the next forward application of \texttt{who} to \texttt{likes(Alice)}, similarly the procedure identifies $n\;(y)$ with $n\;(x)$ and $s\;(w)$ with $s\;(\texttt{likes})$.
In the backward application step following this, $n\;(x)$ and therefore $n\;(y)$ is identified with $n\;(\texttt{Bob})$. Thus, we deduce that the subject $n\;(y)$ of the verb \texttt{likes} is \texttt{Bob} and the object $n\;(z)$ is \texttt{Alice}.

In the rest of the paper, we will generally omit the co-index labels on types.
However, we will usually depict the co-index information in the circuit diagrams -- e.g. wires in the diagram will be labelled with types like $n\;(\texttt{Alice})$ or $n\;(\texttt{Bob})$.

\subsubsection{$\lambda$-calculus semantics}
\label{ssec:ccg-to-lambda}

In CCG, every syntactic derivation has a corresponding semantic interpretation in the language of the simply-typed $\lambda$-calculus.
To convert a CCG parse tree to a $\lambda$-term, we first convert the CCG types to $\lambda$ types.
The atomic types of CCG are directly mapped to the atomic types of the $\lambda$-calculus.
For the functor types, both $a \backslash b$ and $a / b$ are mapped to $b \to a$.
Therefore, a word $\texttt{likes}$ that receives the CCG type $(s\backslash n)/n$ will be interpreted as a $\lambda$-term constant with type $n\to n\to s$.
Finally, we can combine the $\lambda$-terms based on the rules of CCG, as shown below: \\
\makebox[\textwidth][c]{\resizebox{1.1\textwidth}{!}{
\begin{tabular}{ lllll }
	\emph{Forward application} & $f: B\to A$ & $a:B$ & $\Rightarrow$ & $f(a): A$\\ 
	\emph{Backward application} & $a:B$ & $f: B\to A$ & $\Rightarrow$ & $f(a): A$\\ 
	\emph{Forward composition} & $f: B\to A$ & $g: C\to B$ & $\Rightarrow_B$ & $\lambda x. f(g(x)): C\to A$\\ 
	\emph{Backward composition} & $g: C\to B$ & $f: B\to A$ & $\Rightarrow_B$ & $\lambda x. f(g(x)): C\to A$\\ 
	\emph{Forward type-raising} & $a:A$ &  & $\Rightarrow_T$ & $\lambda f.f(a): (A \to B) \to B $\\ 
	\emph{Backward type-raising} & $a:A$ &  & $\Rightarrow_T$ & $\lambda f.f(a): (A \to B) \to B $\\ 
	\emph{Forward crossed composition} & $f: B \to A$ & $g: C \to B$ & $\Rightarrow$ & $\lambda x. f(g(x)): C \to A$\\ 
	\emph{Backward crossed composition} & $g: C \to B$ & $f: B \to A$ & $\Rightarrow$ & $\lambda x. f(g(x)): C \to A$\\ 
	\emph{Generalized forward composition} & $f: B_0 \to A$ & $g: B_n \to \dots \to B_0$ & $\Rightarrow$ & $\lambda x_n\cdots \lambda x_1.f(g (x_n)\cdots (x_1))$\\
	&&&& $\qquad \ : B_n\to\cdots\to B_1\to A$\\ 
	\emph{Generalized backward composition} & $g: B_n \to \dots \to B_0$ & $f: B_0 \to A$ & $\Rightarrow$ & $\lambda x_n\cdots \lambda x_1.f(g (x_n)\cdots (x_1))$\\
	&&&& $\qquad \ : B_n\to\cdots\to B_1\to A$\\ 
\end{tabular}}}

\paragraph{Example}
The CCG parse given earlier for the sentence \texttt{colourless green ideas sleep furiously} is mapped to the $\lambda$-term
\[\begin{tikzpicture}
	\begin{pgfonlayer}{nodelayer}
		\node [style=none] (2) at (1.75, 1.25) {$@$};
		\node [style=none] (3) at (-1.25, 1.25) {\texttt{colourless}};
		\node [style=none] (4) at (2.625, 1.875) {};
		\node [style=none] (5) at (2.125, 1.375) {};
		\node [style=none] (6) at (1.375, 1.375) {};
		\node [style=none] (7) at (0.875, 1.875) {};
		\node [style=none] (8) at (-1.25, 0.875) {$\scriptstyle n\to n$};
		\node [style=none] (9) at (1.75, 0.875) {$\scriptstyle n$};
		\node [style=none] (10) at (0.25, 0.125) {$@$};
		\node [style=none] (11) at (3, 2.375) {\texttt{ideas}};
		\node [style=none] (12) at (1.375, 0.75) {};
		\node [style=none] (13) at (0.625, 0.25) {};
		\node [style=none] (14) at (-0.125, 0.25) {};
		\node [style=none] (15) at (-0.875, 0.75) {};
		\node [style=none] (16) at (3, 2) {$\scriptstyle n$};
		\node [style=none] (17) at (0.25, -0.25) {$\scriptstyle n$};
		\node [style=none] (18) at (0.5, 2.375) {\texttt{green}};
		\node [style=none] (19) at (0.5, 2) {$\scriptstyle n\to n$};
		\node [style=none] (20) at (3, -1) {$@$};
		\node [style=none] (21) at (0.625, -0.375) {};
		\node [style=none] (22) at (2.625, -0.875) {};
		\node [style=none] (23) at (3.375, -0.875) {};
		\node [style=none] (24) at (5.375, -0.375) {};
		\node [style=none] (25) at (3, -1.375) {$\scriptstyle s$};
		\node [style=none] (26) at (5.75, 0.25) {$@$};
		\node [style=none] (27) at (6.625, 0.875) {};
		\node [style=none] (28) at (6.125, 0.375) {};
		\node [style=none] (29) at (5.375, 0.375) {};
		\node [style=none] (30) at (4.875, 0.875) {};
		\node [style=none] (31) at (5.75, -0.125) {$\scriptstyle n \to s$};
		\node [style=none] (32) at (7, 1.375) {\texttt{sleep}};
		\node [style=none] (33) at (7, 1) {$\scriptstyle n\to s$};
		\node [style=none] (34) at (4.5, 1.375) {\texttt{furiously}};
		\node [style=none] (35) at (4.5, 1) {$\scriptstyle (n\to s)\to n\to s$};
	\end{pgfonlayer}
	\begin{pgfonlayer}{edgelayer}
		\draw (4.center) to (5.center);
		\draw (7.center) to (6.center);
		\draw (12.center) to (13.center);
		\draw (15.center) to (14.center);
		\draw (21.center) to (22.center);
		\draw (24.center) to (23.center);
		\draw (27.center) to (28.center);
		\draw (30.center) to (29.center);
	\end{pgfonlayer}
\end{tikzpicture}\]

\section{Converting $\lambda$-terms to diagrams}
\label{sec:lambda-to-diagrams}

In this section, we describe our method for converting a typed $\lambda$-term to a diagram.
First, we describe how we draw constant $\lambda$-terms as diagrams.
Then, we show how we interpret the $\lambda$-calculus construction rules -- application, abstraction, and list -- as diagram composition rules.

\subsection{Constant terms}

Our constant $\lambda$-terms correspond to words and other tokens from text, with types initially obtained from their CCG types.
We draw them based on the `order' of their type.
\begin{itemize}
	\item \emph{0th-order} types are atomic types.
	We call components with these types \textit{states}.
	For us, these will correspond to nouns with type $n$.
	We depict nouns as a box with no input wires.
	For example, the noun \texttt{Alice} will be depicted as
	\begin{center}
		\includegraphics[width=0.15\textwidth]{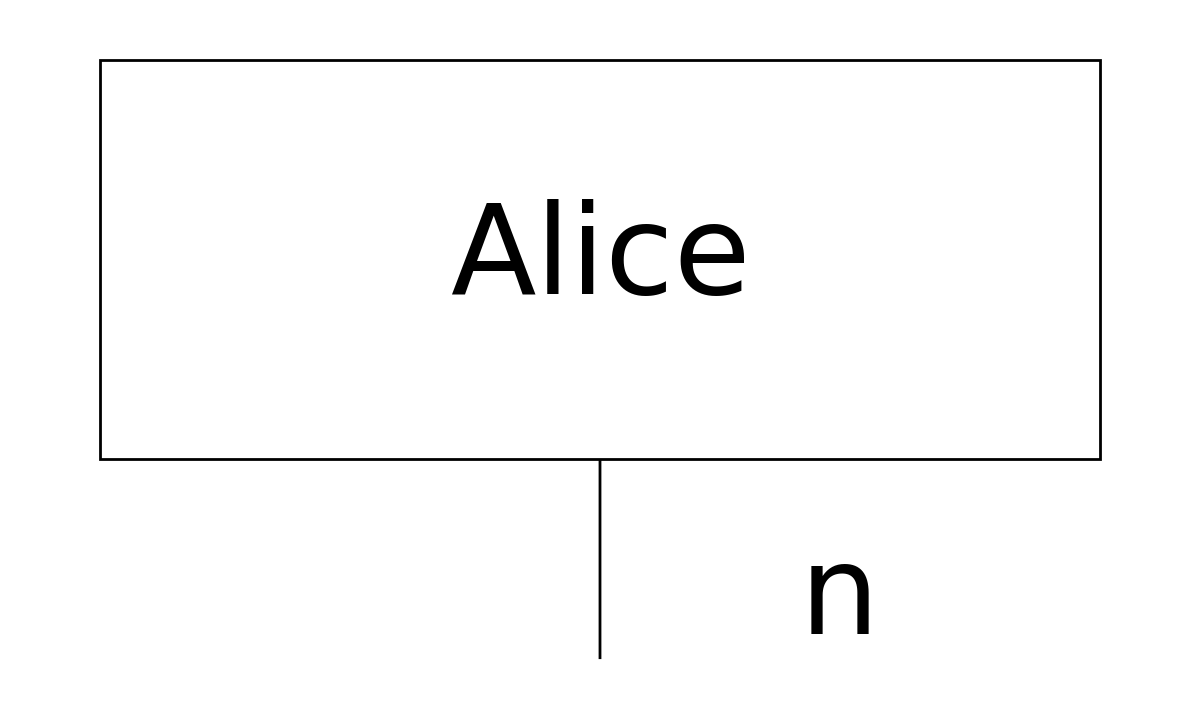}
	\end{center}
	Note that in our diagrams, wires will be labelled with atomic types (i.e. $s$ or $n$).
	We can also include the corresponding co-indices on these types.

	\item \emph{1st-order} types are function types where the type of each input is atomic (i.e.\ 0th order). We call components with these types \textit{gates}.
	These include adjectives, such as \texttt{red} : $n\to n$, and verbs, such as \texttt{walks} : $n\to s$, \texttt{runs} : $n\to n\to s$, or \texttt{gives} : $n\to n\to n\to s$.
	For example, the verb \texttt{walks} will be depicted as
	\begin{center}
		\includegraphics[width=0.18\textwidth]{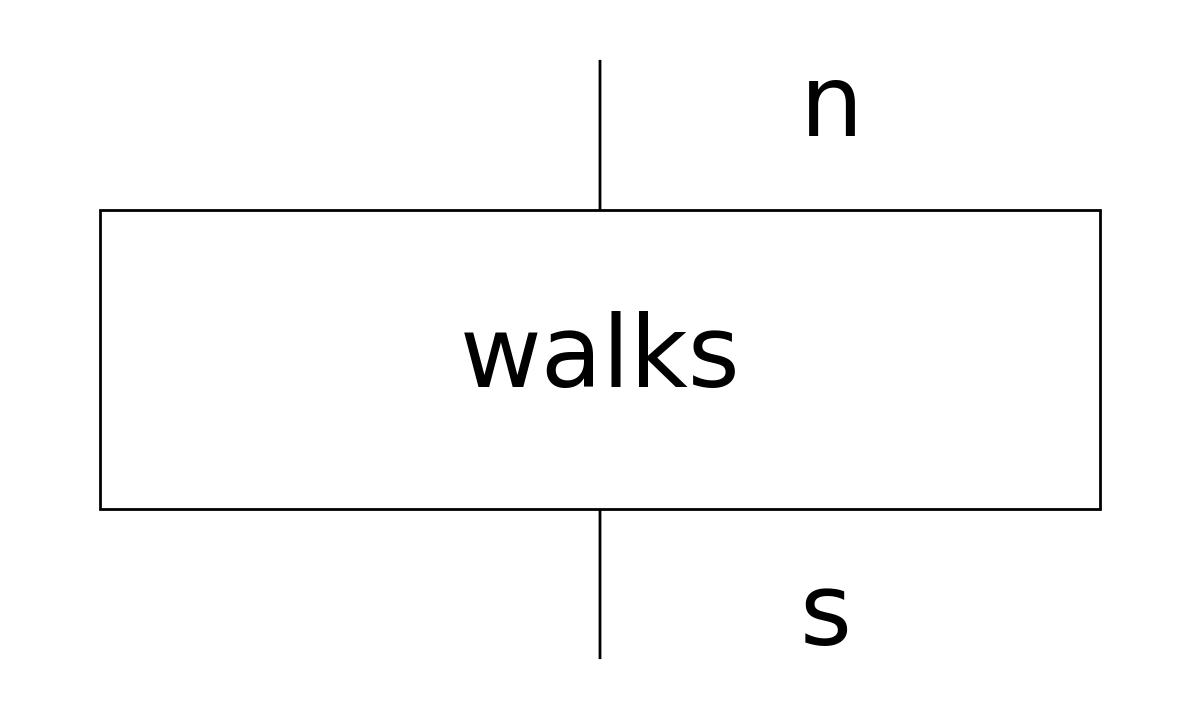}
	\end{center}

	\item \emph{2nd-order} types are function types where at least one of their inputs is 1st-order. 
	We call components with these types \textit{frames} because diagrammatically they are depicted as a box with a hole.
	Sometimes we will also refer to 2nd-order components as \textit{higher-order}.
	These include adverbs, such as \texttt{quickly} : $(n\to s)\to n\to s$, relative pronouns, such as \texttt{who} : $(n\to s)\to n\to n$, and conjunctions, such as \texttt{and} : $(n\to s)\to (n\to s)\to n\to s$.
	For example, the adverb \texttt{quickly} will be depicted as
	\begin{center}
		\includegraphics[width=0.4\textwidth]{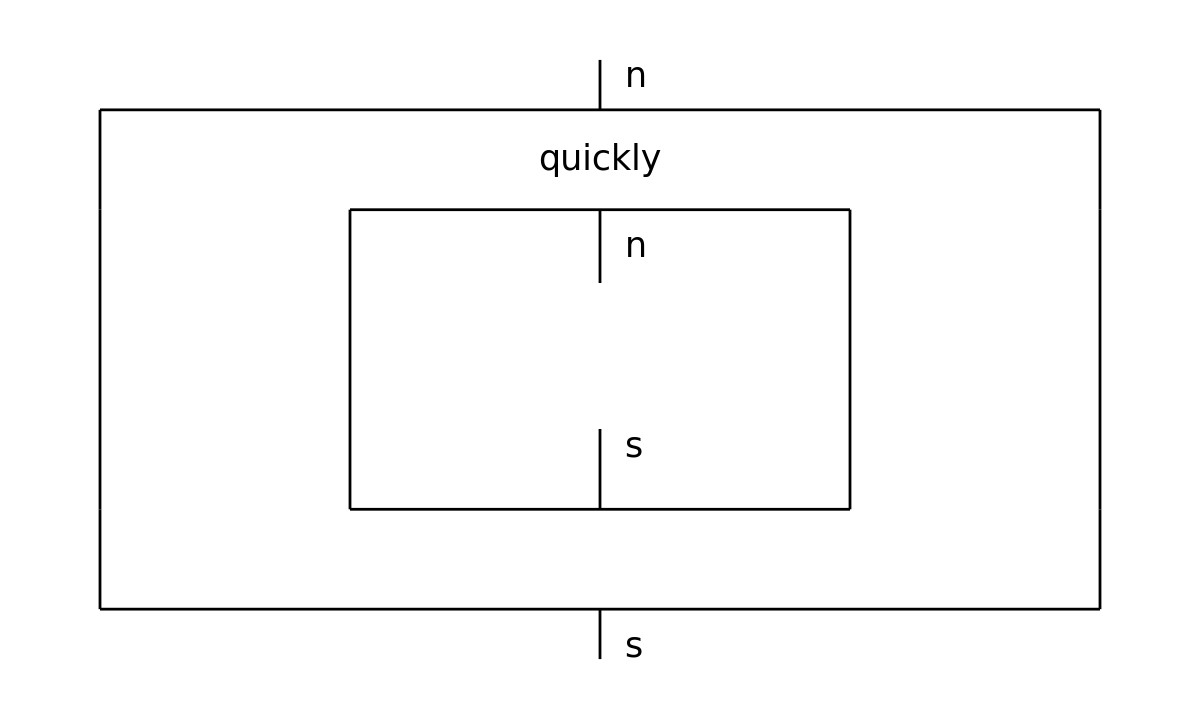}
	\end{center}
\end{itemize}
One notable exception in the above description is that we consider $s$-types to be inherently 1st-order, rather than 0th-order like $n$-types.

The motivation for doing this can be illustrated by considering `sentential complement verbs' like \texttt{dreamt}
in the sentence
\begin{align*}
	\texttt{I dreamt Bob flew}.
\end{align*}
The CCG type of \texttt{dreamt} here corresponds to the simple type $s\to n\to s$.
This type only has atomic input types, and if we treated $s$ as just another atomic type like $n$, the resulting diagram for this sentence would be
\begin{center}
	\includegraphics[width=0.5\textwidth]{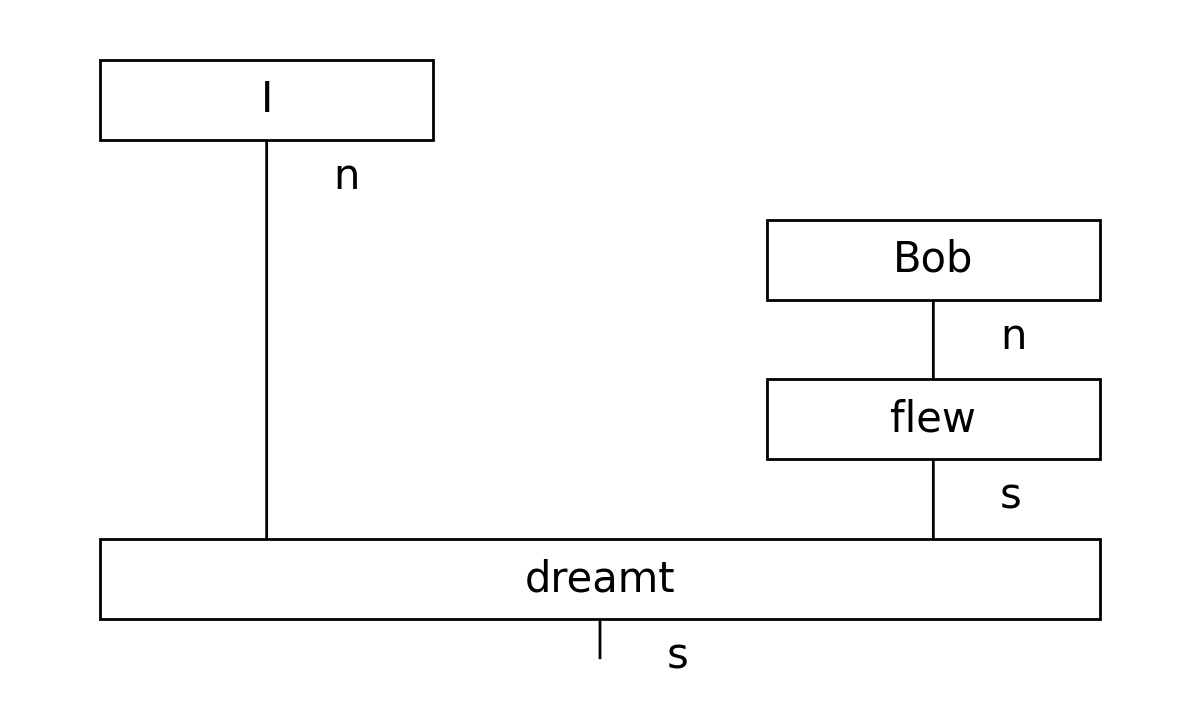}
\end{center}
However, the $s$-type in the input of \texttt{dreamt} represents a clause (\texttt{Bob flew}), which itself must contain a verb (\texttt{flew}), which is a 1st-order component.
Therefore, in order to reflect that the sentential complement verb acts on the content of an entire clause, we want the diagram component corresponding to the clause to go inside a hole of the frame corresponding to the sentential complement verb
\begin{center}
	\includegraphics[width=0.5\textwidth]{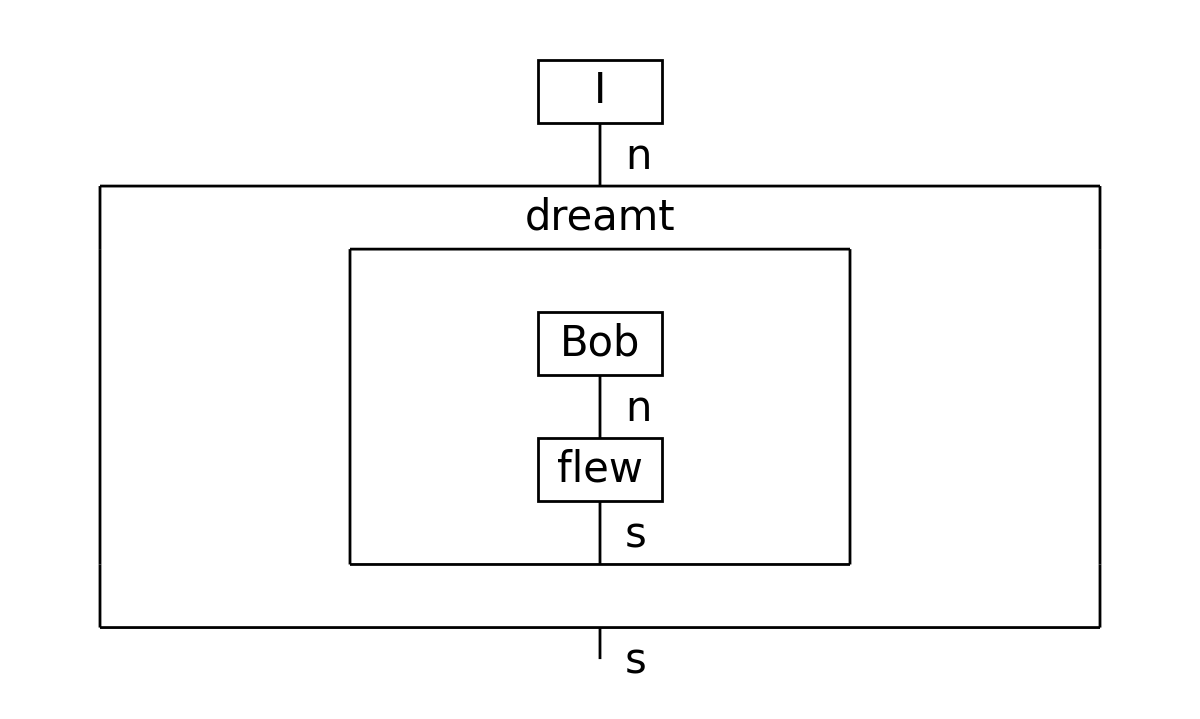}
\end{center}
This is achieved exactly by treating $s$ as a 1st-order type.
The same reasoning applies in other scenarios involving $s$-type inputs.
This treatment of $s$-types as higher-order is important to the algorithms we will perform on these $\lambda$-terms.

\subsection{Composite terms}

Now we look at converting composite $\lambda$-terms to diagrams.
We describe how the rules of the $\lambda$-calculus are interpreted as diagram composition rules.
\\
\\
\textbf{Application:}
Given two terms corresponding to the function and argument, we can compose them to get the application term.
The drawing of this term can be divided into two cases.
If the argument is a state, then we can draw the application term as a gate (or a frame, if the function term is 2nd-order) with the state plugged into the right-most input on top. 
By convention, we compose from right to left for multiple applications.
For example, the term \texttt{dislikes(Alice)} is depicted as
\begin{center}
	\includegraphics[width=0.2\textwidth]{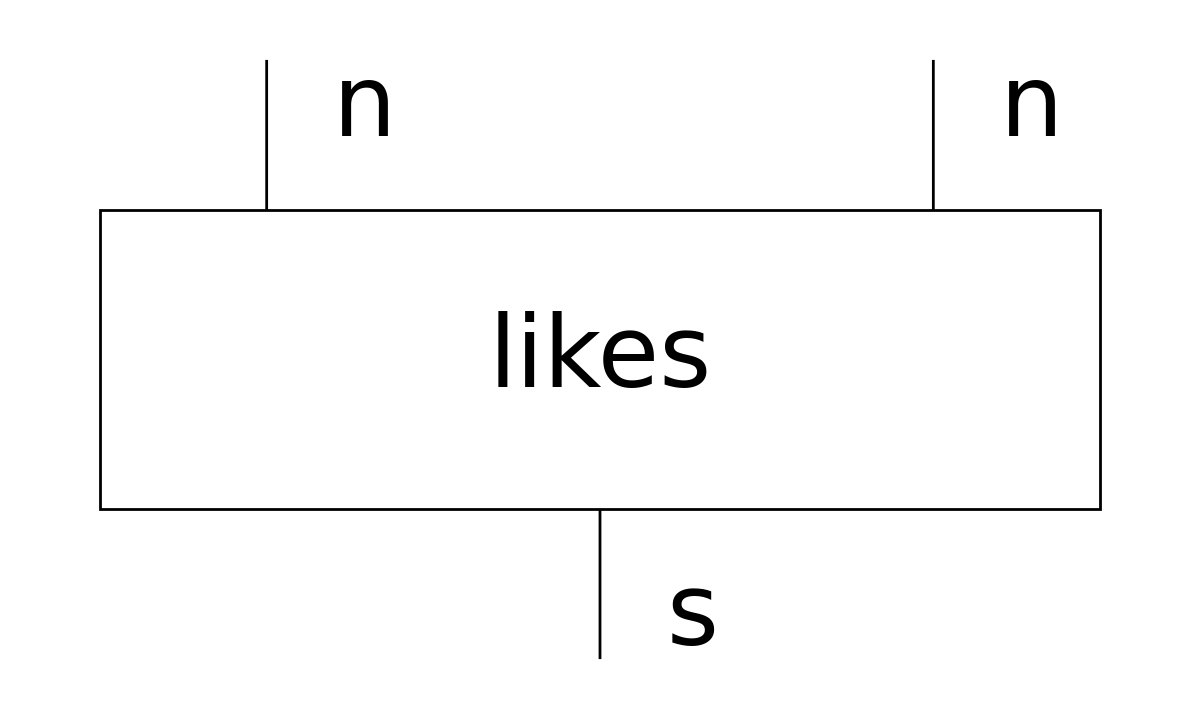}
	\includegraphics[width=0.2\textwidth]{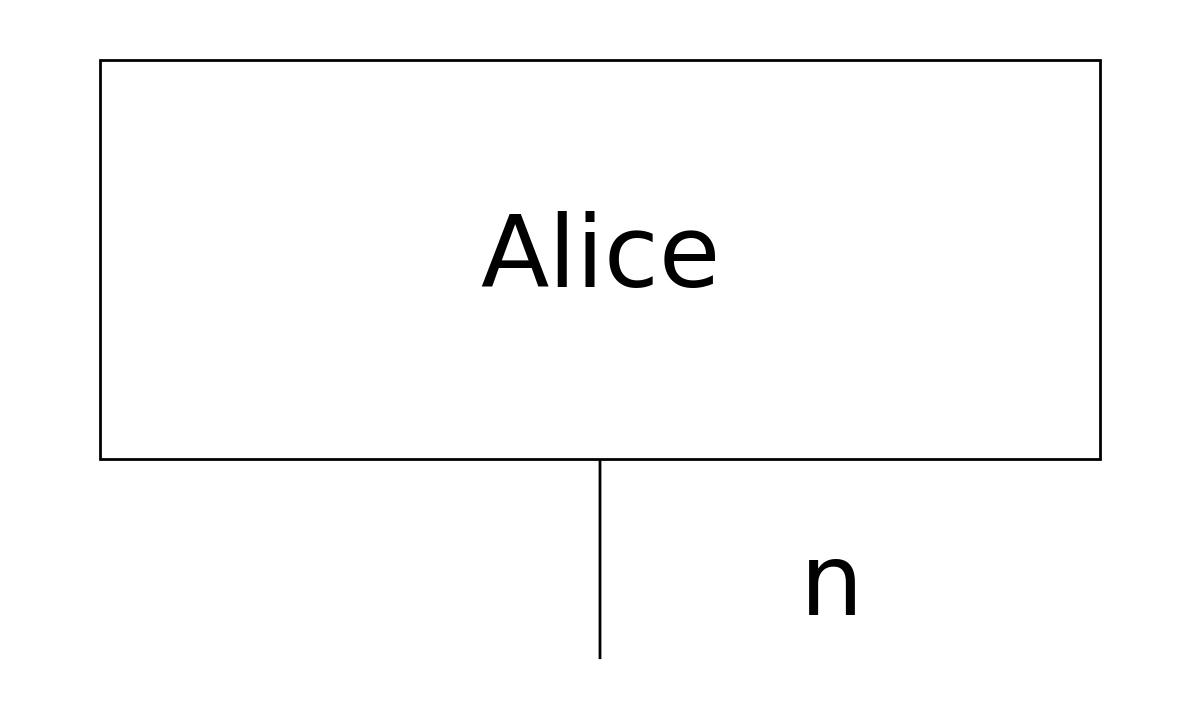}\\
	\rule{8cm}{0.8pt}\\
	\includegraphics[width=0.3\textwidth]{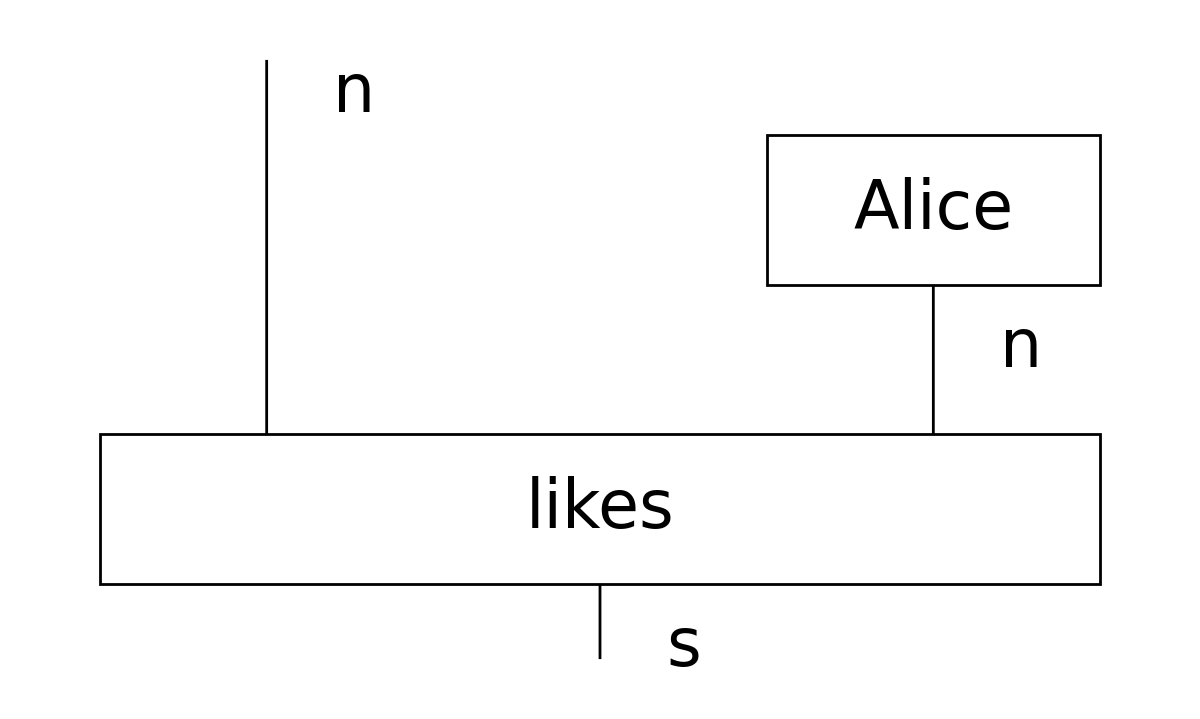}
\end{center}
If the argument is not a state, then we can draw the application term as a frame with the argument plugged into the right-most available hole.
For example, the term \texttt{who(likes(Claire))} is depicted as
\begin{center}
	\includegraphics[width=0.3\textwidth]{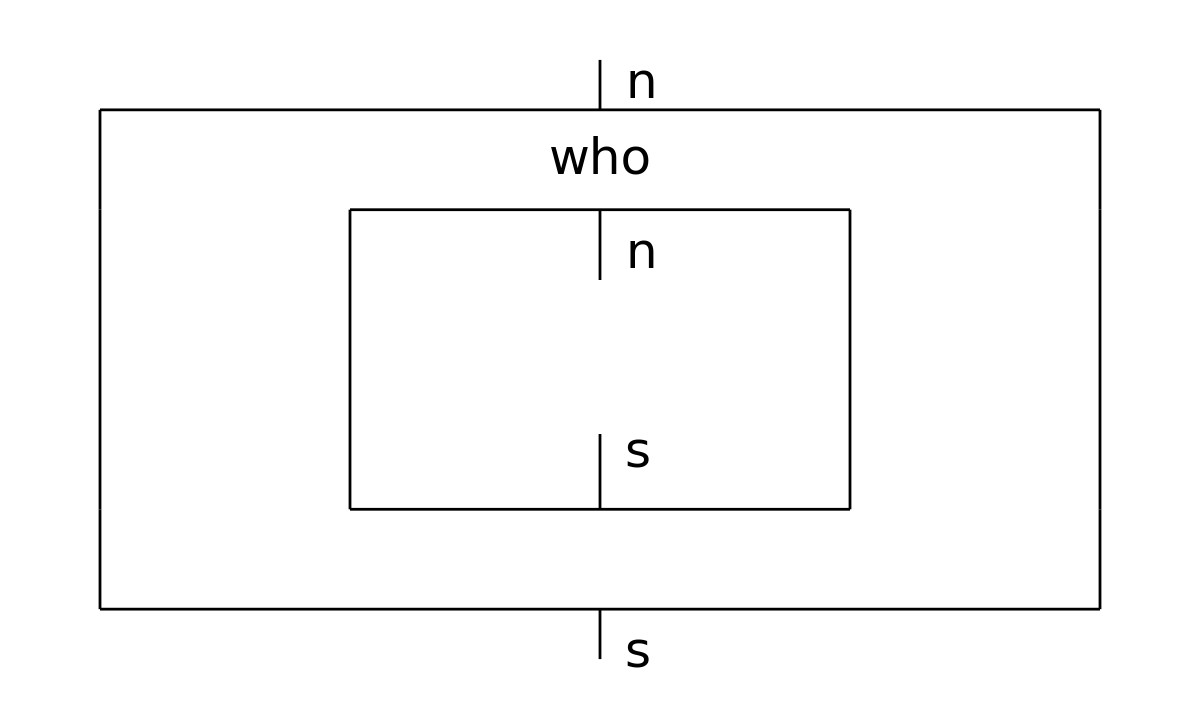}
	\includegraphics[width=0.3\textwidth]{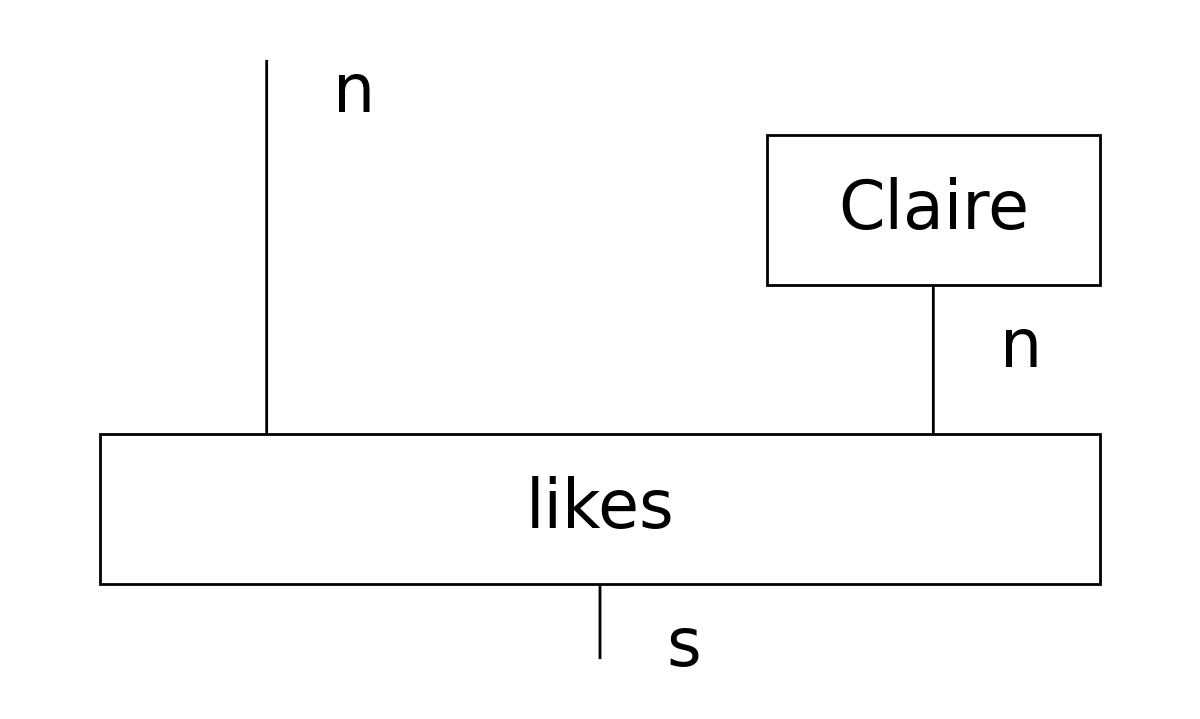}\\
	\rule{8cm}{0.8pt}\\
	\includegraphics[width=0.3\textwidth]{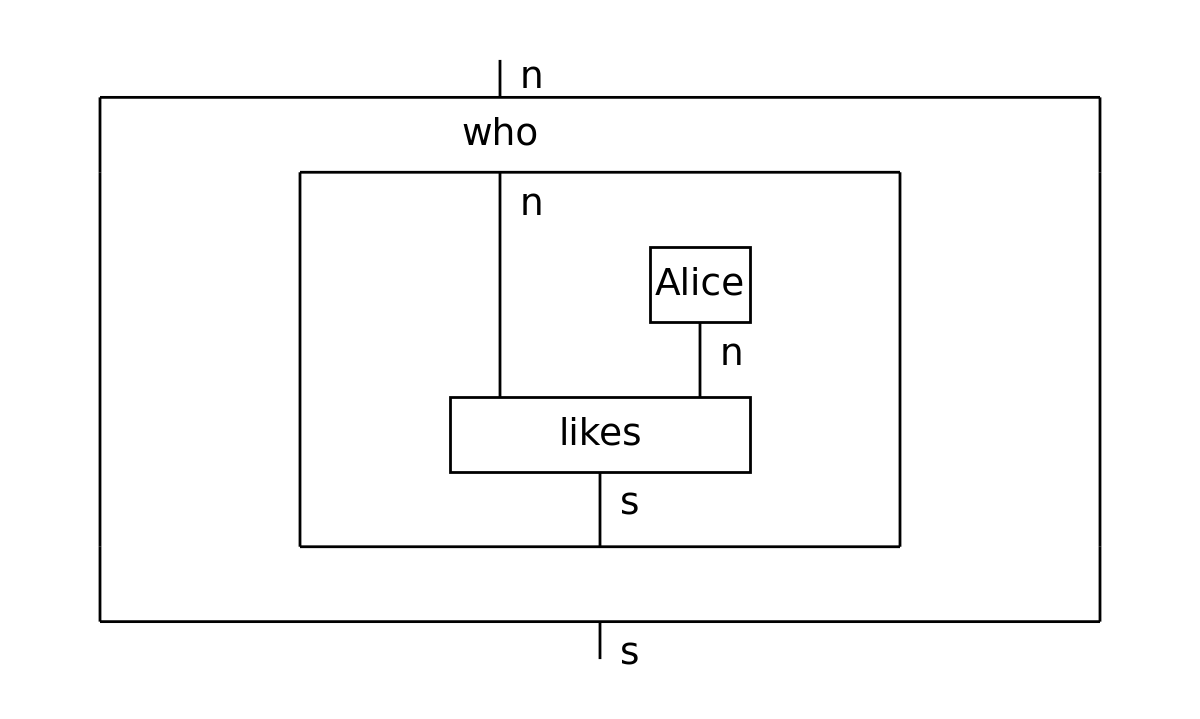}
\end{center}
As we can see from these examples, the open wires appearing in the diagram of a typed term $M$ correspond to the atomic types appearing in the type of $M$.
The application of two terms, i.e. a function term typed $A\to B$ to an argument term typed $A$, effectively unifies the type $A$ of the argument term with the $A$ in the input of the function term.
This is reflected at the diagrammatic level, where application amounts to connecting up the open wires in the argument term with the open wires in the function term.
\\

\noindent \textbf{Abstraction:}
To construct a diagram for an abstraction term, the diagram for the body is first drawn, where the bound variable is simply drawn as a constant.
Then when abstraction is done, the part of the diagram corresponding to the bound variable is removed and replaced with empty space.
The two figures below show examples for abstraction with gates and frames.
\begin{figure}[H]
	\centering
	\begin{subfigure}{.5\textwidth}
	  \centering
	  \includegraphics[width=.4\linewidth]{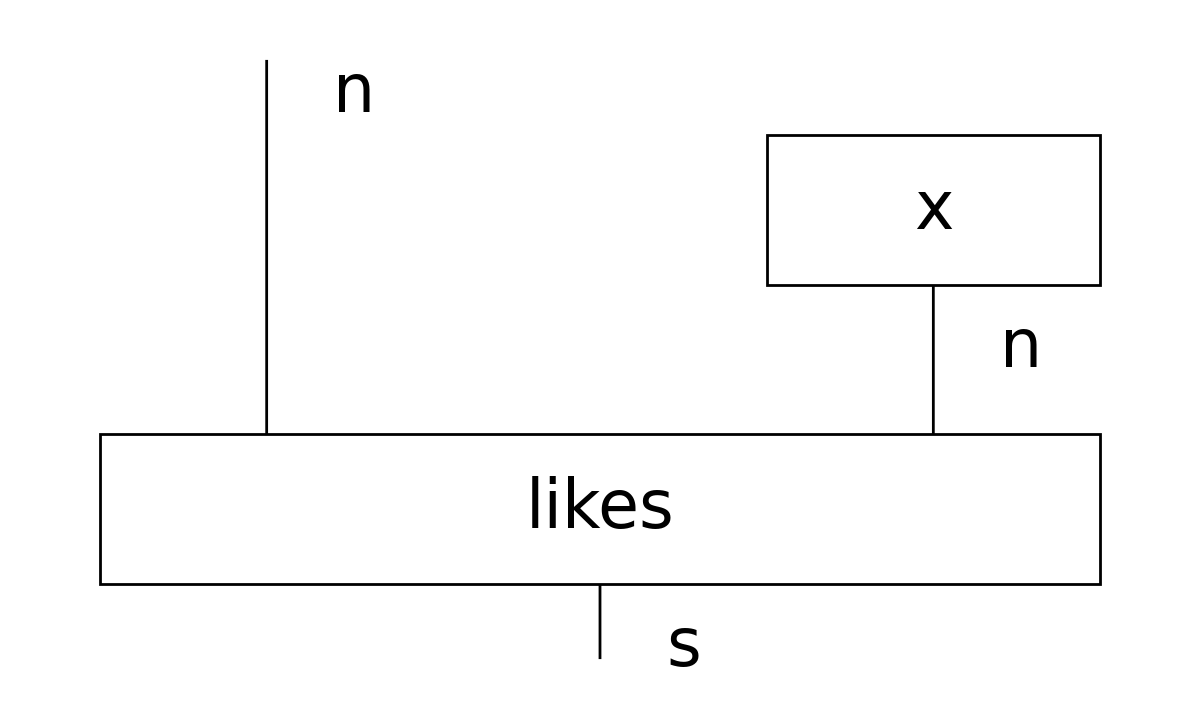} \\
	  \rule{6cm}{0.5pt} $\lambda x$\\
	  \includegraphics[width=.4\linewidth]{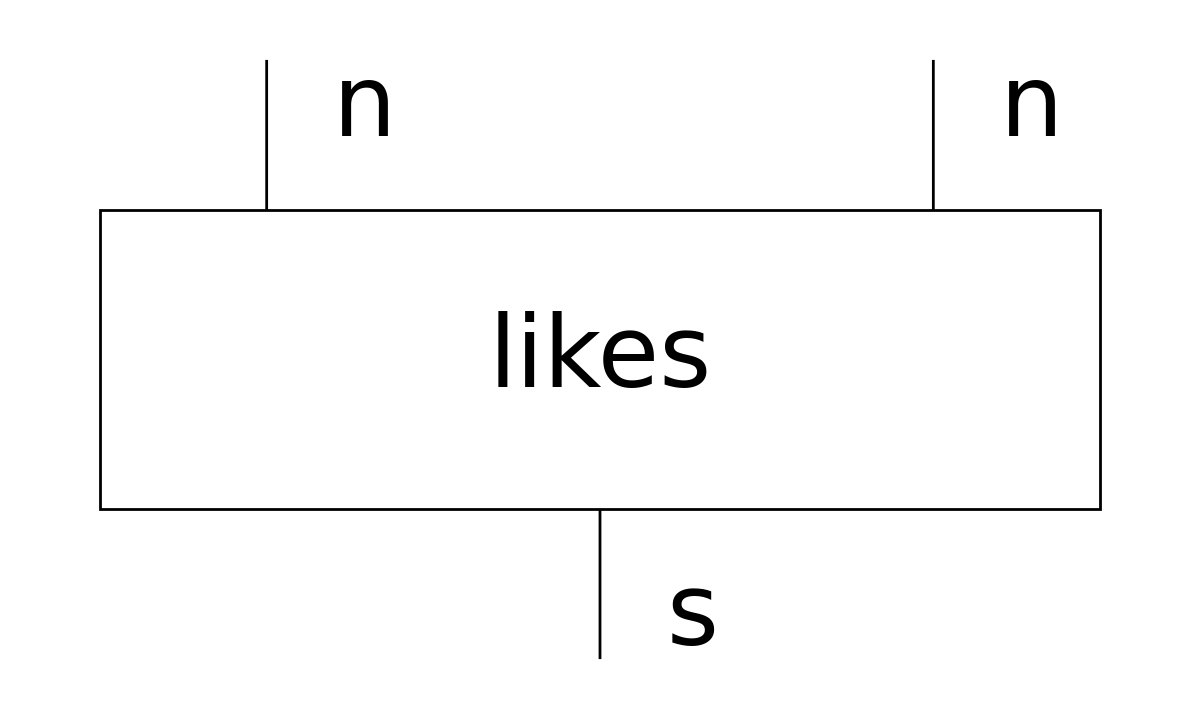} \\
	\end{subfigure}%
	\begin{subfigure}{.5\textwidth}
	  \centering
	  \includegraphics[width=.4\linewidth]{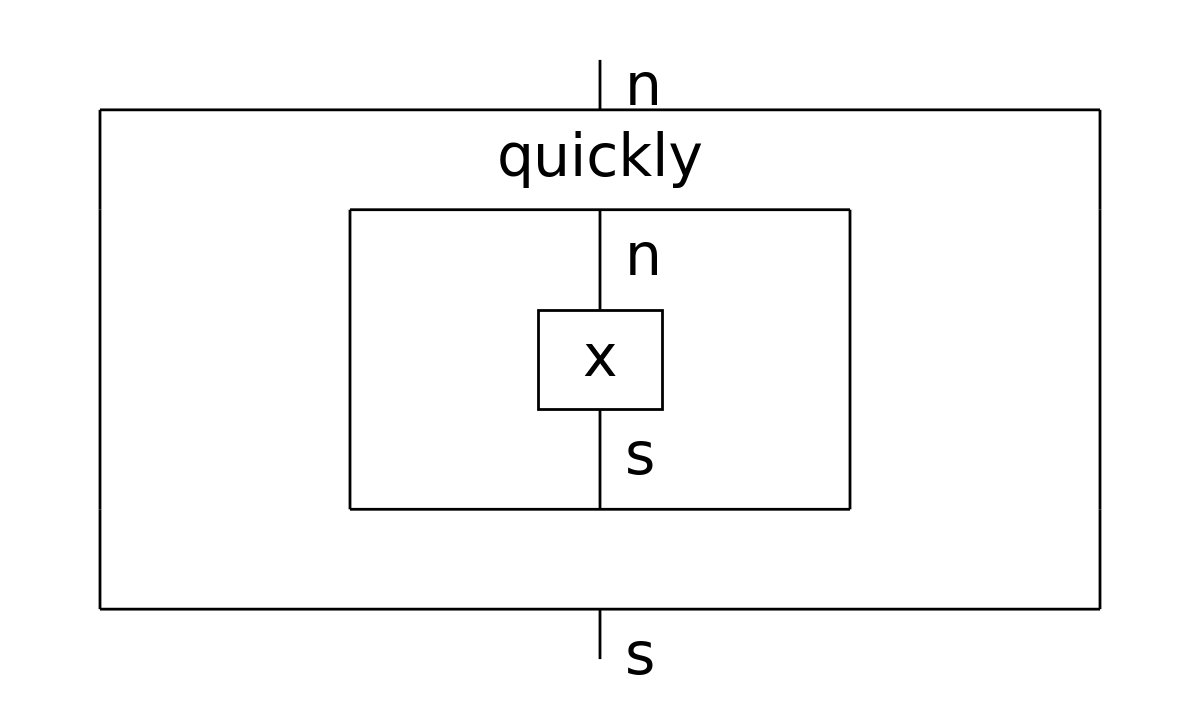} \\
	  \rule{6cm}{0.5pt} $\lambda x$\\
	  \includegraphics[width=.4\linewidth]{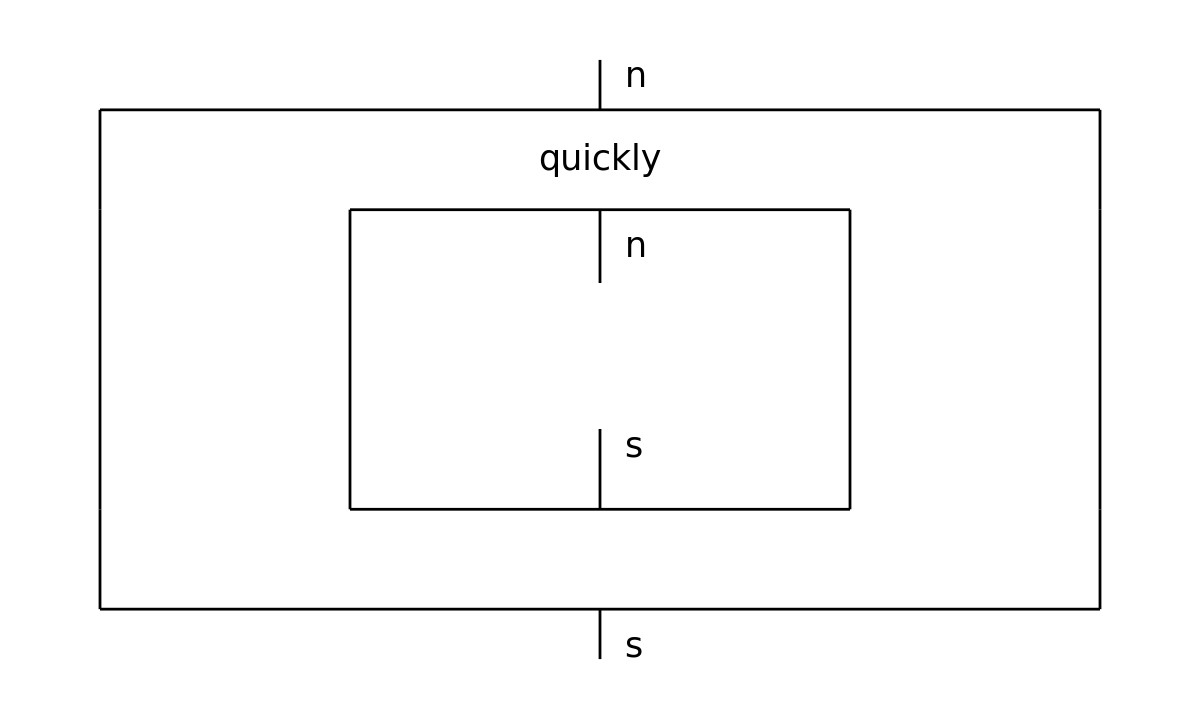} \\
	\end{subfigure}
\end{figure}
The drawing of abstractions relies on the fact that we restrict to linear $\lambda$-terms which guarantees that each bound variable appears exactly once in the expression. 

When the abstraction has multiple variables, we need to be careful about the order in which we draw the open wires.
For example, the term 
$\lambda (x,y).(\texttt{likes}(x))(y)$
is depicted diagrammatically with a swap
\begin{center}
	\includegraphics[width=0.3\textwidth]{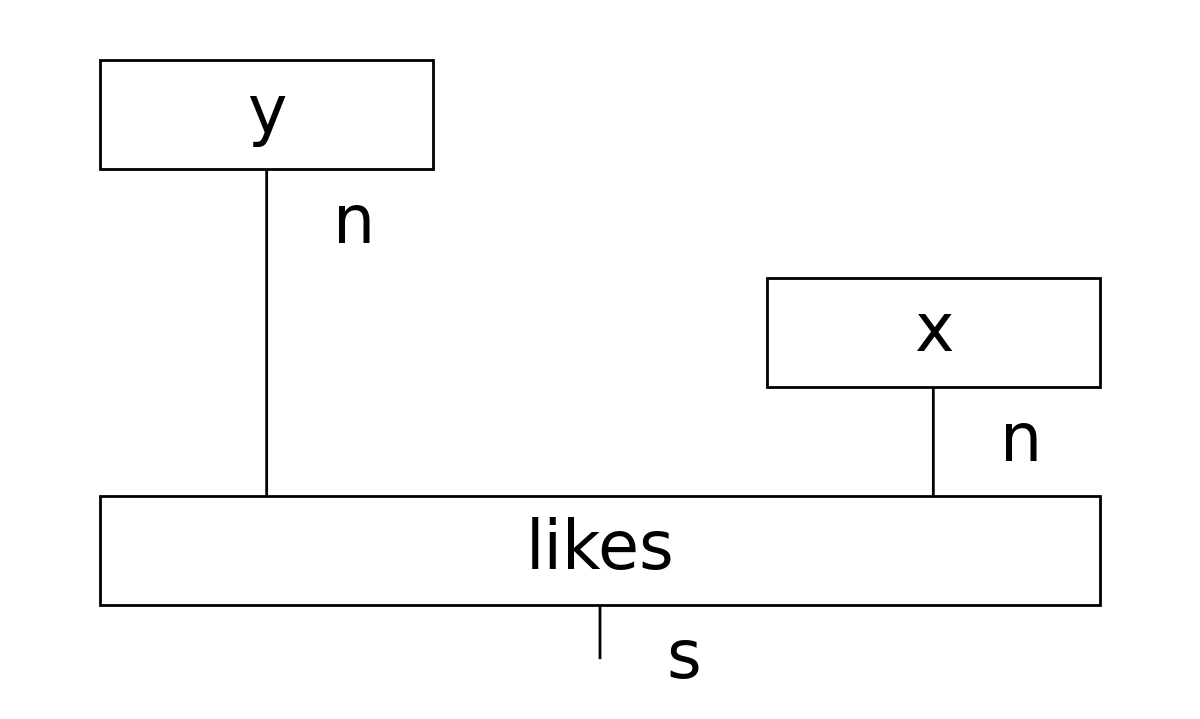}\\
	\rule{8cm}{0.8pt} $\lambda (x,y)$\\
	\includegraphics[width=0.3\textwidth]{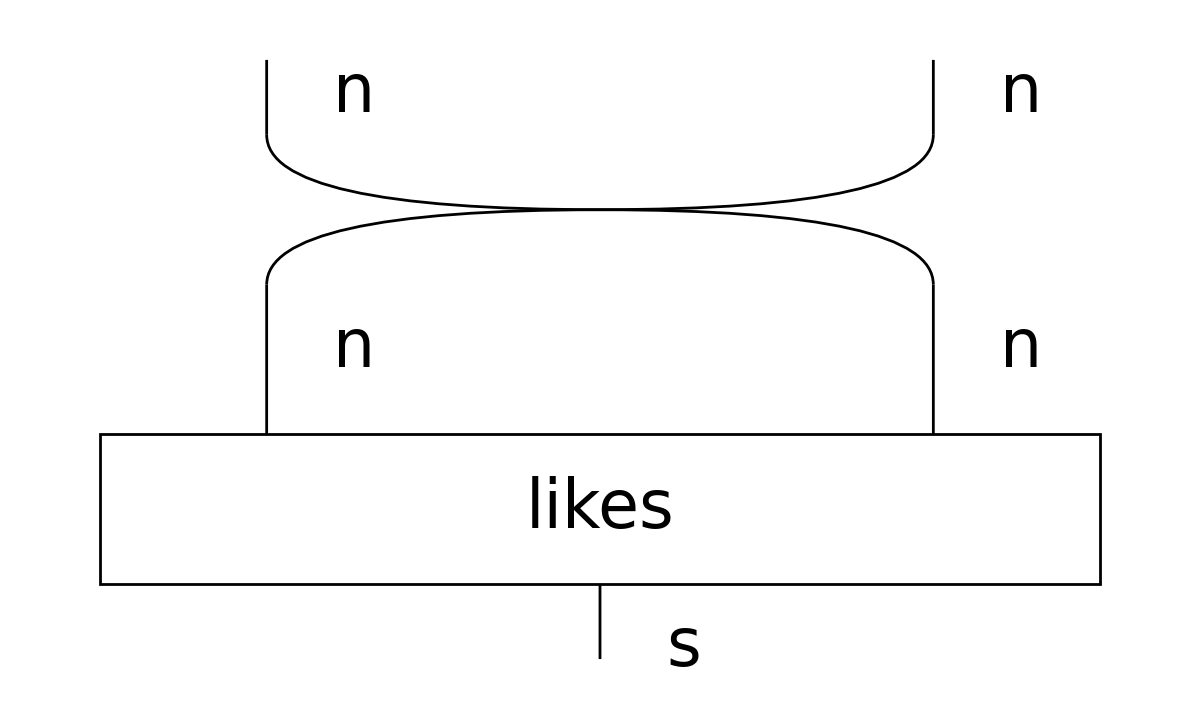}
\end{center}
Indeed, abstraction terms can be used to create arbitrary swaps as needed.

\noindent \textbf{List:}
To draw a list term, the elements are drawn next to each other in order (also known as parallel composition).\\
\begin{minipage}{\linewidth}
\begin{center}
	\includegraphics[width=0.2\textwidth]{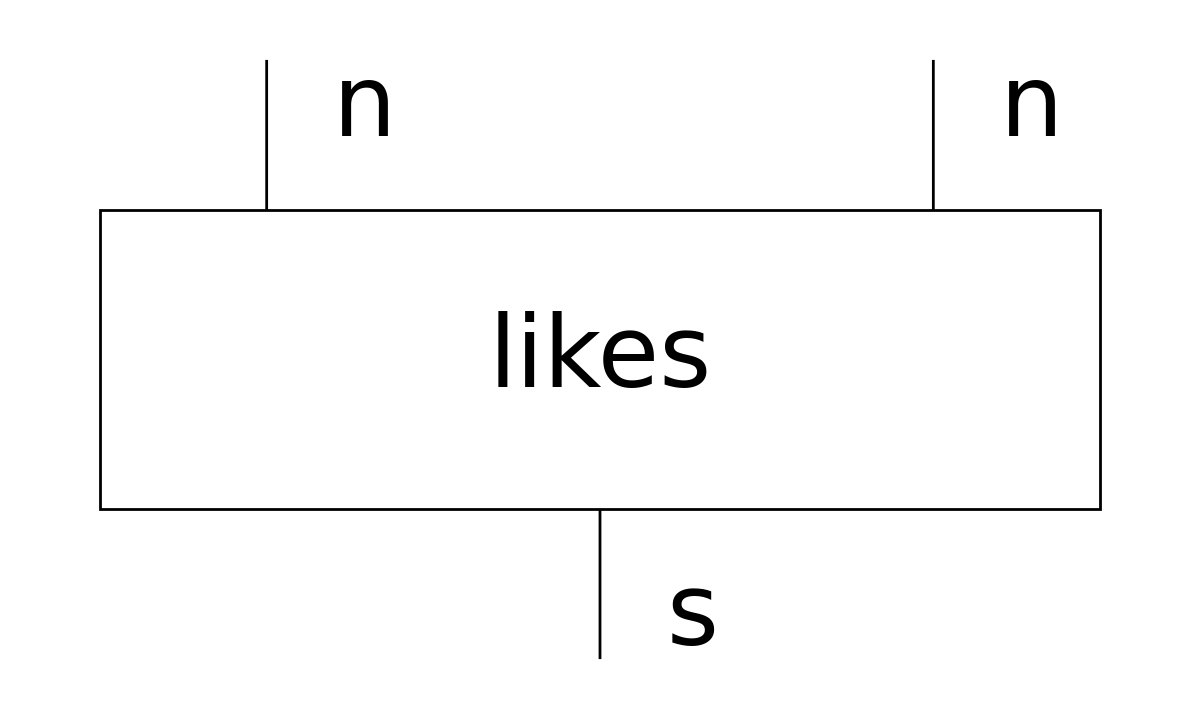} \hspace{5em}
	\includegraphics[width=0.25\textwidth]{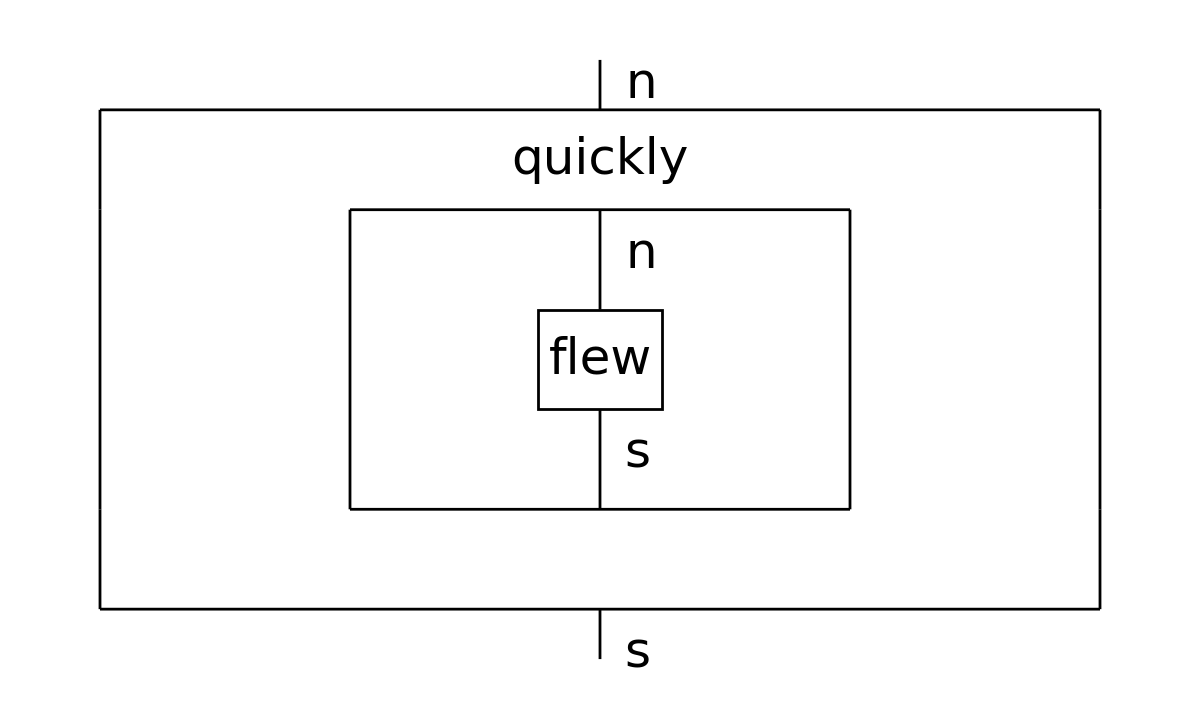}\\
	\rule{10cm}{0.8pt}\\
	\includegraphics[width=0.2\textwidth]{paper-figures/pipeline-output/Section3/parallel_composition1.png}
	\includegraphics[width=0.25\textwidth]{paper-figures/pipeline-output/Section3/parallel_composition2.png}\\
\end{center}
\end{minipage}

\subsection{Example}
Consider the sentence \texttt{Alice dislikes Bob who likes Claire}, which when parsed with CCG yields the $\lambda$-term (depicted as a tree)
\[\begin{tikzpicture}
	\begin{pgfonlayer}{nodelayer}
		\node [style=none] (0) at (0, 2) {\texttt{likes}};
		\node [style=none] (1) at (2, 2) {\texttt{Claire}};
		\node [style=none] (2) at (0, 1.625) {$\scriptstyle n\to n\to s$};
		\node [style=none] (3) at (2, 1.625) {$\scriptstyle n$};
		\node [style=none] (4) at (0.25, 1.375) {};
		\node [style=none] (5) at (1.75, 1.375) {};
		\node [style=none] (6) at (0.75, 1) {};
		\node [style=none] (7) at (1.25, 1) {};
		\node [style=none] (8) at (1, 0.875) {$@$};
		\node [style=none] (9) at (1, 0.5) {$\scriptstyle n\to s$};
		\node [style=none] (10) at (-1, 0.875) {\texttt{who}};
		\node [style=none] (12) at (-1, 0.5) {$\scriptstyle (n\to s)\to n\to n$};
		\node [style=none] (14) at (-0.75, 0.25) {};
		\node [style=none] (15) at (0.75, 0.25) {};
		\node [style=none] (16) at (-0.25, -0.125) {};
		\node [style=none] (17) at (0.25, -0.125) {};
		\node [style=none] (18) at (0, -0.25) {$@$};
		\node [style=none] (19) at (0, -0.625) {$\scriptstyle n\to n$};
		\node [style=none] (21) at (2, -0.125) {\texttt{Bob}};
		\node [style=none] (22) at (2, -0.5) {$\scriptstyle n$};
		\node [style=none] (25) at (0.25, -0.75) {};
		\node [style=none] (26) at (1.75, -0.75) {};
		\node [style=none] (27) at (0.75, -1.125) {};
		\node [style=none] (28) at (1.25, -1.125) {};
		\node [style=none] (29) at (1, -1.25) {$@$};
		\node [style=none] (30) at (1, -1.625) {$\scriptstyle n$};
		\node [style=none] (34) at (-1, -1.25) {\texttt{dislikes}};
		\node [style=none] (35) at (-1, -1.625) {$\scriptstyle n \to n \to s$};
		\node [style=none] (36) at (-0.75, -1.875) {};
		\node [style=none] (37) at (0.75, -1.875) {};
		\node [style=none] (38) at (-0.25, -2.25) {};
		\node [style=none] (39) at (0.25, -2.25) {};
		\node [style=none] (40) at (0, -2.375) {$@$};
		\node [style=none] (41) at (0, -2.75) {$\scriptstyle n\to s$};
		\node [style=none] (46) at (2, -2.25) {\texttt{Alice}};
		\node [style=none] (47) at (2, -2.625) {$\scriptstyle n$};
		\node [style=none] (48) at (0.25, -2.875) {};
		\node [style=none] (49) at (1.75, -2.875) {};
		\node [style=none] (50) at (0.75, -3.25) {};
		\node [style=none] (51) at (1.25, -3.25) {};
		\node [style=none] (52) at (1, -3.375) {$@$};
		\node [style=none] (53) at (1, -3.75) {$\scriptstyle s$};
	\end{pgfonlayer}
	\begin{pgfonlayer}{edgelayer}
		\draw (4.center) to (6.center);
		\draw (7.center) to (5.center);
		\draw (14.center) to (16.center);
		\draw (17.center) to (15.center);
		\draw (25.center) to (27.center);
		\draw (28.center) to (26.center);
		\draw (36.center) to (38.center);
		\draw (39.center) to (37.center);
		\draw (48.center) to (50.center);
		\draw (51.center) to (49.center);
	\end{pgfonlayer}
\end{tikzpicture}\]
Here, \texttt{Alice}, \texttt{Bob}, and \texttt{Claire} become states.
The components \texttt{likes} and \texttt{dislikes} which are typed as transitive verbs $n\to n\to s$ are drawn as gates. The word \texttt{who} has a higher-order type $(n\to s)\to n\to n$ taking an intransitive verb as an input, and so is represented as a frame.
The circuit depiction for the entire sentence is the following
\begin{center}
	\includegraphics[width=0.7\textwidth]{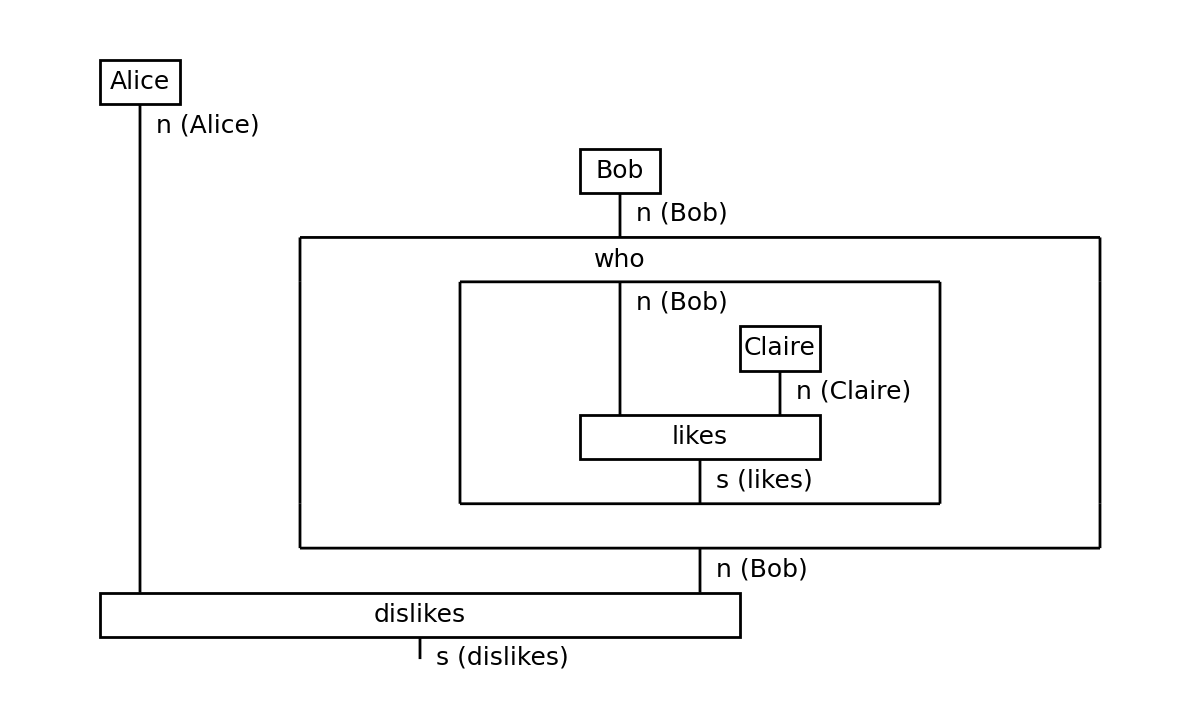}
\end{center}
Here we have labelled the wires with co-indexed types, as discussed in Section~\ref{ssec:coindexing}.
This co-indexing allows us to track which referent each wire in the diagram corresponds to.
For instance, the right-hand wire being fed into the \texttt{dislikes} gate is associated with \texttt{Bob}, reflecting that it is \texttt{Bob} that \texttt{Alice dislikes}, and not \texttt{Claire}.



\section{Rewriting $\lambda$-terms into DisCoCirc form}
\label{sec:rewriting_diags_to_discocirc}

The CCG to $\lambda$-term conversion procedure in Section~\ref{ssec:ccg-to-lambda} combined with the $\lambda$-term to diagram procedure in Section~\ref{sec:lambda-to-diagrams} will suffice to
convert linguistic constituents (up to the level of a sentence) into diagrams that somewhat resemble text circuits.
However, just applying these two steps does not exactly yield the text circuits we desire. 
A number of additional operations need to be performed to bring these diagrams into a text circuit form.
These operations, which we describe in this section, all modify the $\lambda$-term tree.
The order in which we present these operations in this section is conceptually motivated and does not exactly follow the order in which they are applied in the pipeline (see Figure~\ref{fig:pipeline}).

One of the main ideas behind DisCoCirc is that we want to compose circuits corresponding to different sentences along their noun wires to build the circuit of the entire text. 
To enable this, firstly it must be the case that all the noun referents in a diagram are exposed at the top of the diagram, to allow for precomposition. 
This is not always true for diagrams obtained from the CCG parse -- it may be that there are nouns `trapped inside frames'.
In such cases, we apply a procedure called \textit{dragging out} (Section~\ref{ssec:dragging-out})
which changes the order of application of words to ensure that nouns are applied last. 
As the name suggests, this corresponds at the diagram level to dragging the noun out of the frame to the top of the circuit.

To enable the composition of circuits as proposed by DisCoCirc, we further require that the outgoing wires of a diagram are replaced by (i.e. \textit{expanded} into) some number of $n$-type wires, where each wire corresponds to one of the referents involved in the diagram (Section~\ref{ssec:type-expansion}). 
Only then can we postcompose with other circuits.
This notion of \textit{type expansion} was originally introduced in~\cite{coecke2021mathematics} for sentence ($s$) types. 
In this paper, we extend it to also include the expansion of noun phrase ($n$) types. 

These two steps capture the essence of the conversion process from the raw $\lambda$-terms as given by the CCG parse, and the $\lambda$-terms corresponding to DisCoCirc diagrams.
A further technicality we deal with is \textit{noun-coordination expansion}, which arises when we have coordinating conjunctions of noun phrases (Section~\ref{ssec:coord_expansion}).
This is also a new grammatical construction that was not addressed in~\cite{wang2023distilling}.

After these steps are completed, we incorporate semantic information from a coreference parse of the input text, which tells us which nouns refer to the same referents, and how to resolve pronouns (Section~\ref{sec:sentence-composition}).
Concretely, this step leaves the gates of the existing circuits alone, and performs manipulations involving the noun wires.
Firstly, at the level of individual sentences, we use the coreference information to merge together noun wires that refer to the same referent.
Secondly, coreference across sentences tells us how to compose the circuits corresponding to different sentences.

Finally, we may choose to apply \textit{semantic rewrites} (Section~\ref{sec:semantic_rewrites}) which are further rewrites of our $\lambda$-terms and their corresponding diagrams.
These are somewhat optional, and can be tailored to the task or context in which we are using these text circuits.
Many of the rewrite rules we propose here are motivated by rewrites discussed in~\cite{wang2023distilling}.

\subsection{Dragging out}
\label{ssec:dragging-out}

The main part of the dragging out procedure resembles the action of a \textbf{B} combinator~\cite{hindleyseldin2008}.
\textbf{C}~combinators~\cite{hindleyseldin2008} are also used in a structural way to enable flexibility in the order in which terms can accept their inputs.
The \textbf{B} and \textbf{C} combinator together give us a drag\_out procedure (Algorithm~\ref{alg:_drag_out}) that suffices to perform dragging out on any $\lambda$-term that does not involve abstraction. We first give a motivating example of why dragging out is necessary (Section~\ref{sssec:drag_out_examples}) before describing the exact procure (Sections~\ref{ssec:B_combinator},~\ref{sssec:C_combinator}).

Dealing with abstractions is relatively straightforward and requires a structural operation that is conceptually uninteresting, so we relegate its discussion to Appendix~\ref{ssec:beta_expand_appendix}.
In brief, it requires doing a \textit{$\beta$-expansion} pass on the $\lambda$-term before then calling drag\_out.

\subsubsection{Motivating examples}
\label{sssec:drag_out_examples}

We give some simple examples to illustrate the necessity for dragging out and provide intuition for what it entails.
A minimal example is the sentence
\begin{center}
	\texttt{Alice really likes Bob}.
\end{center}
The $\lambda$-term obtained directly from the CCG parse is the following, which corresponds to a diagram in which the noun \texttt{Bob} is inside the higher-order component \texttt{really}.
\begin{center}
	\begin{minipage}{0.49\linewidth}
		\[\begin{tikzpicture}
	\begin{pgfonlayer}{nodelayer}
		\node [style=none, align=center] (0) at (-2, -0.125) {\texttt{really}};
		\node [style=none] (1) at (-0.5, 1) {\texttt{likes}};
		\node [style=none] (2) at (2.5, 1) {\texttt{Bob}};
		\node [style=none] (3) at (3, -1.25) {\texttt{Alice}};
		\node [style=none] (4) at (-0.25, 0.5) {};
		\node [style=none] (5) at (2.25, 0.5) {};
		\node [style=none] (6) at (0.75, 0) {};
		\node [style=none] (7) at (1.25, 0) {};
		\node [style=none, align=center] (8) at (1, -0.125) {$@$};
		\node [style=none] (9) at (0.75, -0.625) {};
		\node [style=none] (10) at (-0.25, -1.125) {};
		\node [style=none] (12) at (-1.75, -0.625) {};
		\node [style=none] (13) at (-0.75, -1.125) {};
		\node [style=none] (14) at (-0.5, -1.25) {$@$};
		\node [style=none] (15) at (1.25, -2.375) {$@$};
		\node [style=none] (16) at (1, -2.25) {};
		\node [style=none] (17) at (-0.25, -1.75) {};
		\node [style=none] (18) at (1.5, -2.25) {};
		\node [style=none] (19) at (2.75, -1.75) {};
		\node [style=none] (20) at (-0.5, 0.625) {$\scriptstyle n\to n\to s$};
		\node [style=none] (21) at (2.5, 0.625) {$\scriptstyle n$};
		\node [style=none] (22) at (1, -0.5) {$\scriptstyle n\to s$};
		\node [style=none] (23) at (-2, -0.5) {$\scriptstyle (n\to s)\to n\to s$};
		\node [style=none] (24) at (-0.5, -1.625) {$\scriptstyle n\to s$};
		\node [style=none] (25) at (3, -1.625) {$\scriptstyle n$};
		\node [style=none] (26) at (1.25, -2.75) {$\scriptstyle s$};
	\end{pgfonlayer}
	\begin{pgfonlayer}{edgelayer}
		\draw (4.center) to (6.center);
		\draw (5.center) to (7.center);
		\draw (12.center) to (13.center);
		\draw (9.center) to (10.center);
		\draw (17.center) to (16.center);
		\draw (18.center) to (19.center);
	\end{pgfonlayer}
\end{tikzpicture}\]
	\end{minipage}
	\begin{minipage}{0.49\linewidth}
		\includegraphics[width=\textwidth]{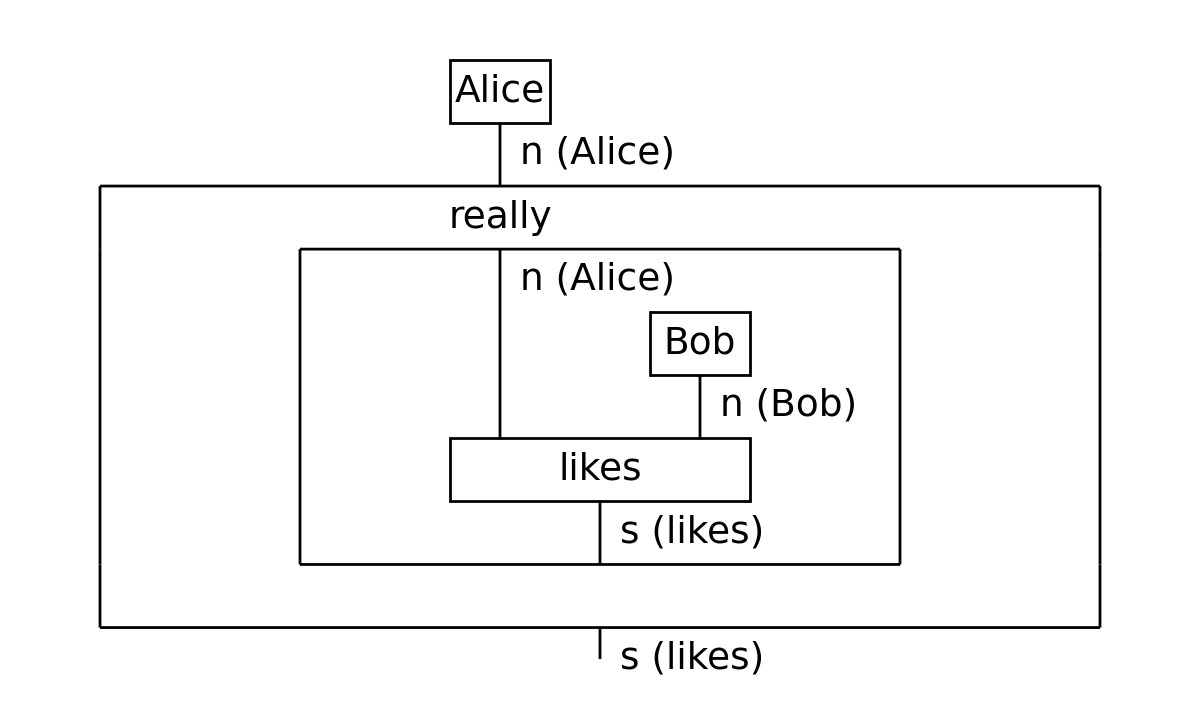}
	\end{minipage}
\end{center}
\texttt{Bob} is inside the higher-order component because, as can be seen from the $\lambda$-tree, it is fed into \texttt{likes} as an argument, and the composite \texttt{likes}(\texttt{Bob}) is in turn fed into \texttt{really} as an argument. 

However, in text circuits, we want all nouns to be at the top of the diagram. Therefore, we want \texttt{likes} first to be fed into \texttt{really} by itself, and then we want \texttt{Bob} to be fed into \texttt{really(likes)} instead. 
We achieve this by performing the dragging out algorithm on the $\lambda$-term.
The resulting $\lambda$-term and accompanying diagram is the following
\begin{center}
	\makebox[\textwidth]{\makebox[1.2\textwidth]{
	\begin{minipage}{0.59\linewidth}
		\[\begin{tikzpicture}
	\begin{pgfonlayer}{nodelayer}
		\node [style=none] (0) at (-1.5, 2) {\texttt{really}};
		\node [style=none] (1) at (1.5, 2) {\texttt{likes}};
		\node [style=none] (2) at (3, 0.875) {\texttt{Bob}};
		\node [style=none] (3) at (5, -0.25) {\texttt{Alice}};
		\node [style=none] (4) at (1.25, 1.5) {};
		\node [style=none] (5) at (2.75, 0.375) {};
		\node [style=none] (6) at (0.25, 1) {};
		\node [style=none] (7) at (1.75, -0.125) {};
		\node [style=none] (8) at (1.5, -0.25) {$@$};
		\node [style=none] (15) at (3.25, -1.375) {$@$};
		\node [style=none] (16) at (3, -1.25) {};
		\node [style=none] (17) at (1.75, -0.75) {};
		\node [style=none] (18) at (3.5, -1.25) {};
		\node [style=none] (19) at (4.75, -0.75) {};
		\node [style=none] (20) at (0, 0.875) {$@$};
		\node [style=none] (21) at (-1.25, 1.5) {};
		\node [style=none] (22) at (-0.25, 1) {};
		\node [style=none] (24) at (1.25, -0.125) {};
		\node [style=none] (25) at (0.25, 0.375) {};
		\node [style=none] (26) at (-1.5, 1.625) {$\scriptstyle (n\to n\to s)\to n\to n\to s$};
		\node [style=none] (27) at (1.5, 1.625) {$\scriptstyle n\to n\to s$};
		\node [style=none] (28) at (0, 0.5) {$\scriptstyle n\to n\to s$};
		\node [style=none] (29) at (3, 0.5) {$\scriptstyle n$};
		\node [style=none] (30) at (1.5, -0.625) {$\scriptstyle n\to s$};
		\node [style=none] (31) at (5, -0.625) {$\scriptstyle n$};
		\node [style=none] (32) at (3.25, -1.75) {$\scriptstyle s$};
	\end{pgfonlayer}
	\begin{pgfonlayer}{edgelayer}
		\draw (4.center) to (6.center);
		\draw (5.center) to (7.center);
		\draw (17.center) to (16.center);
		\draw (18.center) to (19.center);
		\draw (21.center) to (22.center);
		\draw (25.center) to (24.center);
	\end{pgfonlayer}
\end{tikzpicture}\]
	\end{minipage}
	\begin{minipage}{0.59\linewidth}
		\includegraphics[width=\textwidth]{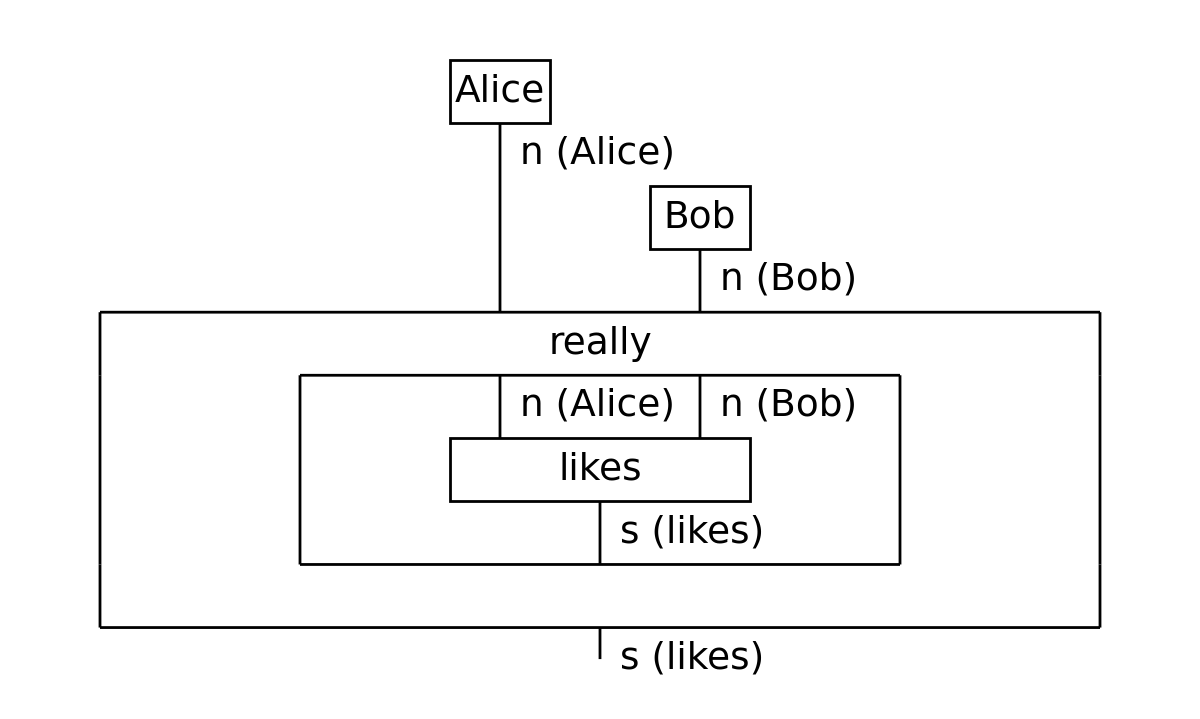}
	\end{minipage}
	}}
\end{center}
Note the structure of the tree has changed, and the type of the frame \texttt{really} has changed, while the types of the other literals have stayed the same.

This dragging out algorithm must work recursively, since we can nest frames inside frames.
The CCG parse for
$$\texttt{Claire knows Alice really likes Bob}$$
yields the following tree 
\begin{center}
	\makebox[\textwidth]{\makebox[1.2\textwidth]{
	\begin{minipage}{0.59\linewidth}
	\[\begin{tikzpicture}
	\begin{pgfonlayer}{nodelayer}
		\node [style=none] (0) at (0, 1) {\texttt{likes}};
		\node [style=none] (1) at (2.5, 1) {\texttt{Bob}};
		\node [style=none] (2) at (1.25, -0.125) {$@$};
		\node [style=none] (3) at (1, 0) {};
		\node [style=none] (4) at (0.25, 0.5) {};
		\node [style=none] (5) at (1.5, 0) {};
		\node [style=none] (6) at (2.25, 0.5) {};
		\node [style=none] (7) at (-1.25, -0.125) {\texttt{really}};
		\node [style=none] (8) at (-1, -0.625) {};
		\node [style=none] (9) at (-0.25, -1.125) {};
		\node [style=none] (10) at (0.25, -1.125) {};
		\node [style=none] (11) at (1, -0.625) {};
		\node [style=none] (12) at (0, -1.25) {$@$};
		\node [style=none] (13) at (2.5, -1.25) {\texttt{Alice}};
		\node [style=none] (14) at (2.25, -1.75) {};
		\node [style=none] (15) at (0.25, -1.75) {};
		\node [style=none] (16) at (1.5, -2.25) {};
		\node [style=none] (17) at (1, -2.25) {};
		\node [style=none] (18) at (1.25, -2.375) {$@$};
		\node [style=none] (19) at (-1.75, -2.375) {\texttt{knows}};
		\node [style=none] (20) at (-0.25, -3.5) {$@$};
		\node [style=none] (21) at (-1.5, -2.875) {};
		\node [style=none] (22) at (-0.5, -3.375) {};
		\node [style=none] (23) at (0, -3.375) {};
		\node [style=none] (24) at (1, -2.875) {};
		\node [style=none] (25) at (3.25, -3.5) {\texttt{Claire}};
		\node [style=none] (26) at (0, -4) {};
		\node [style=none] (27) at (3, -4) {};
		\node [style=none] (28) at (1.5, -4.625) {$@$};
		\node [style=none] (29) at (1.25, -4.5) {};
		\node [style=none] (30) at (1.75, -4.5) {};
		\node [style=none] (31) at (0, 0.625) {$\scriptstyle n\to n\to s$};
		\node [style=none] (32) at (2.5, 0.625) {$\scriptstyle n$};
		\node [style=none] (33) at (-1.25, -0.5) {$\scriptstyle (n\to s)\to n\to s$};
		\node [style=none] (34) at (1.25, -0.5) {$\scriptstyle n\to s$};
		\node [style=none] (35) at (0, -1.625) {$\scriptstyle n\to s$};
		\node [style=none] (36) at (2.5, -1.625) {$\scriptstyle n$};
		\node [style=none] (37) at (-1.75, -2.75) {$\scriptstyle s\to n\to s$};
		\node [style=none] (38) at (1.25, -2.75) {$\scriptstyle s$};
		\node [style=none] (39) at (3.25, -3.875) {$\scriptstyle n$};
		\node [style=none] (40) at (-0.25, -3.875) {$\scriptstyle n\to s$};
		\node [style=none] (41) at (1.5, -5) {$\scriptstyle s$};
	\end{pgfonlayer}
	\begin{pgfonlayer}{edgelayer}
		\draw (4.center) to (3.center);
		\draw (6.center) to (5.center);
		\draw (11.center) to (10.center);
		\draw (9.center) to (8.center);
		\draw (15.center) to (17.center);
		\draw (16.center) to (14.center);
		\draw (23.center) to (24.center);
		\draw (22.center) to (21.center);
		\draw (29.center) to (26.center);
		\draw (30.center) to (27.center);
	\end{pgfonlayer}
\end{tikzpicture}\]
\end{minipage}
\begin{minipage}{0.59\linewidth}
	\includegraphics[width=\textwidth]{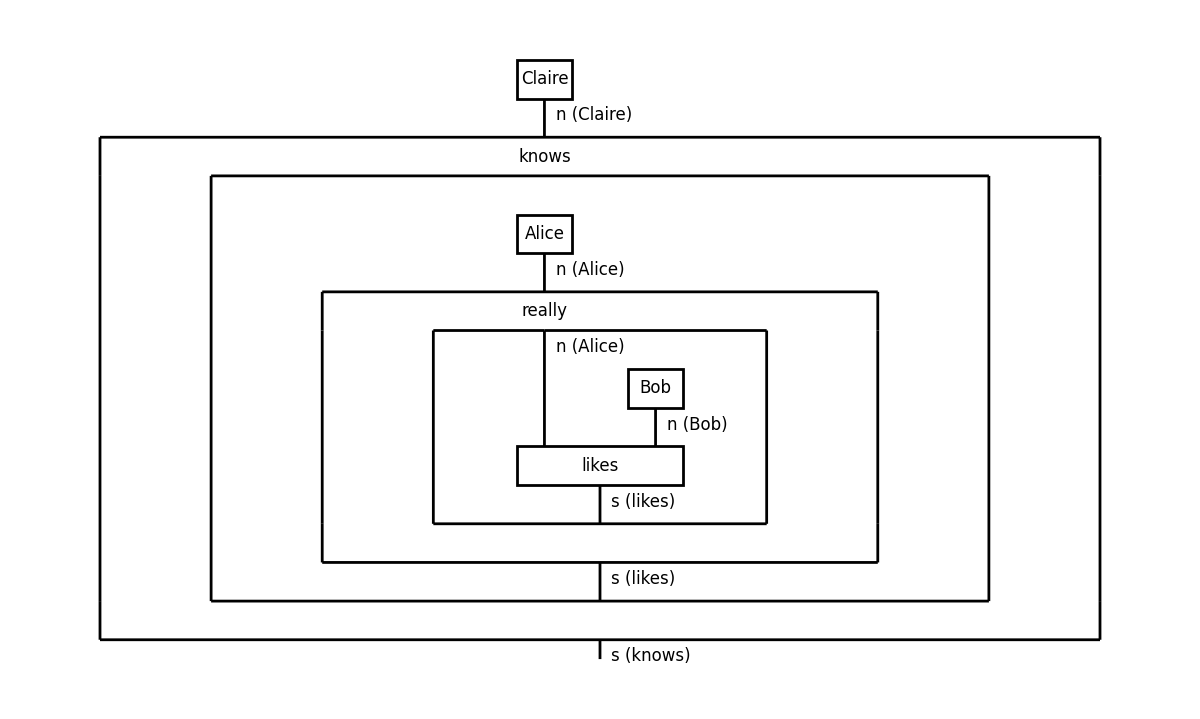}
\end{minipage}
	}}
\end{center}
Note the \texttt{Alice really likes Bob} part of this sentence corresponds to a subtree identical to our earlier sentence.

Here we need to drag \texttt{Bob} out of two frames, one after the other.
We can first drag \texttt{Bob} out of the \texttt{really} frame identically to before, to obtain the intermediate circuit
\begin{center}
	\makebox[\textwidth]{\makebox[1.2\textwidth]{
			\begin{minipage}{0.59\linewidth}
				\[\begin{tikzpicture}
	\begin{pgfonlayer}{nodelayer}
		\node [style=none] (19) at (-2.25, 0.25) {\texttt{knows}};
		\node [style=none] (20) at (-0.775, -0.875) {$@$};
		\node [style=none] (25) at (2.75, -0.875) {\texttt{Claire}};
		\node [style=none] (26) at (-0.525, -1.375) {};
		\node [style=none] (27) at (2.5, -1.375) {};
		\node [style=none] (28) at (1, -2) {$@$};
		\node [style=none] (29) at (0.75, -1.875) {};
		\node [style=none] (30) at (1.25, -1.875) {};
		\node [style=none] (37) at (-1.025, -0.75) {};
		\node [style=none] (38) at (-0.525, -0.75) {};
		\node [style=none] (39) at (-2, -0.25) {};
		\node [style=none] (40) at (0.75, -0.25) {};
		\node [style=none] (41) at (-3.75, 3.625) {\texttt{really}};
		\node [style=none] (42) at (-0.75, 3.625) {\texttt{likes}};
		\node [style=none] (43) at (0.75, 2.5) {\texttt{Bob}};
		\node [style=none] (44) at (2.75, 1.375) {\texttt{Alice}};
		\node [style=none] (45) at (-1, 3.125) {};
		\node [style=none] (46) at (0.5, 2) {};
		\node [style=none] (47) at (-2, 2.625) {};
		\node [style=none] (48) at (-0.5, 1.5) {};
		\node [style=none] (49) at (-0.75, 1.375) {$@$};
		\node [style=none] (50) at (1, 0.25) {$@$};
		\node [style=none] (51) at (0.75, 0.375) {};
		\node [style=none] (52) at (-0.5, 0.875) {};
		\node [style=none] (53) at (1.25, 0.375) {};
		\node [style=none] (54) at (2.5, 0.875) {};
		\node [style=none] (55) at (-2.25, 2.5) {$@$};
		\node [style=none] (56) at (-3.5, 3.125) {};
		\node [style=none] (57) at (-2.5, 2.625) {};
		\node [style=none] (58) at (-1, 1.5) {};
		\node [style=none] (59) at (-2, 2) {};
		\node [style=none] (60) at (-3.75, 3.25) {$\scriptstyle (n\to n\to s)\to n\to n\to s$};
		\node [style=none] (61) at (-0.75, 3.25) {$\scriptstyle n\to n\to s$};
		\node [style=none] (62) at (-2.25, 2.125) {$\scriptstyle n\to n\to s$};
		\node [style=none] (63) at (0.75, 2.125) {$\scriptstyle n$};
		\node [style=none] (64) at (-0.75, 1) {$\scriptstyle n\to s$};
		\node [style=none] (65) at (2.75, 1) {$\scriptstyle n$};
		\node [style=none] (66) at (1, -0.125) {$\scriptstyle s$};
		\node [style=none] (67) at (-2.25, -0.125) {$\scriptstyle s\to n\to s$};
		\node [style=none] (68) at (-0.775, -1.25) {$\scriptstyle n\to s$};
		\node [style=none] (69) at (2.75, -1.25) {$\scriptstyle n$};
		\node [style=none] (70) at (1, -2.375) {$\scriptstyle s$};
	\end{pgfonlayer}
	\begin{pgfonlayer}{edgelayer}
		\draw (29.center) to (26.center);
		\draw (30.center) to (27.center);
		\draw (37.center) to (39.center);
		\draw (38.center) to (40.center);
		\draw (45.center) to (47.center);
		\draw (46.center) to (48.center);
		\draw (52.center) to (51.center);
		\draw (53.center) to (54.center);
		\draw (56.center) to (57.center);
		\draw (59.center) to (58.center);
	\end{pgfonlayer}
\end{tikzpicture}\]
			\end{minipage}
			\begin{minipage}{0.59\linewidth}
				\includegraphics[width=	\textwidth]{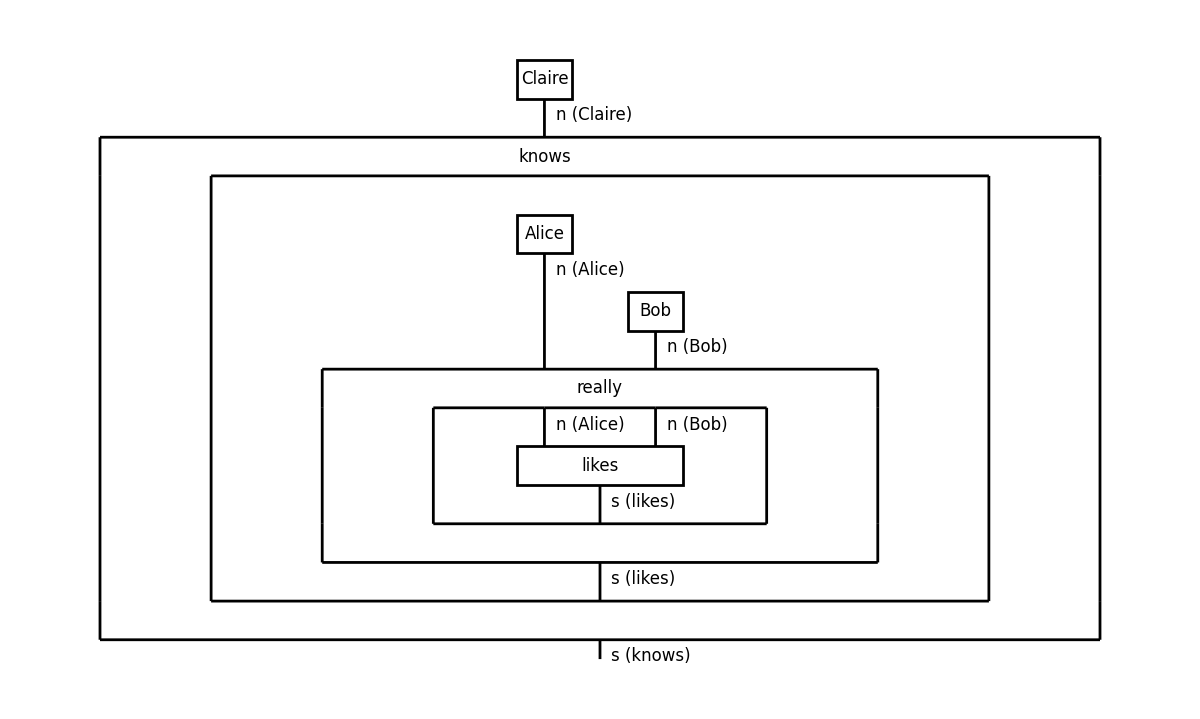}
			\end{minipage}
	}}
\end{center}
Then, we drag \texttt{Alice} and \texttt{Bob} out of \texttt{knows} to obtain the fully dragged-out diagram
\begin{center}
	\makebox[\textwidth]{\makebox[1.2\textwidth]{
			\begin{minipage}{0.59\linewidth}
				\[\begin{tikzpicture}
	\begin{pgfonlayer}{nodelayer}
		\node [style=none] (0) at (0, 5) {\texttt{likes}};
		\node [style=none] (1) at (-0.375, 2.875) {\texttt{Bob}};
		\node [style=none] (2) at (-2, 1.75) {$@$};
		\node [style=none] (3) at (-2.25, 1.875) {};
		\node [style=none] (4) at (-3.25, 2.375) {};
		\node [style=none] (5) at (-1.75, 1.875) {};
		\node [style=none] (6) at (-0.625, 2.375) {};
		\node [style=none] (7) at (-3, 5) {\texttt{really}};
		\node [style=none] (8) at (-1.75, 1.25) {};
		\node [style=none] (9) at (-1, 0.75) {};
		\node [style=none] (10) at (-0.5, 0.75) {};
		\node [style=none] (11) at (0.5, 1.25) {};
		\node [style=none] (12) at (-0.75, 0.625) {$@$};
		\node [style=none] (13) at (0.75, 1.75) {\texttt{Alice}};
		\node [style=none] (19) at (-5.5, 4) {\texttt{knows}};
		\node [style=none] (25) at (1.75, 0.625) {\texttt{Claire}};
		\node [style=none] (28) at (0.5, -0.5) {$@$};
		\node [style=none] (32) at (-1.75, 4.125) {};
		\node [style=none] (33) at (-1.25, 4.125) {};
		\node [style=none] (34) at (-2.75, 4.625) {};
		\node [style=none] (35) at (-0.25, 4.625) {};
		\node [style=none] (36) at (-1.5, 4) {$@$};
		\node [style=none] (41) at (-1.75, 3.5) {};
		\node [style=none] (42) at (-5.25, 3.5) {};
		\node [style=none] (43) at (-3.5, 2.875) {$@$};
		\node [style=none] (44) at (-3.75, 3) {};
		\node [style=none] (45) at (-3.25, 3) {};
		\node [style=none] (46) at (0.75, -0.375) {};
		\node [style=none] (47) at (0.25, -0.375) {};
		\node [style=none] (48) at (-0.5, 0.125) {};
		\node [style=none] (49) at (1.5, 0.125) {};
		\node [style=none] (50) at (-3, 4.75) {$\scriptstyle (n\to n\to s)\to n\to n\to s$};
		\node [style=none] (51) at (0, 4.75) {$\scriptstyle n\to n\to s$};
		\node [style=none] (52) at (-5.5, 3.625) {$\scriptstyle (n\to n\to s)\to n\to n\to n\to s$};
		\node [style=none] (53) at (-1.5, 3.625) {$\scriptstyle n\to n\to s$};
		\node [style=none] (54) at (-3.5, 2.5) {$\scriptstyle n\to n\to n\to s$};
		\node [style=none] (55) at (-0.375, 2.5) {$\scriptstyle n$};
		\node [style=none] (56) at (-2, 1.375) {$\scriptstyle n\to n\to s$};
		\node [style=none] (57) at (0.75, 1.375) {$\scriptstyle n$};
		\node [style=none] (58) at (-0.75, 0.25) {$\scriptstyle n\to s$};
		\node [style=none] (59) at (1.75, 0.25) {$\scriptstyle n$};
		\node [style=none] (60) at (0.5, -0.875) {$\scriptstyle s$};
	\end{pgfonlayer}
	\begin{pgfonlayer}{edgelayer}
		\draw (4.center) to (3.center);
		\draw (6.center) to (5.center);
		\draw (11.center) to (10.center);
		\draw (9.center) to (8.center);
		\draw (34.center) to (32.center);
		\draw (33.center) to (35.center);
		\draw (42.center) to (44.center);
		\draw (45.center) to (41.center);
		\draw (48.center) to (47.center);
		\draw (46.center) to (49.center);
	\end{pgfonlayer}
\end{tikzpicture}\]
			\end{minipage}
			\begin{minipage}{0.59\linewidth}
				\includegraphics[width=\textwidth]{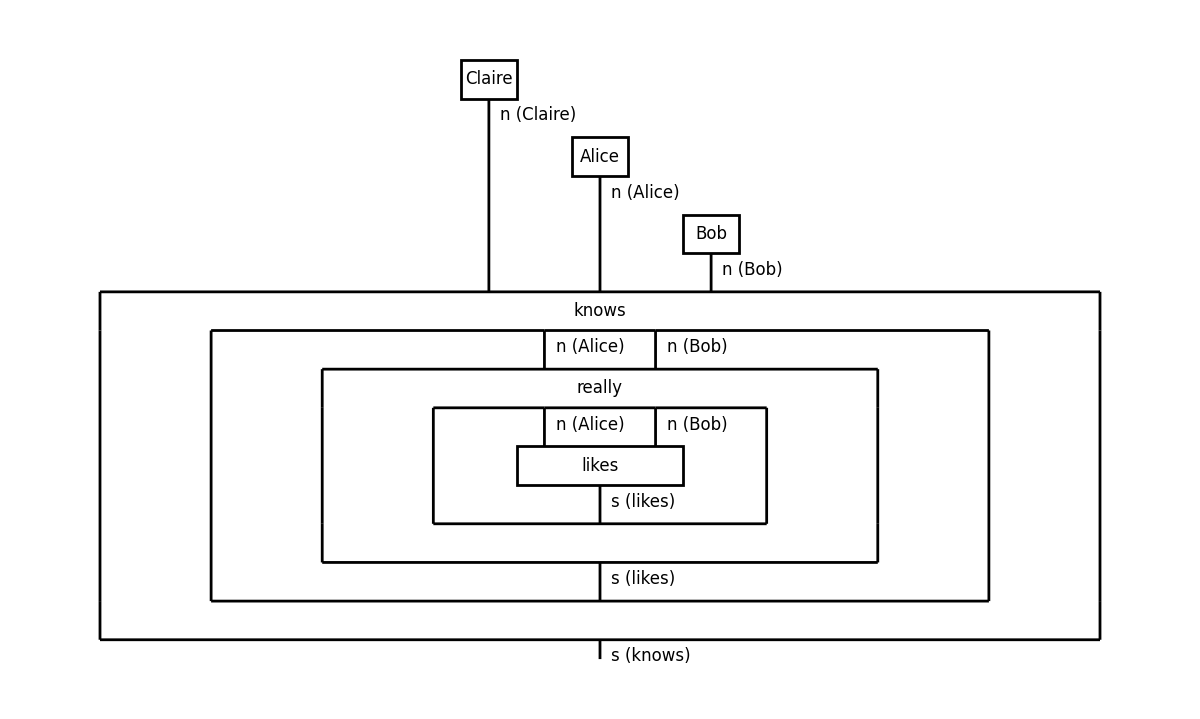}
			\end{minipage}
	}}
\end{center}
Indeed, natural language can yield arbitrarily deep nestings like this -- e.g. \texttt{Fred dreamt Eve thought Dave said \ldots\ Alice really likes Bob}.

\subsubsection{B combinator}\label{ssec:B_combinator}
Recall the first example from the previous section (Section~\ref{sssec:drag_out_examples}) -- \texttt{Alice really likes Bob}.
In this case, the dragging out procedure is accomplished by the introduction of a $\mathbf{B}$ combinator
\begin{align*}
	\mathbf{B} = \lambda f.\lambda g.\lambda h. f(gh)
	\\
	(B\to C)\to (A\to B)\to A \to C
\end{align*}
Applied to some function $f: B \to C$, the $\mathbf{B}$ combinator changes $f$ such that it first accepts a function $g: A \to B$ and then a value $h: A$ for some type $A$. 
Thus $\mathbf{B}$ can be thought of as enabling the composition of functions: $\mathbf{B} f g$ is the composite $f\circ g$.

In the case of \texttt{Alice really likes Bob}, applying the $\mathbf{B}$ combinator to \texttt{really} converts the higher-order component of type $(n \to s) \to n \to s$ to the component $\mathbf{B}$ \texttt{really}: $(n \to n \to s) \to n \to n \to s$. 
This, in turn, allows us first to compose \texttt{really} and \texttt{like} before applying the two nouns \texttt{Bob} and \texttt{Alice} respectively.
Effectively, \texttt{Bob} has been dragged outside of the frame \texttt{really}.
\[\begin{tikzpicture}
	\begin{pgfonlayer}{nodelayer}
		\node [style=none] (0) at (-2.5, -0.375) {\texttt{really}};
		\node [style=none] (1) at (-0.75, 0.875) {\texttt{likes}};
		\node [style=none] (2) at (1.75, 0.875) {$\texttt{Bob}_{n}$};
		\node [style=none] (3) at (0.5, -0.375) {$@$};
		\node [style=none] (4) at (-0.5, 0.25) {};
		\node [style=none] (5) at (0.25, -0.25) {};
		\node [style=none] (6) at (0.75, -0.25) {};
		\node [style=none] (7) at (1.5, 0.25) {};
		\node [style=none] (8) at (-1, -1.5) {$@$};
		\node [style=none] (9) at (-0.75, -1.375) {};
		\node [style=none] (10) at (0.25, -0.875) {};
		\node [style=none] (11) at (-1.25, -1.375) {};
		\node [style=none] (12) at (-2.25, -0.875) {};
		\node [style=none] (13) at (2, -1.5) {\texttt{Alice}};
		\node [style=none] (14) at (0.5, -2.625) {$@$};
		\node [style=none] (15) at (0.25, -2.5) {};
		\node [style=none] (16) at (-0.75, -2) {};
		\node [style=none] (17) at (0.75, -2.5) {};
		\node [style=none] (18) at (1.75, -2) {};
		\node [style=none] (19) at (2.5, 1) {};
		\node [style=none] (20) at (-3.75, -0.25) {};
		\node [style=none] (21) at (-1, -2.25) {};
		\node [style=none] (22) at (4, -0.5) {$=_{\beta}$};
		\node [style=none] (23) at (6.75, 1) {\textbf{B}\texttt{really}};
		\node [style=none] (24) at (9.75, 1) {\texttt{likes}};
		\node [style=none] (25) at (10.75, -0.125) {\texttt{Bob}};
		\node [style=none] (26) at (9.5, -1.375) {$@$};
		\node [style=none] (29) at (9.75, -1.25) {};
		\node [style=none] (30) at (10.5, -0.75) {};
		\node [style=none] (31) at (8.25, -0.125) {$@$};
		\node [style=none] (32) at (8.5, 0) {};
		\node [style=none] (33) at (9.5, 0.5) {};
		\node [style=none] (34) at (8, 0) {};
		\node [style=none] (35) at (7, 0.5) {};
		\node [style=none] (36) at (12, -1.375) {\texttt{Alice}};
		\node [style=none] (37) at (10.75, -2.5) {$@$};
		\node [style=none] (38) at (9.25, -1.25) {};
		\node [style=none] (39) at (8.5, -0.75) {};
		\node [style=none] (40) at (11, -2.375) {};
		\node [style=none] (41) at (11.75, -1.875) {};
		\node [style=none] (42) at (9.75, -1.875) {};
		\node [style=none] (43) at (10.5, -2.375) {};
		\node [style=none] (44) at (5.25, 1) {};
		\node [style=none] (45) at (11.75, 0.5) {};
		\node [style=none] (46) at (9.25, -2) {};
		\node [style=none] (47) at (-0.75, 0.5) {$\scriptstyle n\to n\to s$};
		\node [style=none] (48) at (1.75, 0.5) {$\scriptstyle n$};
		\node [style=none] (49) at (0.5, -0.75) {$\scriptstyle n\to s$};
		\node [style=none] (50) at (-2.5, -0.75) {$\scriptstyle (n\to s)\to n\to s$};
		\node [style=none] (51) at (-1, -1.75) {$\scriptstyle n\to s$};
		\node [style=none] (52) at (0.5, -3) {$\scriptstyle s$};
		\node [style=none] (53) at (2, -1.875) {$\scriptstyle n$};
		\node [style=none] (54) at (6.75, 0.625) {$\scriptstyle (n\to n\to s)\to n\to n\to s$};
		\node [style=none] (55) at (9.75, 0.625) {$\scriptstyle n\to n\to s$};
		\node [style=none] (56) at (8.25, -0.5) {$\scriptstyle n\to n\to s$};
		\node [style=none] (57) at (10.75, -0.5) {$\scriptstyle n$};
		\node [style=none] (58) at (9.5, -1.75) {$\scriptstyle n\to s$};
		\node [style=none] (59) at (12, -1.75) {$\scriptstyle n$};
		\node [style=none] (60) at (10.75, -2.875) {$\scriptstyle s$};
	\end{pgfonlayer}
	\begin{pgfonlayer}{edgelayer}
		\draw (4.center) to (5.center);
		\draw (7.center) to (6.center);
		\draw (10.center) to (9.center);
		\draw (12.center) to (11.center);
		\draw (16.center) to (15.center);
		\draw (17.center) to (18.center);
		\draw [style=H] (20.center)
			 to [in=-180, out=-90, looseness=0.75] (21.center)
			 to [in=-90, out=0, looseness=0.75] (19.center)
			 to [in=90, out=90, looseness=0.50] cycle;
		\draw (30.center) to (29.center);
		\draw (33.center) to (32.center);
		\draw (35.center) to (34.center);
		\draw (39.center) to (38.center);
		\draw (40.center) to (41.center);
		\draw (42.center) to (43.center);
		\draw [style=H] (44.center)
			 to [in=90, out=60, looseness=0.50] (45.center)
			 to [in=-15, out=-90] (46.center)
			 to [in=-120, out=165, looseness=0.75] cycle;
	\end{pgfonlayer}
\end{tikzpicture}\]
The subtree that is modified is circled.
When we apply the \textbf{B}-combinator to drag something out of a frame, we always look to operate on a subtree of this form.
More precisely, the conditions that must be satisfied in order to trigger the application of a \textbf{B} combinator are: 
\begin{itemize}
	\item the subterm being acted on must be an application: 
	in this example, the subterm is \texttt{really}(\texttt{likes}(\texttt{Bob})), which is an application of \texttt{really} to \texttt{likes}(\texttt{Bob})
	\item the function of the subterm must be a higher-order component (i.e. a frame), so that there is something to be dragged out of: 
	here the function is the constant term \texttt{really} which has the higher-order type $(n\to s)\to n\to s$
	\item the argument term of the subterm is also an instance of application, so that there is something to drag out: 
	the argument term here is \texttt{likes}(\texttt{Bob}), which is itself an application
	\item the argument of the argument (i.e. the thing to be dragged out) is of noun type (or possibly a product of noun types): 
	the argument of the argument here is \texttt{Bob}, which is $n$-type
\end{itemize}

Note the two $\lambda$-terms represented by the trees in the figure above are $\beta$-equivalent.
In practice however, we break this $\beta$-equivalence when doing dragging out.
Instead of keeping the label $\mathbf{B}\texttt{really}$, we forget the presence of $\mathbf{B}$ and just turn this into a constant labelled \texttt{really}.
This gives the resulting dragged-out $\lambda$-term and diagram as shown in Section~\ref{sssec:drag_out_examples}.

Note also that this procedure has to be compatible with co-indexing.
That is, the new $n$-types introduced to the type of \texttt{really} must inherit the co-index of the thing getting dragged out.
In this example, the co-indexed type of \texttt{really} changed from
$$(n\; (\texttt{Alice})\to s\;(\texttt{likes}))\to n\;(\texttt{Alice})\to s\;(\texttt{likes})$$
to
$$(n\;(\texttt{Bob}) \to n\;(\texttt{Alice})\to s\;(\texttt{likes}))\to n\;(\texttt{Bob})\to n\;(\texttt{Alice})\to s\;(\texttt{likes})$$

\subsubsection{C combinator}
\label{sssec:C_combinator}

The dragging out operation via the $\mathbf{B}$ combinator described in the previous section can only be applied to the last input of a term. 
To drag out other inputs in more general examples, we have to pair the $\mathbf{B}$ combinator with $\mathbf{C}$ combinators which interchange the order of arguments. 
\begin{align*}
	\mathbf{C} = \lambda x.\lambda y.\lambda z. xzy
	\\
	(A\to B\to C)\to B\to A\to C
\end{align*}
Given a function $f:A\to B\to C$, the function $\mathbf{C}f:B\to A\to C$ accepts input arguments in the reverse order.

To see the necessity of the \textbf{C} combinator, consider the sentence
$$\texttt{Alice quickly runs to Bob}.$$
\begin{figure*}[h]
	\centering
	\makebox[\textwidth]{\makebox[1.2\textwidth]{
			\begin{minipage}{0.59\linewidth}
				\[\begin{tikzpicture}
	\begin{pgfonlayer}{nodelayer}
		\node [style=none] (0) at (0, 1) {\texttt{to}};
		\node [style=none] (1) at (1.375, -0.125) {$@$};
		\node [style=none] (2) at (2.75, 1) {\texttt{Bob}};
		\node [style=none] (3) at (0.25, 0.5) {};
		\node [style=none] (4) at (1, 0) {};
		\node [style=none] (5) at (1.75, 0) {};
		\node [style=none] (6) at (2.5, 0.5) {};
		\node [style=none] (7) at (4.5, -0.125) {\texttt{runs}};
		\node [style=none] (8) at (4.25, -0.75) {};
		\node [style=none] (9) at (1.75, -0.75) {};
		\node [style=none] (10) at (3, -1.375) {$@$};
		\node [style=none] (11) at (2.5, -1.25) {};
		\node [style=none] (12) at (3.5, -1.25) {};
		\node [style=none] (13) at (0, -1.375) {\texttt{quickly}};
		\node [style=none] (14) at (1.375, -2.625) {$@$};
		\node [style=none] (15) at (1, -2.5) {};
		\node [style=none] (16) at (0.25, -2) {};
		\node [style=none] (17) at (1.75, -2.5) {};
		\node [style=none] (18) at (2.5, -2) {};
		\node [style=none] (19) at (4, -2.625) {\texttt{Alice}};
		\node [style=none] (20) at (2.75, -3.875) {$@$};
		\node [style=none] (21) at (1.75, -3.25) {};
		\node [style=none] (22) at (2.5, -3.75) {};
		\node [style=none] (23) at (3, -3.75) {};
		\node [style=none] (24) at (3.75, -3.25) {};
		\node [style=none] (25) at (0, 0.625) {$\scriptstyle n\to (n\to s)\to n\to s$};
		\node [style=none] (26) at (2.75, 0.625) {$\scriptstyle n$};
		\node [style=none] (27) at (1.375, -0.5) {$\scriptstyle (n\to s)\to n\to s$};
		\node [style=none] (28) at (4.5, -0.5) {$\scriptstyle n\to s$};
		\node [style=none] (29) at (3, -1.75) {$\scriptstyle n\to s$};
		\node [style=none] (30) at (0, -1.75) {$\scriptstyle (n\to s)\to n\to s$};
		\node [style=none] (31) at (1.375, -3) {$\scriptstyle n\to s$};
		\node [style=none] (32) at (4, -3) {$\scriptstyle n$};
		\node [style=none] (33) at (2.75, -4.25) {$\scriptstyle s$};
	\end{pgfonlayer}
	\begin{pgfonlayer}{edgelayer}
		\draw (3.center) to (4.center);
		\draw (6.center) to (5.center);
		\draw (11.center) to (9.center);
		\draw (12.center) to (8.center);
		\draw (16.center) to (15.center);
		\draw (18.center) to (17.center);
		\draw (21.center) to (22.center);
		\draw (24.center) to (23.center);
	\end{pgfonlayer}
\end{tikzpicture}\]
			\end{minipage}
			\begin{minipage}{0.59\linewidth}
				\includegraphics[width=\textwidth]{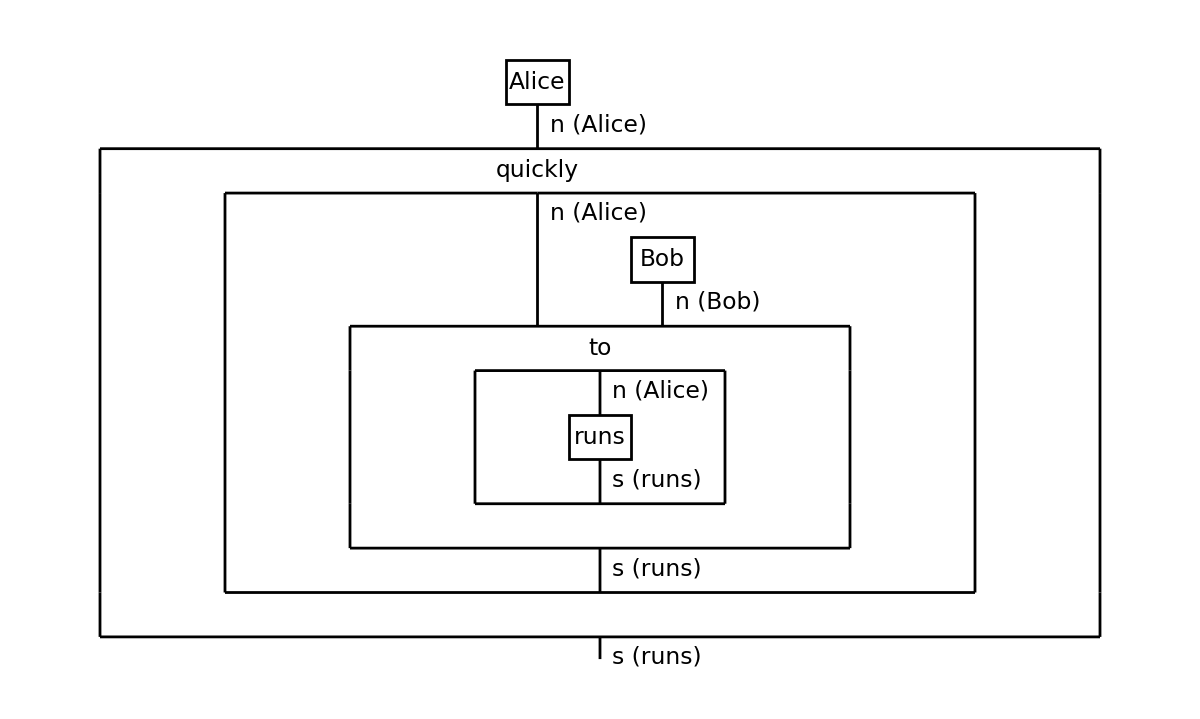}
			\end{minipage}
	}}
\end{figure*}
Here, \texttt{Bob} is a noun that must be dragged out of the frame \texttt{quickly}, but \texttt{runs} is an intransitive verb which does not need to be dragged out.
So, before we can apply our \textbf{B} combinator, it is necessary to first interchange the order of \texttt{Bob} and \texttt{runs}, which is done by applying the \textbf{C} combinator to \texttt{to}.
The subtree that is actually modified by the introduction of the combinator is circled below.
\[\begin{tikzpicture}
	\begin{pgfonlayer}{nodelayer}
		\node [style=none] (19) at (-1.25, 1.25) {};
		\node [style=none] (20) at (-7.5, 2.25) {};
		\node [style=none] (21) at (-3.25, -0.75) {};
		\node [style=none] (22) at (0, 0) {$=_{\beta}$};
		\node [style=none] (59) at (-6.25, 2.25) {\texttt{to}};
		\node [style=none] (60) at (-4.875, 1.125) {$@$};
		\node [style=none] (61) at (-3.5, 2.25) {\texttt{Bob}};
		\node [style=none] (62) at (-6, 1.75) {};
		\node [style=none] (63) at (-5.25, 1.25) {};
		\node [style=none] (64) at (-4.5, 1.25) {};
		\node [style=none] (65) at (-3.75, 1.75) {};
		\node [style=none] (66) at (-1.75, 1.125) {\texttt{runs}};
		\node [style=none] (67) at (-2, 0.5) {};
		\node [style=none] (68) at (-4.5, 0.5) {};
		\node [style=none] (69) at (-3.25, -0.125) {$@$};
		\node [style=none] (70) at (-3.75, 0) {};
		\node [style=none] (71) at (-2.75, 0) {};
		\node [style=none] (72) at (-6.25, -0.125) {\texttt{quickly}};
		\node [style=none] (73) at (-4.875, -1.375) {$@$};
		\node [style=none] (74) at (-5.25, -1.25) {};
		\node [style=none] (75) at (-6, -0.75) {};
		\node [style=none] (76) at (-4.5, -1.25) {};
		\node [style=none] (77) at (-3.75, -0.75) {};
		\node [style=none] (78) at (-2.25, -1.375) {\texttt{Alice}};
		\node [style=none] (79) at (-3.5, -2.625) {$@$};
		\node [style=none] (80) at (-4.5, -2) {};
		\node [style=none] (81) at (-3.75, -2.5) {};
		\node [style=none] (82) at (-3.25, -2.5) {};
		\node [style=none] (83) at (-2.5, -2) {};
		\node [style=none] (84) at (-6.25, 1.875) {$\scriptstyle n\to (n\to s)\to n\to s$};
		\node [style=none] (85) at (-3.5, 1.875) {$\scriptstyle n$};
		\node [style=none] (86) at (-4.875, 0.75) {$\scriptstyle (n\to s)\to n\to s$};
		\node [style=none] (87) at (-1.75, 0.75) {$\scriptstyle n\to s$};
		\node [style=none] (88) at (-3.25, -0.5) {$\scriptstyle n\to s$};
		\node [style=none] (89) at (-6.25, -0.5) {$\scriptstyle (n\to s)\to n\to s$};
		\node [style=none] (90) at (-4.875, -1.75) {$\scriptstyle n\to s$};
		\node [style=none] (91) at (-2.25, -1.75) {$\scriptstyle n$};
		\node [style=none] (92) at (-3.5, -3) {$\scriptstyle s$};
		\node [style=none] (94) at (2.5, 2.25) {\textbf{C}\texttt{to}};
		\node [style=none] (95) at (3.875, 1.125) {$@$};
		\node [style=none] (96) at (5.25, 2.25) {\texttt{runs}};
		\node [style=none] (97) at (2.75, 1.75) {};
		\node [style=none] (98) at (3.5, 1.25) {};
		\node [style=none] (99) at (4.25, 1.25) {};
		\node [style=none] (100) at (5, 1.75) {};
		\node [style=none] (101) at (7, 1.125) {\texttt{Bob}};
		\node [style=none] (102) at (6.75, 0.5) {};
		\node [style=none] (103) at (4.25, 0.5) {};
		\node [style=none] (104) at (5.5, -0.125) {$@$};
		\node [style=none] (105) at (5, 0) {};
		\node [style=none] (106) at (6, 0) {};
		\node [style=none] (107) at (2.5, -0.125) {\texttt{quickly}};
		\node [style=none] (108) at (3.875, -1.375) {$@$};
		\node [style=none] (109) at (3.5, -1.25) {};
		\node [style=none] (110) at (2.75, -0.75) {};
		\node [style=none] (111) at (4.25, -1.25) {};
		\node [style=none] (112) at (5, -0.75) {};
		\node [style=none] (113) at (6.5, -1.375) {\texttt{Alice}};
		\node [style=none] (114) at (5.25, -2.625) {$@$};
		\node [style=none] (115) at (4.25, -2) {};
		\node [style=none] (116) at (5, -2.5) {};
		\node [style=none] (117) at (5.5, -2.5) {};
		\node [style=none] (118) at (6.25, -2) {};
		\node [style=none] (119) at (2.5, 1.875) {$\scriptstyle (n\to s)\to n\to n\to s$};
		\node [style=none] (120) at (5.25, 1.875) {$\scriptstyle n\to s$};
		\node [style=none] (121) at (3.875, 0.75) {$\scriptstyle n \to n \to s$};
		\node [style=none] (122) at (7, 0.75) {$\scriptstyle n$};
		\node [style=none] (123) at (5.5, -0.5) {$\scriptstyle n\to s$};
		\node [style=none] (124) at (2.5, -0.5) {$\scriptstyle (n\to s)\to n\to s$};
		\node [style=none] (125) at (3.875, -1.75) {$\scriptstyle n\to s$};
		\node [style=none] (126) at (6.5, -1.75) {$\scriptstyle n$};
		\node [style=none] (127) at (5.25, -3) {$\scriptstyle s$};
		\node [style=none] (128) at (7.5, 1.25) {};
		\node [style=none] (129) at (1.25, 2.25) {};
		\node [style=none] (130) at (5.5, -0.75) {};
	\end{pgfonlayer}
	\begin{pgfonlayer}{edgelayer}
		\draw [style=H] (20.center)
			 to [in=165, out=-90, looseness=0.75] (21.center)
			 to [in=-75, out=-15] (19.center)
			 to [in=90, out=105, looseness=0.75] cycle;
		\draw (62.center) to (63.center);
		\draw (65.center) to (64.center);
		\draw (70.center) to (68.center);
		\draw (71.center) to (67.center);
		\draw (75.center) to (74.center);
		\draw (77.center) to (76.center);
		\draw (80.center) to (81.center);
		\draw (83.center) to (82.center);
		\draw (97.center) to (98.center);
		\draw (100.center) to (99.center);
		\draw (105.center) to (103.center);
		\draw (106.center) to (102.center);
		\draw (110.center) to (109.center);
		\draw (112.center) to (111.center);
		\draw (115.center) to (116.center);
		\draw (118.center) to (117.center);
		\draw [style=H, in=165, out=-90, looseness=0.75] (129.center) to (130.center);
		\draw [style=H, in=-75, out=-15] (130.center) to (128.center);
		\draw [style=H, in=90, out=105, looseness=0.75] (128.center) to (129.center);
	\end{pgfonlayer}
\end{tikzpicture}\]
Note that, just like when we apply the \textbf{B} combinator, in practice we break the $\beta$-equivalence and forget the presence of the \textbf{C} combinator.
That is, in the term on the right we would relabel the component \textbf{C} \texttt{to} to \texttt{to}.
Note also that even after we do this, the resulting $\lambda$-term will look identical to the initial $\lambda$-term when drawn as diagrams as per Section~\ref{sec:lambda-to-diagrams}.
This is because the two arguments that had their order swapped by the \textbf{C} combinator have different types ($n$ and $n\to s$), so this ordering change will not show up in the diagram. 

After this step, we can then apply the \textbf{B} combinator as before to drag \texttt{Bob} out of \texttt{quickly} and arrive at the fully dragged-out term.

\[\begin{tikzpicture}
	\begin{pgfonlayer}{nodelayer}
		\node [style=none] (3) at (8, 0) {$=_{\beta}$};
		\node [style=none] (75) at (1.75, 2.25) {\textbf{C}\texttt{to}};
		\node [style=none] (76) at (3.125, 1.125) {$@$};
		\node [style=none] (77) at (4.5, 2.25) {\texttt{runs}};
		\node [style=none] (78) at (2, 1.75) {};
		\node [style=none] (79) at (2.75, 1.25) {};
		\node [style=none] (80) at (3.5, 1.25) {};
		\node [style=none] (81) at (4.25, 1.75) {};
		\node [style=none] (82) at (6.25, 1.125) {\texttt{Bob}};
		\node [style=none] (83) at (6, 0.5) {};
		\node [style=none] (84) at (3.5, 0.5) {};
		\node [style=none] (85) at (4.75, -0.125) {$@$};
		\node [style=none] (86) at (4.25, 0) {};
		\node [style=none] (87) at (5.25, 0) {};
		\node [style=none] (88) at (1.75, -0.125) {\texttt{quickly}};
		\node [style=none] (89) at (3.125, -1.375) {$@$};
		\node [style=none] (90) at (2.75, -1.25) {};
		\node [style=none] (91) at (2, -0.75) {};
		\node [style=none] (92) at (3.5, -1.25) {};
		\node [style=none] (93) at (4.25, -0.75) {};
		\node [style=none] (94) at (5.75, -1.375) {\texttt{Alice}};
		\node [style=none] (95) at (4.5, -2.625) {$@$};
		\node [style=none] (96) at (3.5, -2) {};
		\node [style=none] (97) at (4.25, -2.5) {};
		\node [style=none] (98) at (4.75, -2.5) {};
		\node [style=none] (99) at (5.5, -2) {};
		\node [style=none] (100) at (1.75, 1.875) {$\scriptstyle (n\to s)\to n\to n\to s$};
		\node [style=none] (101) at (4.5, 1.875) {$\scriptstyle n\to s$};
		\node [style=none] (102) at (3.125, 0.75) {$\scriptstyle n \to n \to s$};
		\node [style=none] (103) at (6.25, 0.75) {$\scriptstyle n$};
		\node [style=none] (104) at (4.75, -0.5) {$\scriptstyle n\to s$};
		\node [style=none] (105) at (1.75, -0.5) {$\scriptstyle (n\to s)\to n\to s$};
		\node [style=none] (106) at (3.125, -1.75) {$\scriptstyle n\to s$};
		\node [style=none] (107) at (5.75, -1.75) {$\scriptstyle n$};
		\node [style=none] (108) at (4.5, -3) {$\scriptstyle s$};
		\node [style=none] (109) at (6.5, 1.75) {};
		\node [style=none] (110) at (0.5, 2.25) {};
		\node [style=none] (111) at (3.125, -2.25) {};
		\node [style=none] (112) at (11.75, 2.25) {\textbf{C}\texttt{to}};
		\node [style=none] (113) at (13.125, 1.125) {$@$};
		\node [style=none] (114) at (14.5, 2.25) {\texttt{runs}};
		\node [style=none] (115) at (12, 1.75) {};
		\node [style=none] (116) at (12.75, 1.25) {};
		\node [style=none] (117) at (13.5, 1.25) {};
		\node [style=none] (118) at (14.25, 1.75) {};
		\node [style=none] (119) at (10, 1.125) {\textbf{B}\texttt{quickly}};
		\node [style=none] (120) at (10.25, 0.5) {};
		\node [style=none] (121) at (12.75, 0.5) {};
		\node [style=none] (122) at (11.5, -0.125) {$@$};
		\node [style=none] (123) at (12, 0) {};
		\node [style=none] (124) at (11, 0) {};
		\node [style=none] (125) at (14.5, -0.125) {\texttt{Bob}};
		\node [style=none] (126) at (13.125, -1.375) {$@$};
		\node [style=none] (127) at (13.5, -1.25) {};
		\node [style=none] (128) at (14.25, -0.75) {};
		\node [style=none] (129) at (12.75, -1.25) {};
		\node [style=none] (130) at (12, -0.75) {};
		\node [style=none] (131) at (15.5, -1.375) {\texttt{Alice}};
		\node [style=none] (132) at (14.25, -2.625) {$@$};
		\node [style=none] (133) at (13.25, -2) {};
		\node [style=none] (134) at (14, -2.5) {};
		\node [style=none] (135) at (14.5, -2.5) {};
		\node [style=none] (136) at (15.25, -2) {};
		\node [style=none] (137) at (11.75, 1.875) {$\scriptstyle (n\to s)\to n\to n\to s$};
		\node [style=none] (138) at (14.5, 1.875) {$\scriptstyle n\to s$};
		\node [style=none] (139) at (13.125, 0.75) {$\scriptstyle n \to n \to s$};
		\node [style=none] (140) at (10, 0.75) {$\scriptstyle (n\to n\to s)\to n\to n\to s$};
		\node [style=none] (141) at (11.5, -0.5) {$\scriptstyle n\to n\to s$};
		\node [style=none] (142) at (14.5, -0.5) {$\scriptstyle n$};
		\node [style=none] (143) at (13.125, -1.75) {$\scriptstyle n\to s$};
		\node [style=none] (144) at (15.5, -1.75) {$\scriptstyle n$};
		\node [style=none] (145) at (14.25, -3) {$\scriptstyle s$};
		\node [style=none] (146) at (15.75, 1.75) {};
		\node [style=none] (147) at (8.75, 1.75) {};
		\node [style=none] (148) at (13.125, -2) {};
	\end{pgfonlayer}
	\begin{pgfonlayer}{edgelayer}
		\draw (78.center) to (79.center);
		\draw (81.center) to (80.center);
		\draw (86.center) to (84.center);
		\draw (87.center) to (83.center);
		\draw (91.center) to (90.center);
		\draw (93.center) to (92.center);
		\draw (96.center) to (97.center);
		\draw (99.center) to (98.center);
		\draw [style=H] (110.center)
			 to [in=-180, out=-120] (111.center)
			 to [in=-60, out=0, looseness=0.75] (109.center)
			 to [in=60, out=120, looseness=0.75] cycle;
		\draw (115.center) to (116.center);
		\draw (118.center) to (117.center);
		\draw (123.center) to (121.center);
		\draw (124.center) to (120.center);
		\draw (128.center) to (127.center);
		\draw (130.center) to (129.center);
		\draw (133.center) to (134.center);
		\draw (136.center) to (135.center);
		\draw [style=H] (147.center)
			 to [in=-180, out=-120] (148.center)
			 to [in=-75, out=0, looseness=0.75] (146.center)
			 to [in=60, out=105, looseness=0.75] cycle;
	\end{pgfonlayer}
\end{tikzpicture}\]
In this example we just had to make one swap using the $\mathbf{C}$ combinator.
However, in general it is possible that the noun argument to be dragged out of a frame could be further up the tree of arguments than in this example.
That is, the noun argument to be dragged out may not be the last input or the second last input, but say the third last input of an term.
An example of this is given by the sentence 
$$\texttt{I think you dance better than Bob}.$$
The initial term returned by the \textit{Bobcat} parser is depicted on the left.
\begin{center}
	\makebox[\textwidth]{\makebox[1.2\textwidth]{
				\begin{tikzpicture}
	\begin{pgfonlayer}{nodelayer}
		\node [style=none] (0) at (7, 0) {$=_{\beta}$};
		\node [style=none] (75) at (-1.5, 3.25) {\texttt{than}};
		\node [style=none] (76) at (-0.125, 1.875) {$@$};
		\node [style=none] (77) at (1.25, 3) {\texttt{Bob}};
		\node [style=none] (78) at (-1.25, 2.5) {};
		\node [style=none] (79) at (-0.5, 2) {};
		\node [style=none] (80) at (0.25, 2) {};
		\node [style=none] (81) at (1, 2.5) {};
		\node [style=none] (83) at (-1.5, 2.875) {$\scriptstyle n\to ((n\to s)\to n\to s)$};
		\node [style=none] (84) at (1.25, 2.625) {$\scriptstyle n$};
		\node [style=none] (85) at (-0.125, 1.5) {$\scriptstyle ((n\to s)\to n\to s)$};
		\node [style=none] (87) at (1.25, 0.5) {$@$};
		\node [style=none] (88) at (2.625, 1.625) {\texttt{better}};
		\node [style=none] (89) at (0.125, 1.125) {};
		\node [style=none] (90) at (0.875, 0.625) {};
		\node [style=none] (91) at (1.625, 0.625) {};
		\node [style=none] (92) at (2.375, 1.125) {};
		\node [style=none] (95) at (2.625, 1.25) {$\scriptstyle (n\to s)\to n\to s$};
		\node [style=none] (96) at (1.25, 0.125) {$\scriptstyle (n\to s) \to n \to s$};
		\node [style=none] (97) at (2.625, -0.75) {$@$};
		\node [style=none] (98) at (4, 0.375) {\texttt{dance}};
		\node [style=none] (99) at (1.5, -0.125) {};
		\node [style=none] (100) at (2.25, -0.625) {};
		\node [style=none] (101) at (3, -0.625) {};
		\node [style=none] (102) at (3.75, -0.125) {};
		\node [style=none] (104) at (4, 0) {$\scriptstyle n\to s$};
		\node [style=none] (105) at (2.625, -1.125) {$\scriptstyle n \to s$};
		\node [style=none] (106) at (4.125, -2) {$@$};
		\node [style=none] (107) at (5.5, -0.875) {\texttt{you}};
		\node [style=none] (108) at (3, -1.375) {};
		\node [style=none] (109) at (3.75, -1.875) {};
		\node [style=none] (110) at (4.5, -1.875) {};
		\node [style=none] (111) at (5.25, -1.375) {};
		\node [style=none] (113) at (5.5, -1.25) {$\scriptstyle n$};
		\node [style=none] (114) at (4.125, -2.375) {$\scriptstyle s$};
		\node [style=none] (116) at (2.625, -3.25) {$@$};
		\node [style=none] (117) at (1.25, -2.125) {\texttt{think}};
		\node [style=none] (118) at (3.75, -2.625) {};
		\node [style=none] (119) at (3, -3.125) {};
		\node [style=none] (120) at (2.25, -3.125) {};
		\node [style=none] (121) at (1.5, -2.625) {};
		\node [style=none] (123) at (1.25, -2.5) {$\scriptstyle s\to n\to s$};
		\node [style=none] (124) at (2.625, -3.625) {$\scriptstyle n \to s$};
		\node [style=none] (125) at (4.125, -4.375) {$@$};
		\node [style=none] (126) at (5.5, -3.25) {\texttt{I}};
		\node [style=none] (127) at (3, -3.75) {};
		\node [style=none] (128) at (3.75, -4.25) {};
		\node [style=none] (129) at (4.5, -4.25) {};
		\node [style=none] (130) at (5.25, -3.75) {};
		\node [style=none] (132) at (5.5, -3.625) {$\scriptstyle n$};
		\node [style=none] (133) at (4.125, -4.75) {$\scriptstyle s$};
		\node [style=none] (134) at (6.25, -0.5) {};
		\node [style=none] (135) at (-0.25, 4) {};
		\node [style=none] (136) at (-3, 2.75) {};
		\node [style=none] (137) at (2.625, -3.875) {};
		\node [style=none] (138) at (-1.5, 2.625) {$\scriptstyle \to (n\to s)\to n\to s$};
		\node [style=none] (139) at (-0.125, 1.25) {$\scriptstyle \to (n\to s)\to n\to s$};
		\node [style=none] (140) at (8.5, 3.25) {\texttt{than}};
		\node [style=none] (141) at (9.875, 1.875) {$@$};
		\node [style=none] (142) at (11.25, 3) {\texttt{Bob}};
		\node [style=none] (143) at (8.75, 2.5) {};
		\node [style=none] (144) at (9.5, 2) {};
		\node [style=none] (145) at (10.25, 2) {};
		\node [style=none] (146) at (11, 2.5) {};
		\node [style=none] (147) at (8.5, 2.875) {$\scriptstyle n\to ((n\to s)\to n\to s)$};
		\node [style=none] (148) at (11.25, 2.625) {$\scriptstyle n$};
		\node [style=none] (149) at (9.875, 1.5) {$\scriptstyle ((n\to s)\to n\to s)$};
		\node [style=none] (150) at (11.25, 0.5) {$@$};
		\node [style=none] (151) at (12.625, 1.625) {\texttt{better}};
		\node [style=none] (152) at (10.125, 1.125) {};
		\node [style=none] (153) at (10.875, 0.625) {};
		\node [style=none] (154) at (11.625, 0.625) {};
		\node [style=none] (155) at (12.375, 1.125) {};
		\node [style=none] (156) at (12.625, 1.25) {$\scriptstyle (n\to s)\to n\to s$};
		\node [style=none] (157) at (11.25, 0.125) {$\scriptstyle (n\to s) \to n \to s$};
		\node [style=none] (158) at (12.625, -0.75) {$@$};
		\node [style=none] (159) at (14, 0.375) {\texttt{dance}};
		\node [style=none] (160) at (11.5, -0.125) {};
		\node [style=none] (161) at (12.25, -0.625) {};
		\node [style=none] (162) at (13, -0.625) {};
		\node [style=none] (163) at (13.75, -0.125) {};
		\node [style=none] (164) at (14, 0) {$\scriptstyle n\to s$};
		\node [style=none] (165) at (12.625, -1.125) {$\scriptstyle n \to s$};
		\node [style=none] (166) at (11.125, -2) {$@$};
		\node [style=none] (167) at (9.75, -0.875) {\textbf{B}\texttt{think}};
		\node [style=none] (168) at (12.25, -1.375) {};
		\node [style=none] (169) at (11.5, -1.875) {};
		\node [style=none] (170) at (10.75, -1.875) {};
		\node [style=none] (171) at (10, -1.375) {};
		\node [style=none] (172) at (9.75, -1.25) {$\scriptstyle (n\to s)\to n\to n\to s$};
		\node [style=none] (173) at (11.125, -2.375) {$\scriptstyle n\to n\to s$};
		\node [style=none] (174) at (12.625, -3.25) {$@$};
		\node [style=none] (175) at (14, -2.125) {\texttt{you}};
		\node [style=none] (176) at (11.5, -2.625) {};
		\node [style=none] (177) at (12.25, -3.125) {};
		\node [style=none] (178) at (13, -3.125) {};
		\node [style=none] (179) at (13.75, -2.625) {};
		\node [style=none] (180) at (14, -2.5) {$\scriptstyle n$};
		\node [style=none] (181) at (12.625, -3.625) {$\scriptstyle n \to s$};
		\node [style=none] (182) at (14.125, -4.375) {$@$};
		\node [style=none] (183) at (15.5, -3.25) {\texttt{I}};
		\node [style=none] (184) at (13, -3.75) {};
		\node [style=none] (185) at (13.75, -4.25) {};
		\node [style=none] (186) at (14.5, -4.25) {};
		\node [style=none] (187) at (15.25, -3.75) {};
		\node [style=none] (188) at (15.5, -3.625) {$\scriptstyle n$};
		\node [style=none] (189) at (14.125, -4.75) {$\scriptstyle s$};
		\node [style=none] (190) at (15.5, -0.5) {};
		\node [style=none] (191) at (9.75, 4) {};
		\node [style=none] (192) at (7, 2.75) {};
		\node [style=none] (193) at (12.625, -3.875) {};
		\node [style=none] (194) at (8.5, 2.625) {$\scriptstyle \to (n\to s)\to n\to s$};
		\node [style=none] (195) at (9.875, 1.25) {$\scriptstyle \to (n\to s)\to n\to s$};
	\end{pgfonlayer}
	\begin{pgfonlayer}{edgelayer}
		\draw (78.center) to (79.center);
		\draw (81.center) to (80.center);
		\draw (89.center) to (90.center);
		\draw (92.center) to (91.center);
		\draw (99.center) to (100.center);
		\draw (102.center) to (101.center);
		\draw (108.center) to (109.center);
		\draw (111.center) to (110.center);
		\draw (118.center) to (119.center);
		\draw (121.center) to (120.center);
		\draw (127.center) to (128.center);
		\draw (130.center) to (129.center);
		\draw [style=H] (136.center)
			 to [in=-180, out=-90, looseness=0.50] (137.center)
			 to [in=-90, out=0, looseness=0.75] (134.center)
			 to [in=0, out=90, looseness=0.75] (135.center)
			 to [in=90, out=180, looseness=0.75] cycle;
		\draw (143.center) to (144.center);
		\draw (146.center) to (145.center);
		\draw (152.center) to (153.center);
		\draw (155.center) to (154.center);
		\draw (160.center) to (161.center);
		\draw (163.center) to (162.center);
		\draw (168.center) to (169.center);
		\draw (171.center) to (170.center);
		\draw (176.center) to (177.center);
		\draw (179.center) to (178.center);
		\draw (184.center) to (185.center);
		\draw (187.center) to (186.center);
		\draw [style=H] (192.center)
			 to [in=-180, out=-90] (193.center)
			 to [in=-90, out=0, looseness=0.75] (190.center)
			 to [in=0, out=90, looseness=0.75] (191.center)
			 to [in=90, out=180, looseness=0.75] cycle;
	\end{pgfonlayer}
\end{tikzpicture}
	}}
\end{center}
We can first drag the noun \texttt{you} out of the \texttt{think} frame as usual by applying the \textbf{B} combinator, as shown.
However, at this point, \texttt{Bob} still needs to be dragged out, but \texttt{better} and \texttt{dance} do not need to be.
Hence \texttt{Bob} will need to be shuffled down two positions so we can apply the \textbf{B} combinator.

To this end we define a generalized family of \textbf{C} combinators: $\mathbf{C}_n$ takes the first input and shifts it $n$ places down while leaving order of the remaining inputs intact. For $n=1$ this is just the usual \textbf{C} combinator.
These $\mathbf{C}_n$ combinators can be constructed from vanilla \textbf{C} combinators, therefore, they are just a notational convenience.

In the example sentence above, we would thus apply $\mathbf{C}_2$ to \texttt{than} to shift \texttt{Bob} down the tree and obtain
\begin{center}
	\makebox[\textwidth]{\makebox[1.2\textwidth]{
				\begin{tikzpicture}
	\begin{pgfonlayer}{nodelayer}
		\node [style=none] (0) at (1.5, 3.75) {$\mathbf{C}_2 \texttt{than}$};
		\node [style=none] (1) at (2.875, 2.375) {$@$};
		\node [style=none] (2) at (7, 1.25) {\texttt{Bob}};
		\node [style=none] (3) at (1.75, 3) {};
		\node [style=none] (4) at (2.5, 2.5) {};
		\node [style=none] (5) at (3.25, 2.5) {};
		\node [style=none] (6) at (4, 3) {};
		\node [style=none] (7) at (1.5, 3.375) {$\scriptstyle ((n\to s)\to n\to s)$};
		\node [style=none] (8) at (7, 0.875) {$\scriptstyle n$};
		\node [style=none] (10) at (4.25, 1.25) {$@$};
		\node [style=none] (11) at (4.125, 3.625) {\texttt{better}};
		\node [style=none] (12) at (3.125, 1.875) {};
		\node [style=none] (13) at (3.875, 1.375) {};
		\node [style=none] (14) at (4.625, 1.375) {};
		\node [style=none] (15) at (5.375, 1.875) {};
		\node [style=none] (16) at (4.125, 3.25) {$\scriptstyle (n\to s)\to n\to s$};
		\node [style=none] (17) at (4.25, 0.875) {$\scriptstyle n \to n \to s$};
		\node [style=none] (18) at (5.625, 0) {$@$};
		\node [style=none] (19) at (5.5, 2.375) {\texttt{dance}};
		\node [style=none] (20) at (4.5, 0.625) {};
		\node [style=none] (21) at (5.25, 0.125) {};
		\node [style=none] (22) at (6, 0.125) {};
		\node [style=none] (23) at (6.75, 0.625) {};
		\node [style=none] (24) at (5.5, 2) {$\scriptstyle n\to s$};
		\node [style=none] (25) at (5.625, -0.375) {$\scriptstyle n \to s$};
		\node [style=none] (26) at (4.125, -1.25) {$@$};
		\node [style=none] (27) at (2.75, -0.125) {\textbf{B}\texttt{think}};
		\node [style=none] (28) at (5.25, -0.625) {};
		\node [style=none] (29) at (4.5, -1.125) {};
		\node [style=none] (30) at (3.75, -1.125) {};
		\node [style=none] (31) at (3, -0.625) {};
		\node [style=none] (32) at (2.75, -0.5) {$\scriptstyle (n\to s)\to n\to n\to s$};
		\node [style=none] (33) at (4.125, -1.625) {$\scriptstyle n\to n\to s$};
		\node [style=none] (34) at (5.625, -2.5) {$@$};
		\node [style=none] (35) at (7, -1.375) {\texttt{you}};
		\node [style=none] (36) at (4.5, -1.875) {};
		\node [style=none] (37) at (5.25, -2.375) {};
		\node [style=none] (38) at (6, -2.375) {};
		\node [style=none] (39) at (6.75, -1.875) {};
		\node [style=none] (40) at (7, -1.75) {$\scriptstyle n$};
		\node [style=none] (41) at (5.625, -2.875) {$\scriptstyle n \to s$};
		\node [style=none] (42) at (7.125, -3.625) {$@$};
		\node [style=none] (43) at (8.5, -2.5) {\texttt{I}};
		\node [style=none] (44) at (6, -3) {};
		\node [style=none] (45) at (6.75, -3.5) {};
		\node [style=none] (46) at (7.5, -3.5) {};
		\node [style=none] (47) at (8.25, -3) {};
		\node [style=none] (48) at (8.5, -2.875) {$\scriptstyle n$};
		\node [style=none] (49) at (7.125, -4) {$\scriptstyle s$};
		\node [style=none] (53) at (1.5, 3.125) {$\scriptstyle \to (n\to s)\to n\to n\to s$};
		\node [style=none] (54) at (2.875, 2) {$\scriptstyle (n\to s)\to n\to n\to s$};
		\node [style=none] (55) at (-7.5, 4) {\texttt{than}};
		\node [style=none] (56) at (-6.125, 2.625) {$@$};
		\node [style=none] (57) at (-4.75, 3.75) {\texttt{Bob}};
		\node [style=none] (58) at (-7.25, 3.25) {};
		\node [style=none] (59) at (-6.5, 2.75) {};
		\node [style=none] (60) at (-5.75, 2.75) {};
		\node [style=none] (61) at (-5, 3.25) {};
		\node [style=none] (62) at (-7.5, 3.625) {$\scriptstyle n\to ((n\to s)\to n\to s)$};
		\node [style=none] (63) at (-4.75, 3.375) {$\scriptstyle n$};
		\node [style=none] (64) at (-6.125, 2.25) {$\scriptstyle ((n\to s)\to n\to s)$};
		\node [style=none] (65) at (-4.75, 1.25) {$@$};
		\node [style=none] (66) at (-3.375, 2.375) {\texttt{better}};
		\node [style=none] (67) at (-5.875, 1.875) {};
		\node [style=none] (68) at (-5.125, 1.375) {};
		\node [style=none] (69) at (-4.375, 1.375) {};
		\node [style=none] (70) at (-3.625, 1.875) {};
		\node [style=none] (71) at (-3.375, 2) {$\scriptstyle (n\to s)\to n\to s$};
		\node [style=none] (72) at (-4.75, 0.875) {$\scriptstyle (n\to s) \to n \to s$};
		\node [style=none] (73) at (-3.375, 0) {$@$};
		\node [style=none] (74) at (-2, 1.125) {\texttt{dance}};
		\node [style=none] (75) at (-4.5, 0.625) {};
		\node [style=none] (76) at (-3.75, 0.125) {};
		\node [style=none] (77) at (-3, 0.125) {};
		\node [style=none] (78) at (-2.25, 0.625) {};
		\node [style=none] (79) at (-2, 0.75) {$\scriptstyle n\to s$};
		\node [style=none] (80) at (-3.375, -0.375) {$\scriptstyle n \to s$};
		\node [style=none] (81) at (-4.875, -1.25) {$@$};
		\node [style=none] (82) at (-6.25, -0.125) {\textbf{B}\texttt{think}};
		\node [style=none] (83) at (-3.75, -0.625) {};
		\node [style=none] (84) at (-4.5, -1.125) {};
		\node [style=none] (85) at (-5.25, -1.125) {};
		\node [style=none] (86) at (-6, -0.625) {};
		\node [style=none] (87) at (-6.25, -0.5) {$\scriptstyle (n\to s)\to n\to n\to s$};
		\node [style=none] (88) at (-4.875, -1.625) {$\scriptstyle n\to n\to s$};
		\node [style=none] (89) at (-3.375, -2.5) {$@$};
		\node [style=none] (90) at (-2, -1.375) {\texttt{you}};
		\node [style=none] (91) at (-4.5, -1.875) {};
		\node [style=none] (92) at (-3.75, -2.375) {};
		\node [style=none] (93) at (-3, -2.375) {};
		\node [style=none] (94) at (-2.25, -1.875) {};
		\node [style=none] (95) at (-2, -1.75) {$\scriptstyle n$};
		\node [style=none] (96) at (-3.375, -2.875) {$\scriptstyle n \to s$};
		\node [style=none] (97) at (-1.875, -3.625) {$@$};
		\node [style=none] (98) at (-0.5, -2.5) {\texttt{I}};
		\node [style=none] (99) at (-3, -3) {};
		\node [style=none] (100) at (-2.25, -3.5) {};
		\node [style=none] (101) at (-1.5, -3.5) {};
		\node [style=none] (102) at (-0.75, -3) {};
		\node [style=none] (103) at (-0.5, -2.875) {$\scriptstyle n$};
		\node [style=none] (104) at (-1.875, -4) {$\scriptstyle s$};
		\node [style=none] (105) at (-1, 1.75) {};
		\node [style=none] (106) at (-6.25, 4.75) {};
		\node [style=none] (107) at (-9, 3.75) {};
		\node [style=none] (108) at (-3.375, -0.625) {};
		\node [style=none] (109) at (-7.5, 3.375) {$\scriptstyle \to (n\to s)\to n\to s$};
		\node [style=none] (110) at (-6.125, 2) {$\scriptstyle \to (n\to s)\to n\to s$};
		\node [style=none] (111) at (0, 0) {$=_{\beta}$};
		\node [style=none] (112) at (8, 1.75) {};
		\node [style=none] (113) at (2.75, 4.75) {};
		\node [style=none] (114) at (0, 3.75) {};
		\node [style=none] (115) at (5.625, -0.625) {};
	\end{pgfonlayer}
	\begin{pgfonlayer}{edgelayer}
		\draw (3.center) to (4.center);
		\draw (6.center) to (5.center);
		\draw (12.center) to (13.center);
		\draw (15.center) to (14.center);
		\draw (20.center) to (21.center);
		\draw (23.center) to (22.center);
		\draw (28.center) to (29.center);
		\draw (31.center) to (30.center);
		\draw (36.center) to (37.center);
		\draw (39.center) to (38.center);
		\draw (44.center) to (45.center);
		\draw (47.center) to (46.center);
		\draw (58.center) to (59.center);
		\draw (61.center) to (60.center);
		\draw (67.center) to (68.center);
		\draw (70.center) to (69.center);
		\draw (75.center) to (76.center);
		\draw (78.center) to (77.center);
		\draw (83.center) to (84.center);
		\draw (86.center) to (85.center);
		\draw (91.center) to (92.center);
		\draw (94.center) to (93.center);
		\draw (99.center) to (100.center);
		\draw (102.center) to (101.center);
		\draw [style=H, in=165, out=-90] (107.center) to (108.center);
		\draw [style=H, in=-90, out=-15, looseness=0.75] (108.center) to (105.center);
		\draw [style=H, in=0, out=90, looseness=0.75] (105.center) to (106.center);
		\draw [style=H, in=90, out=180, looseness=0.75] (106.center) to (107.center);
		\draw [style=H, in=165, out=-90] (114.center) to (115.center);
		\draw [style=H, in=-90, out=-15, looseness=0.75] (115.center) to (112.center);
		\draw [style=H, in=0, out=90, looseness=0.75] (112.center) to (113.center);
		\draw [style=H, in=90, out=180, looseness=0.75] (113.center) to (114.center);
	\end{pgfonlayer}
\end{tikzpicture}
	}}
\end{center}
At this point we can then use another \textbf{B} combinator to drag \texttt{Bob} out of the frame \texttt{think} and arrive at a fully dragged-out term.

By recursively exchanging and dragging out the arguments (Algorithm~\ref{alg:_drag_out}), the \textbf{C} and \textbf{B} combinators allow us to drag any nouns out of higher-order frames, in the case where our term only has application types (as well as list types, if we do the obvious recursion).
Thus, at this point only abstraction $\lambda$-terms pose a problem -- but as previously mentioned, this is easily dealt with by the method described in Appendix~\ref{ssec:beta_expand_appendix}.

\begin{algorithm}[H]
	\caption{DragOut}\label{alg:_drag_out}
	\begin{algorithmic}
		\STATE Input: a $\lambda$-term
		\STATE Output: a dragged out $\lambda$-term
		\IF{term is a constant or variable}
		\RETURN term
		\ELSIF{term is a list}
		\RETURN [DragOut(t) for t in term]
		\ELSIF{term is an application}
		\STATE term $\leftarrow$ DragOut(term.function)(DragOut(term.argument))
		\FORALL{argument in GetArguments(term)}
		\STATE permutedTerm $\leftarrow$ 
		permute the argument to the root with generalized \textbf{C} combinator
		\IF{permutedTerm fulfills the four \textbf{B} combinator conditions (see Section~\ref{ssec:B_combinator})}
		\STATE term $\leftarrow$ drag out the permutedTerm using a \textbf{B} combinator
		\ENDIF
		\ENDFOR
		\RETURN term
		\ELSIF{term is an abstraction}
		\RETURN $\lambda$ (term.variable).(DragOut(term.body))
		\ENDIF
	\end{algorithmic}
\end{algorithm}

\subsection{Type expansion}
\label{ssec:type-expansion}

We now describe a kind of dual to dragging out, which is the `expansion' of certain outgoing wires into multiple $n$-wires.
This expansion comes in two flavours -- $s$-type expansion, which is simpler and conceptually cleaner, and $n$-type expansion, which is slightly more involved and requires the grammatical head information contained in the co-indices on the types.

\subsubsection{$s$-type expansion}
\label{sssec:s_type_expansion}
In the examples of the previous section, after performing dragging out, the top halves of the diagrams were in the appropriate form, but the bottom halves were not -- for instance, there is only a single $s$-type output wire. As we only have a single output wire, it is not possible to compose the circuits of multiple sentences along the wires representing the discourse referents (as done in Figure~\ref{fig:circuit_example}). This is rectified by $s$-type expansion.
A minimal example illustrating it is the sentence
$$\texttt{Alice likes Bob},$$
the CCG parse for which yields the following $\lambda$-term and accompanying diagram
\begin{center}
	\begin{minipage}{0.49\linewidth}
	\[\begin{tikzpicture}
	\begin{pgfonlayer}{nodelayer}
		\node [style=none] (0) at (11.125, -2.125) {\texttt{likes}};
		\node [style=none] (1) at (11.125, -2.5) {$\scriptstyle n\to n\to s$};
		\node [style=none] (2) at (12.625, -3.25) {$@$};
		\node [style=none] (3) at (14.125, -2.125) {\texttt{Bob}};
		\node [style=none] (4) at (11.5, -2.625) {};
		\node [style=none] (5) at (12.25, -3.125) {};
		\node [style=none] (6) at (13, -3.125) {};
		\node [style=none] (7) at (13.75, -2.625) {};
		\node [style=none] (8) at (14.125, -2.5) {$\scriptstyle n$};
		\node [style=none] (9) at (12.625, -3.625) {$\scriptstyle n \to s$};
		\node [style=none] (10) at (14.125, -4.375) {$@$};
		\node [style=none] (11) at (15.5, -3.25) {\texttt{Alice}};
		\node [style=none] (12) at (13, -3.75) {};
		\node [style=none] (13) at (13.75, -4.25) {};
		\node [style=none] (14) at (14.5, -4.25) {};
		\node [style=none] (15) at (15.25, -3.75) {};
		\node [style=none] (16) at (15.5, -3.625) {$\scriptstyle n$};
		\node [style=none] (17) at (14.125, -4.75) {$\scriptstyle s$};
	\end{pgfonlayer}
	\begin{pgfonlayer}{edgelayer}
		\draw (4.center) to (5.center);
		\draw (7.center) to (6.center);
		\draw (12.center) to (13.center);
		\draw (15.center) to (14.center);
	\end{pgfonlayer}
\end{tikzpicture}\]
	\end{minipage}
	\begin{minipage}{0.49\linewidth}
	\includegraphics[width=\textwidth]{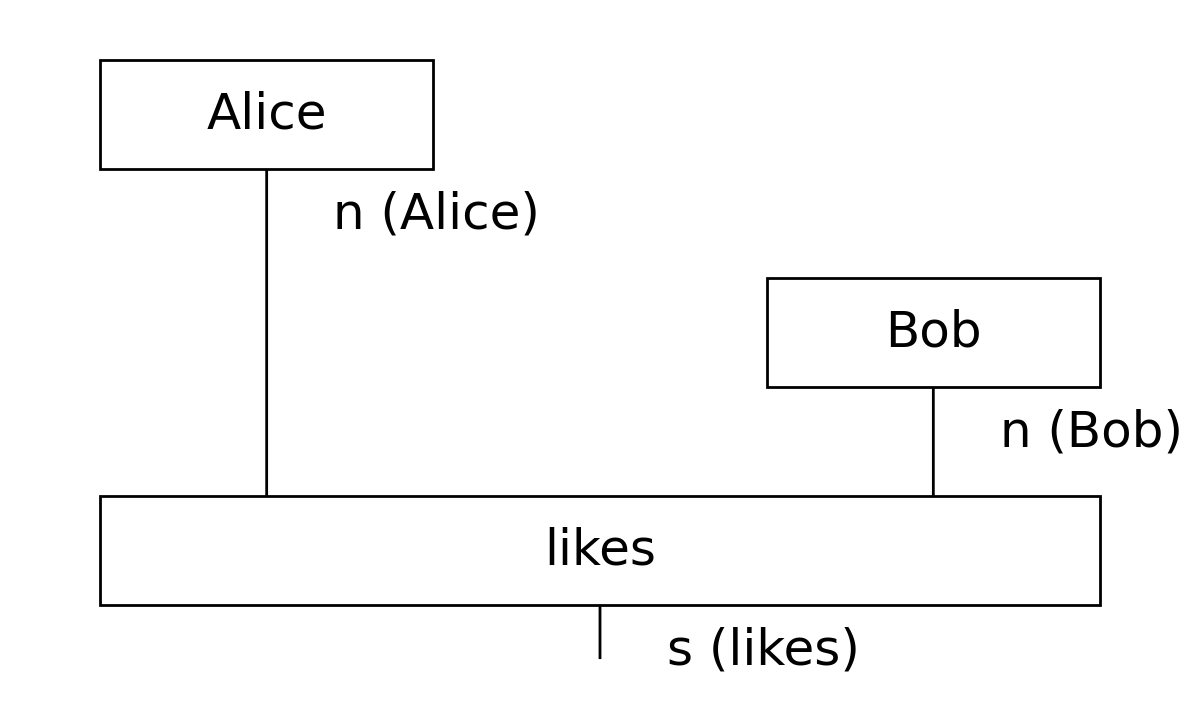}
	\end{minipage}
\end{center}
Following the philosophy of DisCoCirc means that we would like to replace the outgoing $s$-type wire with two $n$-type wires -- one for \texttt{Alice} and one for \texttt{Bob}.
That is, we would like to rewrite the above to the following (note the introduction of product types $\times$ in the $\lambda$-tree)
\begin{center}
	\begin{minipage}{0.49\linewidth}
	\[\begin{tikzpicture}
	\begin{pgfonlayer}{nodelayer}
		\node [style=none] (0) at (11.125, -2.125) {\texttt{likes}};
		\node [style=none] (1) at (11.125, -2.5) {$\scriptstyle n\to n\to n\times n$};
		\node [style=none] (2) at (12.625, -3.25) {$@$};
		\node [style=none] (3) at (14.125, -2.125) {\texttt{Bob}};
		\node [style=none] (4) at (11.5, -2.625) {};
		\node [style=none] (5) at (12.25, -3.125) {};
		\node [style=none] (6) at (13, -3.125) {};
		\node [style=none] (7) at (13.75, -2.625) {};
		\node [style=none] (8) at (14.125, -2.5) {$\scriptstyle n$};
		\node [style=none] (9) at (12.625, -3.625) {$\scriptstyle n \to n\times n$};
		\node [style=none] (10) at (14.125, -4.375) {$@$};
		\node [style=none] (11) at (15.5, -3.25) {\texttt{Alice}};
		\node [style=none] (12) at (13, -3.75) {};
		\node [style=none] (13) at (13.75, -4.25) {};
		\node [style=none] (14) at (14.5, -4.25) {};
		\node [style=none] (15) at (15.25, -3.75) {};
		\node [style=none] (16) at (15.5, -3.625) {$\scriptstyle n$};
		\node [style=none] (17) at (14.125, -4.75) {$\scriptstyle n\times n$};
	\end{pgfonlayer}
	\begin{pgfonlayer}{edgelayer}
		\draw (4.center) to (5.center);
		\draw (7.center) to (6.center);
		\draw (12.center) to (13.center);
		\draw (15.center) to (14.center);
	\end{pgfonlayer}
\end{tikzpicture}\]
	\end{minipage}
	\begin{minipage}{0.49\linewidth}
	\includegraphics[width=\textwidth]{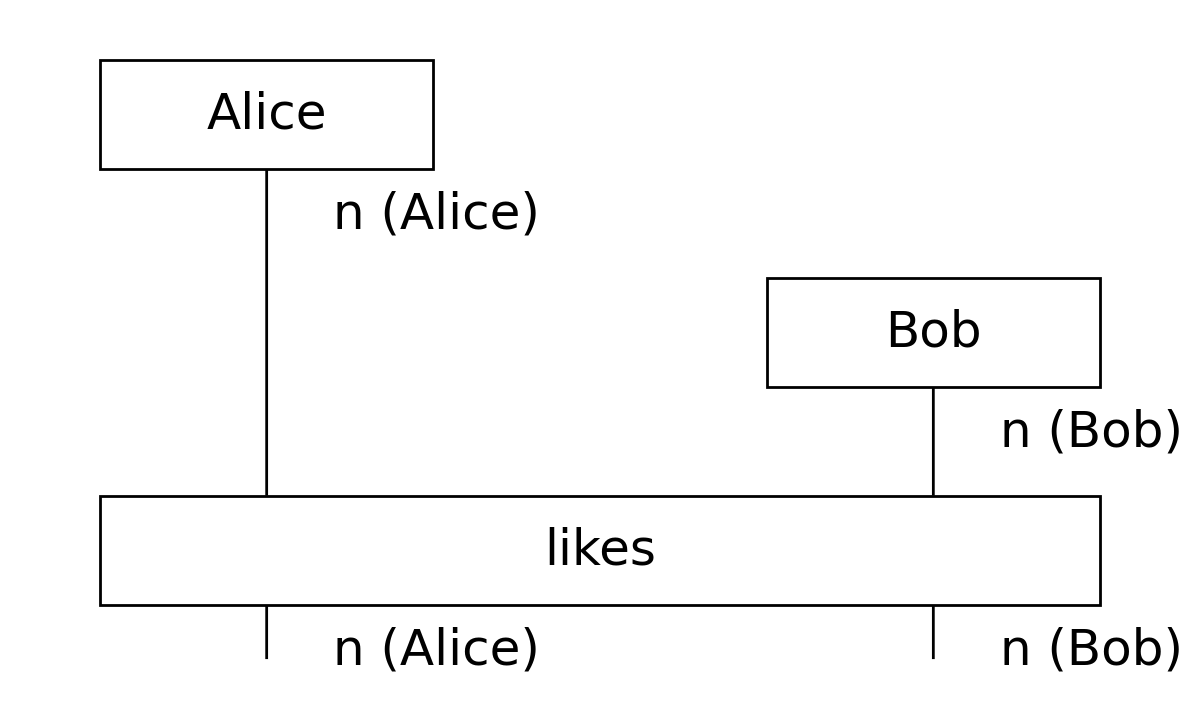}
	\end{minipage}
\end{center}
This is now a valid text circuit.

Of course, more complicated cases than this can arise.
For instance, we can have higher-order frames where different $s$ wires inside and outside the frame may need to be expanded into different numbers of $n$ wires. 
Consider the sentence
$$\texttt{I dreamt Bob punched Charlie}$$
The type expansion is performed after dragging out (see Figure~\ref{fig:pipeline}). Thus, after dragging out, we have 
\begin{center}
	\makebox[\textwidth]{\makebox[1.2\textwidth]{
			\begin{minipage}{0.59\linewidth}
			\[\begin{tikzpicture}
	\begin{pgfonlayer}{nodelayer}
		\node [style=none] (0) at (11.125, -2.125) {\texttt{punched}};
		\node [style=none] (1) at (11.125, -2.5) {$\scriptstyle n\to n\to s$};
		\node [style=none] (2) at (9.625, -3.25) {$@$};
		\node [style=none] (3) at (12.625, -3.125) {\texttt{Charlie}};
		\node [style=none] (4) at (8.5, -2.625) {};
		\node [style=none] (5) at (9.25, -3.125) {};
		\node [style=none] (6) at (10, -3.125) {};
		\node [style=none] (7) at (10.75, -2.625) {};
		\node [style=none] (8) at (12.625, -3.5) {$\scriptstyle n$};
		\node [style=none] (9) at (9.625, -3.625) {$\scriptstyle n\to n\to n \to s$};
		\node [style=none] (10) at (11.125, -4.375) {$@$};
		\node [style=none] (11) at (14, -4.25) {\texttt{Bob}};
		\node [style=none] (12) at (10, -3.75) {};
		\node [style=none] (13) at (10.75, -4.25) {};
		\node [style=none] (14) at (11.5, -4.25) {};
		\node [style=none] (15) at (12.25, -3.75) {};
		\node [style=none] (16) at (14, -4.625) {$\scriptstyle n$};
		\node [style=none] (17) at (11.125, -4.75) {$\scriptstyle n\to n\to s$};
		\node [style=none] (18) at (8.125, -1.875) {\texttt{dreamt}};
		\node [style=none] (19) at (8.125, -2.25) {$\scriptstyle (n\to n\to s)$};
		\node [style=none] (20) at (12.625, -5.5) {$@$};
		\node [style=none] (21) at (13.75, -4.875) {};
		\node [style=none] (22) at (13, -5.375) {};
		\node [style=none] (23) at (12.25, -5.375) {};
		\node [style=none] (24) at (11.5, -4.875) {};
		\node [style=none] (25) at (12.6, -5.875) {$\scriptstyle n\to s$};
		\node [style=none] (26) at (14.1, -6.625) {$@$};
		\node [style=none] (27) at (15.6, -5.5) {\texttt{I}};
		\node [style=none] (28) at (12.975, -6) {};
		\node [style=none] (29) at (13.725, -6.5) {};
		\node [style=none] (30) at (14.475, -6.5) {};
		\node [style=none] (31) at (15.225, -6) {};
		\node [style=none] (32) at (15.6, -5.875) {$\scriptstyle n$};
		\node [style=none] (33) at (14.1, -7) {$\scriptstyle s$};
		\node [style=none] (34) at (8.125, -2.5) {$\scriptstyle \to n\to n\to n\to s$};
	\end{pgfonlayer}
	\begin{pgfonlayer}{edgelayer}
		\draw (4.center) to (5.center);
		\draw (7.center) to (6.center);
		\draw (12.center) to (13.center);
		\draw (15.center) to (14.center);
		\draw (21.center) to (22.center);
		\draw (24.center) to (23.center);
		\draw (28.center) to (29.center);
		\draw (31.center) to (30.center);
	\end{pgfonlayer}
\end{tikzpicture}\]
			\end{minipage}
			\begin{minipage}{0.59\linewidth}
			\includegraphics[width=\textwidth]{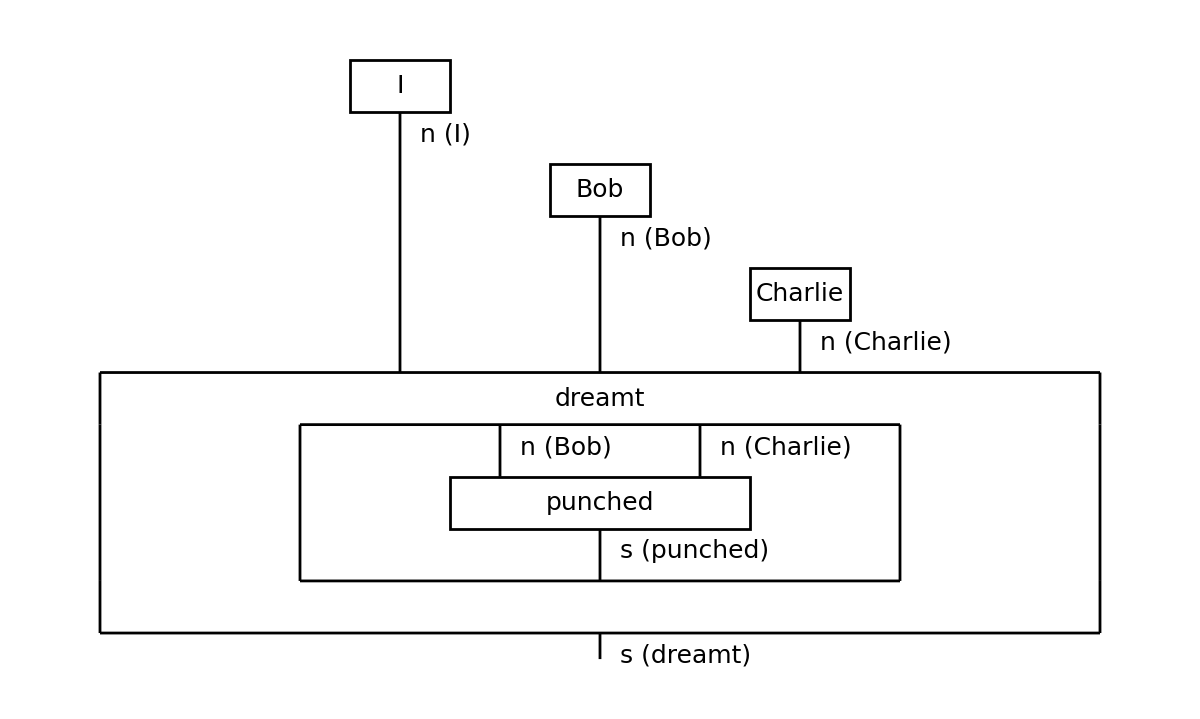}
			\end{minipage}
	}}
\end{center}
To do $s$-expansion here, the $s$-wire inside the frame should be expanded to two $n$ wires (co-indexed as \texttt{Bob} and \texttt{Charlie} respectively), whereas the $s$-wire output for the overall sentence should be expanded into three wires (co-indexed with \texttt{I}, \texttt{Bob}, and \texttt{Charlie} respectively).
Thus, the $s$-expanded form is
\begin{center}
	\makebox[\textwidth]{\makebox[1.2\textwidth]{
			\begin{minipage}{0.59\linewidth}
				\[\begin{tikzpicture}
	\begin{pgfonlayer}{nodelayer}
		\node [style=none] (0) at (0.875, 2.125) {\texttt{punched}};
		\node [style=none] (1) at (0.875, 1.75) {$\scriptstyle n\to n\to n\times n$};
		\node [style=none] (2) at (-0.625, 1) {$@$};
		\node [style=none] (3) at (2.375, 1.125) {\texttt{Charlie}};
		\node [style=none] (4) at (-1.75, 1.625) {};
		\node [style=none] (5) at (-1, 1.125) {};
		\node [style=none] (6) at (-0.25, 1.125) {};
		\node [style=none] (7) at (0.5, 1.625) {};
		\node [style=none] (8) at (2.375, 0.75) {$\scriptstyle n$};
		\node [style=none] (9) at (-0.625, 0.625) {$\scriptstyle n\to n\to n \to n\times n\times n$};
		\node [style=none] (10) at (0.875, -0.125) {$@$};
		\node [style=none] (11) at (3.75, 0) {\texttt{Bob}};
		\node [style=none] (12) at (-0.25, 0.5) {};
		\node [style=none] (13) at (0.5, 0) {};
		\node [style=none] (14) at (1.25, 0) {};
		\node [style=none] (15) at (2, 0.5) {};
		\node [style=none] (16) at (3.75, -0.375) {$\scriptstyle n$};
		\node [style=none] (17) at (0.875, -0.5) {$\scriptstyle n\to n\to n\times n\times n$};
		\node [style=none] (18) at (-2.125, 2.375) {\texttt{dreamt}};
		\node [style=none] (19) at (-2.125, 2) {$\scriptstyle (n\to n\to n\times n)$};
		\node [style=none] (20) at (2.375, -1.25) {$@$};
		\node [style=none] (21) at (3.5, -0.625) {};
		\node [style=none] (22) at (2.75, -1.125) {};
		\node [style=none] (23) at (2, -1.125) {};
		\node [style=none] (24) at (1.25, -0.625) {};
		\node [style=none] (25) at (2.35, -1.625) {$\scriptstyle n\to n\times n\times n$};
		\node [style=none] (26) at (3.85, -2.375) {$@$};
		\node [style=none] (27) at (5.35, -1.25) {\texttt{I}};
		\node [style=none] (28) at (2.725, -1.75) {};
		\node [style=none] (29) at (3.475, -2.25) {};
		\node [style=none] (30) at (4.225, -2.25) {};
		\node [style=none] (31) at (4.975, -1.75) {};
		\node [style=none] (32) at (5.35, -1.625) {$\scriptstyle n$};
		\node [style=none] (33) at (3.85, -2.75) {$\scriptstyle n\times n\times n$};
		\node [style=none] (34) at (-2.125, 1.75) {$\scriptstyle \to n\to n\to n\to n\times n\times n$};
	\end{pgfonlayer}
	\begin{pgfonlayer}{edgelayer}
		\draw (4.center) to (5.center);
		\draw (7.center) to (6.center);
		\draw (12.center) to (13.center);
		\draw (15.center) to (14.center);
		\draw (21.center) to (22.center);
		\draw (24.center) to (23.center);
		\draw (28.center) to (29.center);
		\draw (31.center) to (30.center);
	\end{pgfonlayer}
\end{tikzpicture}\]
			\end{minipage}
			\begin{minipage}{0.59\linewidth}
	\includegraphics[width=\textwidth]{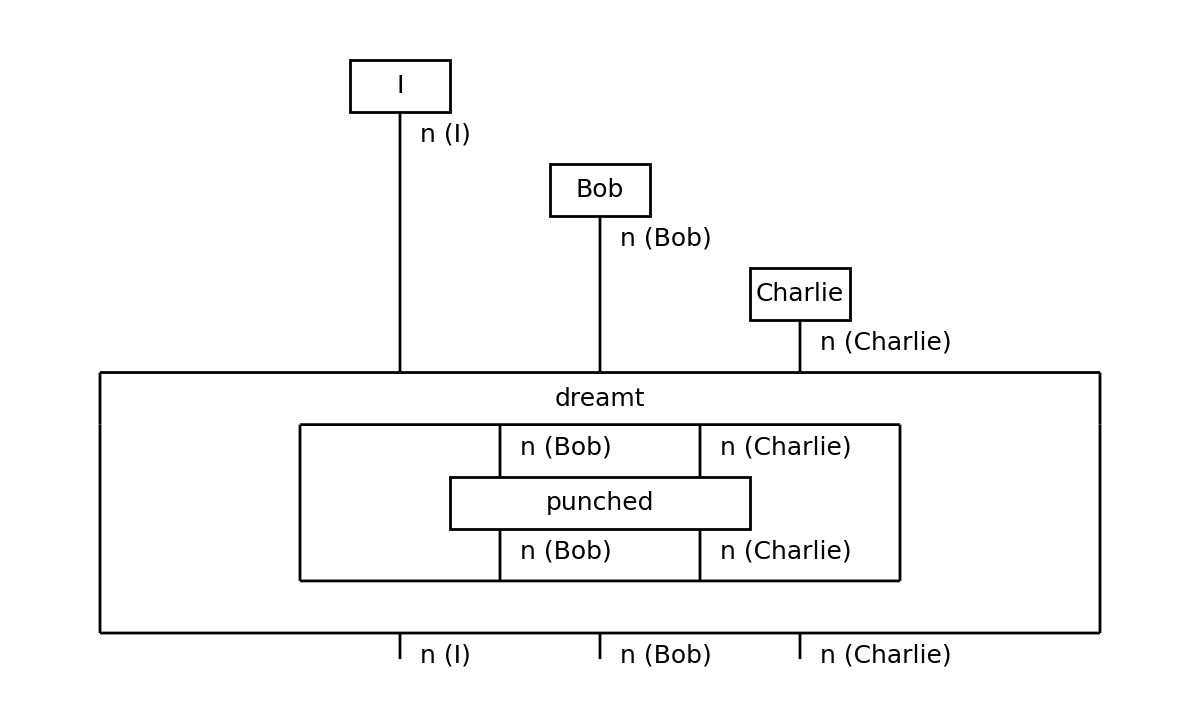}
			\end{minipage}
	}}
\end{center}
For some intuition on how $s$-expansion affects the types, consider the original type of \texttt{dreamt} in this example, which is
$$(n\to n \to s)\to n \to n\to n \to s.$$
In its uncurried form, this is
$$(n\times n\to s) \times n \times n\times n \to s.$$
To turn this type into an $s$-expanded form, we just have to match the $s$ with the number of $n$ wire inputs at the relevant level.
So, the 1st-order type in the input $(n\times n\to s)$ becomes $(n\times n\to n\times n)$, whereas the outer $s$ becomes $n\times n\times n$ to match the three $n$'s in its corresponding input.
This gives
$$(n\times n\to n\times n)\times n\times n\times n \to n\times n\times n$$
which is exactly the uncurried version of the new type of \texttt{dreamt}.
The same logic explains how $s$-type expansion modifies the type of \texttt{likes} in the first example.

Note that in both the example sentences above, the structure of the $\lambda$-tree and the connectivity of the diagram remains unchanged by $s$-expansion.
In both cases the change is that a few $s$ types/wires have been replaced by a product of $n$ types/wires.
Indeed, it turns out that under certain assumptions, $s$-type expansion can be done by precisely this simple type-rewriting operation.
Under these assumptions, in order to perform $s$-expansion on a $\lambda$-term it suffices to
\begin{itemize}
	\item first recurse up to the constants (the leaves of the tree), and rewrite their types into an $s$-expanded form
	\item recursively move back down to the root of the tree, recombining the $s$-expanded subtrees as they originally were.
\end{itemize}
That is, we leave the overall structure of the $\lambda$-tree unchanged.
By recombining the $s$-expanded constants, the changes made to their types will propagate through the types of the whole tree.

We can determine what a constant or variable's type should be rewritten into purely by looking at its type and without any other information -- like we did with the type of \texttt{dreamt} above.
The function that does this rewriting of types into $s$-expanded form is described in Algorithm~\ref{alg:expand_type}. It is important to observe that $s$-type expansion adds noun wires with head information (as can be seen in the circuit representation of the terms). Therefore, the algorithm has to identify the missing noun wires, adding minor additional complexity in the implementation.
\begin{algorithm}
	\caption{SExpandType}\label{alg:expand_type}
	\begin{algorithmic}
		\STATE Input: a type
		\STATE Output: an $s$-expanded type
		\IF{type is atomic}
		\RETURN type
		\ELSIF{type is a product}
		\RETURN product(SExpandType(t) for t in type)
		\ELSIF{type is a function}
		\STATE nounInputs $\leftarrow$ [arg \textbf{if} arg.type = \textit{n} \textbf{for} arg in inputs(type)]
		\STATE nounOutputs $\leftarrow$ [arg \textbf{if} arg.type = \textit{n} \textbf{for} arg in outputs(type)]
		\STATE missingNouns
		$\leftarrow$ [noun \textbf{if} noun not in nounOutputs \textbf{for} noun in nounInputs]
		\FORALL{types t in outputs(type)}
		\STATE \# we assume that we have at most one $s$-type
		\IF{t is $s$-type}
		\STATE type $\leftarrow$ replace $t$ in outputs(type) with missingNouns
		\RETURN type
		\ENDIF
		\ENDFOR
		\ENDIF
	\end{algorithmic}
\end{algorithm}


\subsubsection{$n$-type expansion}
\label{sssec:n_type_expansion}

It is not only $s$-type wires that may carry multiple noun referents -- sometimes $n$-type wires can too, in which case they also need to be expanded.
$n$-type expansion is more complicated than $s$-type expansion in that it substantively changes the structure of the $\lambda$-term and is not a purely local algorithm.
In the actual pipeline, this step is performed before $s$-type expansion.

A prototypical example requiring $n$-type expansion is a relative clause introduced by a relative pronoun
$$\texttt{Bob who loves Alice runs}.$$
After dragging out we have
\begin{center}
	\begin{minipage}{0.49\linewidth}
		\[\begin{tikzpicture}
	\begin{pgfonlayer}{nodelayer}
		\node [style=none] (0) at (1.5, 1.375) {\texttt{Alice}};
		\node [style=none] (1) at (1.5, 1) {$\scriptstyle n$};
		\node [style=none] (2) at (-1.5, 1.25) {$@$};
		\node [style=none] (3) at (0, 2.375) {\texttt{loves}};
		\node [style=none] (4) at (-2.625, 1.875) {};
		\node [style=none] (5) at (-1.875, 1.375) {};
		\node [style=none] (6) at (-1.125, 1.375) {};
		\node [style=none] (7) at (-0.375, 1.875) {};
		\node [style=none] (8) at (0, 2) {$\scriptstyle n\to n\to s$};
		\node [style=none] (9) at (-1.5, 0.875) {$\scriptstyle n \to n\to n$};
		\node [style=none] (10) at (0, 0.125) {$@$};
		\node [style=none] (11) at (-3, 2.5) {\texttt{who}};
		\node [style=none] (12) at (-1.125, 0.75) {};
		\node [style=none] (13) at (-0.375, 0.25) {};
		\node [style=none] (14) at (0.375, 0.25) {};
		\node [style=none] (15) at (1.125, 0.75) {};
		\node [style=none] (16) at (-3, 2.125) {$\scriptstyle (n\to n\to s)\to n\to n\to n$};
		\node [style=none] (17) at (0, -0.25) {$\scriptstyle n\to n$};
		\node [style=none] (18) at (3, 0.125) {\texttt{Bob}};
		\node [style=none] (19) at (3, -0.25) {$\scriptstyle n$};
		\node [style=none] (20) at (1.5, -1) {$@$};
		\node [style=none] (21) at (0.375, -0.375) {};
		\node [style=none] (22) at (1.125, -0.875) {};
		\node [style=none] (23) at (1.875, -0.875) {};
		\node [style=none] (24) at (2.625, -0.375) {};
		\node [style=none] (25) at (1.5, -1.375) {$\scriptstyle n$};
		\node [style=none] (26) at (0, -2.125) {$@$};
		\node [style=none] (27) at (-1.5, -1) {\texttt{runs}};
		\node [style=none] (28) at (1.125, -1.5) {};
		\node [style=none] (29) at (0.375, -2) {};
		\node [style=none] (30) at (-0.375, -2) {};
		\node [style=none] (31) at (-1.125, -1.5) {};
		\node [style=none] (32) at (-1.5, -1.375) {$\scriptstyle n\to s$};
		\node [style=none] (33) at (0, -2.5) {$\scriptstyle s$};
	\end{pgfonlayer}
	\begin{pgfonlayer}{edgelayer}
		\draw (4.center) to (5.center);
		\draw (7.center) to (6.center);
		\draw (12.center) to (13.center);
		\draw (15.center) to (14.center);
		\draw (21.center) to (22.center);
		\draw (24.center) to (23.center);
		\draw (28.center) to (29.center);
		\draw (31.center) to (30.center);
	\end{pgfonlayer}
\end{tikzpicture}\]
	\end{minipage}
	\begin{minipage}{0.49\linewidth}
	\includegraphics[width=\textwidth]{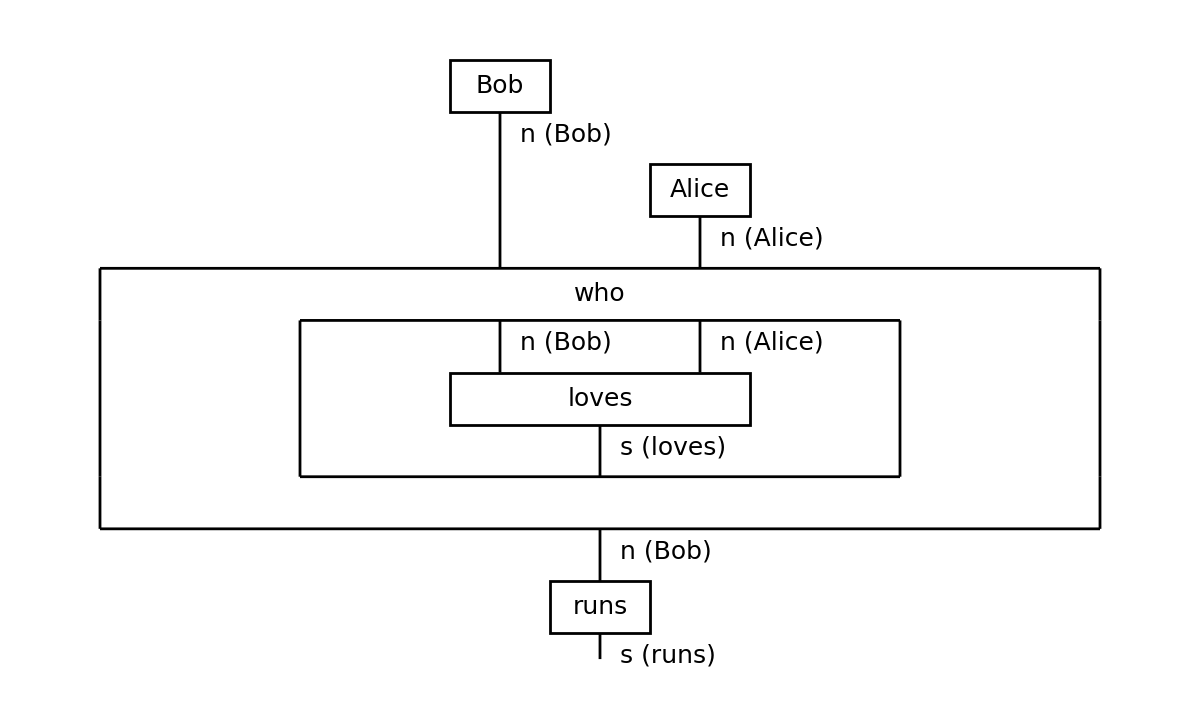}
	\end{minipage}
\end{center}
Now, if we were to just call $s$-expansion at this point, we would obtain
\begin{center}
	\includegraphics[width=0.49\textwidth]{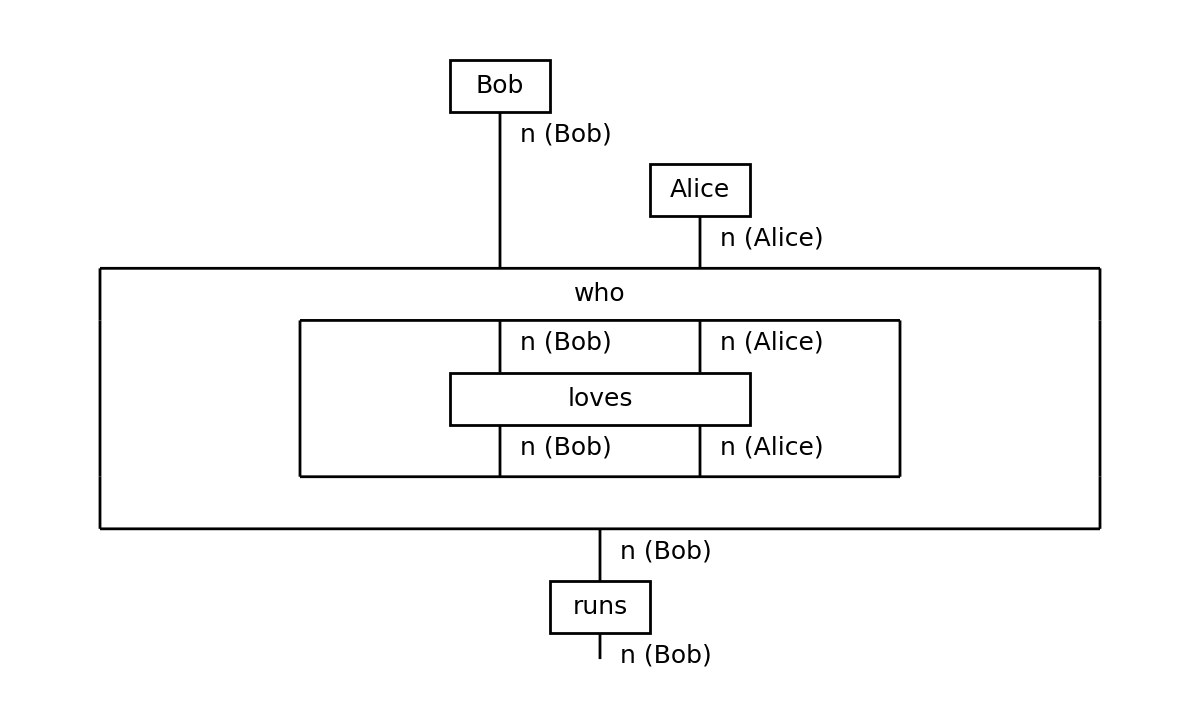}
\end{center}
where the one $n$-wire output of the component \texttt{Bob who loves Alice} stands in for \texttt{Bob}.
\texttt{Alice}, however, is not passed on as a wire.

Instead of calling $s$-expansion here then, we first perform $n$-expansion.
In this case, the $n$-wire corresponding to the phrase \texttt{Bob who loves Alice} will be modified --  effectively we just introduce a second $n$-wire that is indexed as \texttt{Alice}
\begin{center}
	\begin{minipage}{0.49\linewidth}
		\[\begin{tikzpicture}
	\begin{pgfonlayer}{nodelayer}
		\node [style=none] (0) at (3, 1.375) {\texttt{Alice}};
		\node [style=none] (1) at (3, 1) {$\scriptstyle n$};
		\node [style=none] (2) at (0, 1.25) {$@$};
		\node [style=none] (3) at (1.5, 2.375) {\texttt{loves}};
		\node [style=none] (4) at (-1.125, 1.875) {};
		\node [style=none] (5) at (-0.375, 1.375) {};
		\node [style=none] (6) at (0.375, 1.375) {};
		\node [style=none] (7) at (1.125, 1.875) {};
		\node [style=none] (8) at (1.5, 2) {$\scriptstyle n\to n\to s$};
		\node [style=none] (9) at (0, 0.875) {$\scriptstyle n \to n\to n\times n$};
		\node [style=none] (10) at (1.5, 0.125) {$@$};
		\node [style=none] (11) at (-1.5, 2.375) {\texttt{who}};
		\node [style=none] (12) at (0.375, 0.75) {};
		\node [style=none] (13) at (1.125, 0.25) {};
		\node [style=none] (14) at (1.875, 0.25) {};
		\node [style=none] (15) at (2.625, 0.75) {};
		\node [style=none] (16) at (-1.5, 2) {$\scriptstyle (n\to n\to s)\to n\to n\to n\times n$};
		\node [style=none] (17) at (1.5, -0.25) {$\scriptstyle n\to n\times n$};
		\node [style=none] (18) at (4.5, 0.125) {\texttt{Bob}};
		\node [style=none] (19) at (4.5, -0.25) {$\scriptstyle n$};
		\node [style=none] (20) at (3, -1) {$@$};
		\node [style=none] (21) at (1.875, -0.375) {};
		\node [style=none] (22) at (2.625, -0.875) {};
		\node [style=none] (23) at (3.375, -0.875) {};
		\node [style=none] (24) at (4.125, -0.375) {};
		\node [style=none] (25) at (3, -1.375) {$\scriptstyle n\times n$};
	\end{pgfonlayer}
	\begin{pgfonlayer}{edgelayer}
		\draw (4.center) to (5.center);
		\draw (7.center) to (6.center);
		\draw (12.center) to (13.center);
		\draw (15.center) to (14.center);
		\draw (21.center) to (22.center);
		\draw (24.center) to (23.center);
	\end{pgfonlayer}
\end{tikzpicture}\]
	\end{minipage}
	\begin{minipage}{0.49\linewidth}
	\includegraphics[width=\textwidth]{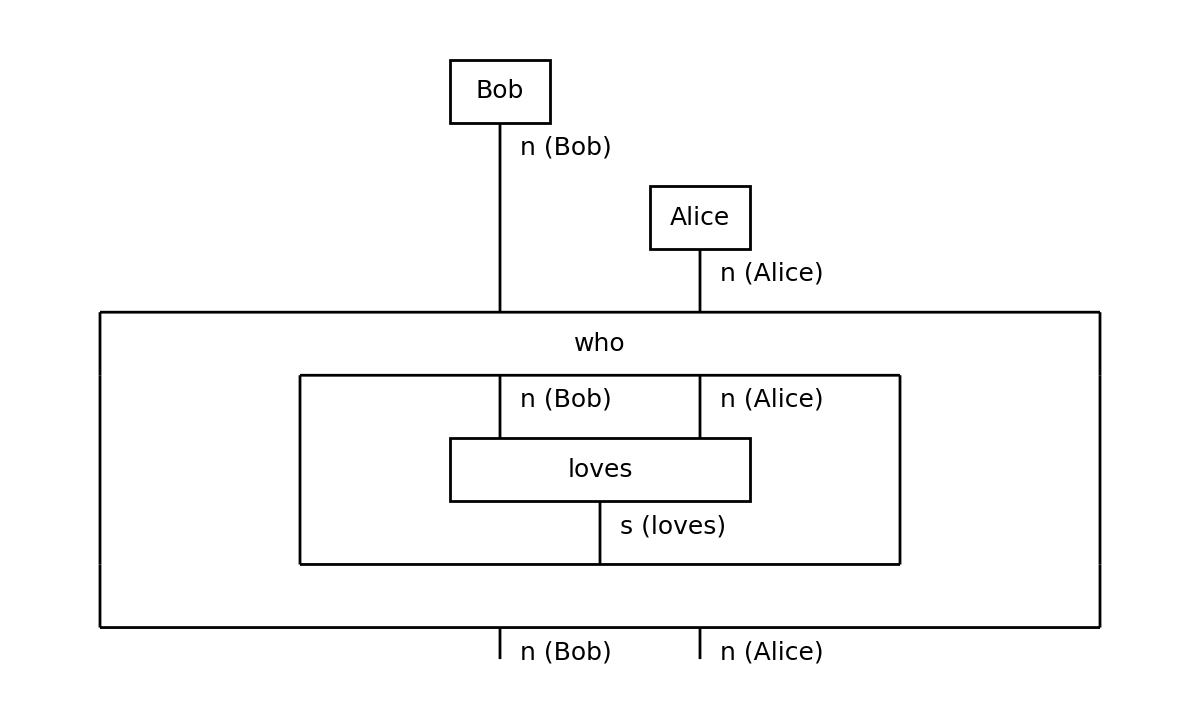}
	\end{minipage}
\end{center}
Algorithmically, this so far works similarly to $s$-type expansion, in that we recurse up to the leaves, $n$-expand their types, and then by recombining the $\lambda$-tree, the $n$-expanded types percolate down towards the root of the tree.

However, a departure from $s$-type expansion occurs at this point, because we now need to compose this $n$-expanded component with the latter part of the circuit -- a \texttt{runs} gate that acts on the head noun \texttt{Bob}.
Using the head information carried in the wire co-indices, we reinstate this and obtain the following term
\[\begin{tikzpicture}
	\begin{pgfonlayer}{nodelayer}
		\node [style=none] (0) at (2.5, 1.875) {\texttt{Alice}};
		\node [style=none] (1) at (2.5, 1.5) {$\scriptstyle n$};
		\node [style=none] (2) at (0, 1.75) {$@$};
		\node [style=none] (3) at (1.25, 2.875) {\texttt{loves}};
		\node [style=none] (4) at (-0.875, 2.375) {};
		\node [style=none] (5) at (-0.375, 1.875) {};
		\node [style=none] (6) at (0.375, 1.875) {};
		\node [style=none] (7) at (0.875, 2.375) {};
		\node [style=none] (8) at (1.25, 2.5) {$\scriptstyle n\to n\to s$};
		\node [style=none] (9) at (0, 1.375) {$\scriptstyle n \to n\to n\times n$};
		\node [style=none] (10) at (1.25, 0.625) {$@$};
		\node [style=none] (11) at (-1.25, 2.875) {\texttt{who}};
		\node [style=none] (12) at (0.375, 1.25) {};
		\node [style=none] (13) at (0.875, 0.75) {};
		\node [style=none] (14) at (1.625, 0.75) {};
		\node [style=none] (15) at (2.125, 1.25) {};
		\node [style=none] (16) at (-1.25, 2.5) {$\scriptstyle (n\to n\to s)\to n\to n\to n\times n$};
		\node [style=none] (17) at (1.25, 0.25) {$\scriptstyle n\to n\times n$};
		\node [style=none] (18) at (3.75, 0.625) {\texttt{Bob}};
		\node [style=none] (19) at (3.75, 0.25) {$\scriptstyle n$};
		\node [style=none] (20) at (2.5, -0.5) {$@$};
		\node [style=none] (21) at (1.625, 0.125) {};
		\node [style=none] (22) at (2.125, -0.375) {};
		\node [style=none] (23) at (2.875, -0.375) {};
		\node [style=none] (24) at (3.375, 0.125) {};
		\node [style=none] (25) at (2.5, -0.875) {$\scriptstyle n\times n$};
		\node [style=none] (26) at (-3.75, 1.875) {$y$};
		\node [style=none] (27) at (-3.75, 1.5) {$\scriptstyle n$};
		\node [style=none] (28) at (-6.25, 1.75) {$@$};
		\node [style=none] (29) at (-5, 2.875) {$x$};
		\node [style=none] (30) at (-7.125, 2.375) {};
		\node [style=none] (31) at (-6.625, 1.875) {};
		\node [style=none] (32) at (-5.875, 1.875) {};
		\node [style=none] (33) at (-5.375, 2.375) {};
		\node [style=none] (34) at (-5, 2.5) {$\scriptstyle n$};
		\node [style=none] (35) at (-6.25, 1.375) {$\scriptstyle s$};
		\node [style=none] (36) at (-5, 0.625) {$\times$};
		\node [style=none] (37) at (-7.5, 2.875) {\texttt{runs}};
		\node [style=none] (38) at (-5.875, 1.25) {};
		\node [style=none] (39) at (-5.375, 0.75) {};
		\node [style=none] (40) at (-4.625, 0.75) {};
		\node [style=none] (41) at (-4.125, 1.25) {};
		\node [style=none] (42) at (-7.5, 2.5) {$\scriptstyle n\to s$};
		\node [style=none] (43) at (-5, 0.25) {$\scriptstyle s\times n$};
		\node [style=none] (44) at (-5, -0.75) {$\lambda (x,y)$};
		\node [style=none] (47) at (-5, -1.125) {$\scriptstyle n\times n\to s\times n$};
		\node [style=none] (48) at (-5, -0.5) {};
		\node [style=none] (49) at (-1, -1.875) {$@$};
		\node [style=none] (50) at (-4.625, -1.375) {};
		\node [style=none] (51) at (-1.25, -1.75) {};
		\node [style=none] (52) at (-0.75, -1.75) {};
		\node [style=none] (53) at (2.125, -1.125) {};
		\node [style=none] (54) at (-1, -2.25) {$\scriptstyle s\times n$};
		\node [style=none] (55) at (-5, 0) {};
	\end{pgfonlayer}
	\begin{pgfonlayer}{edgelayer}
		\draw (4.center) to (5.center);
		\draw (7.center) to (6.center);
		\draw (12.center) to (13.center);
		\draw (15.center) to (14.center);
		\draw (21.center) to (22.center);
		\draw (24.center) to (23.center);
		\draw (30.center) to (31.center);
		\draw (33.center) to (32.center);
		\draw (38.center) to (39.center);
		\draw (41.center) to (40.center);
		\draw (50.center) to (51.center);
		\draw (53.center) to (52.center);
		\draw (48.center) to (55.center);
	\end{pgfonlayer}
\end{tikzpicture}\]
\begin{center}
	\includegraphics[width=0.6\textwidth]{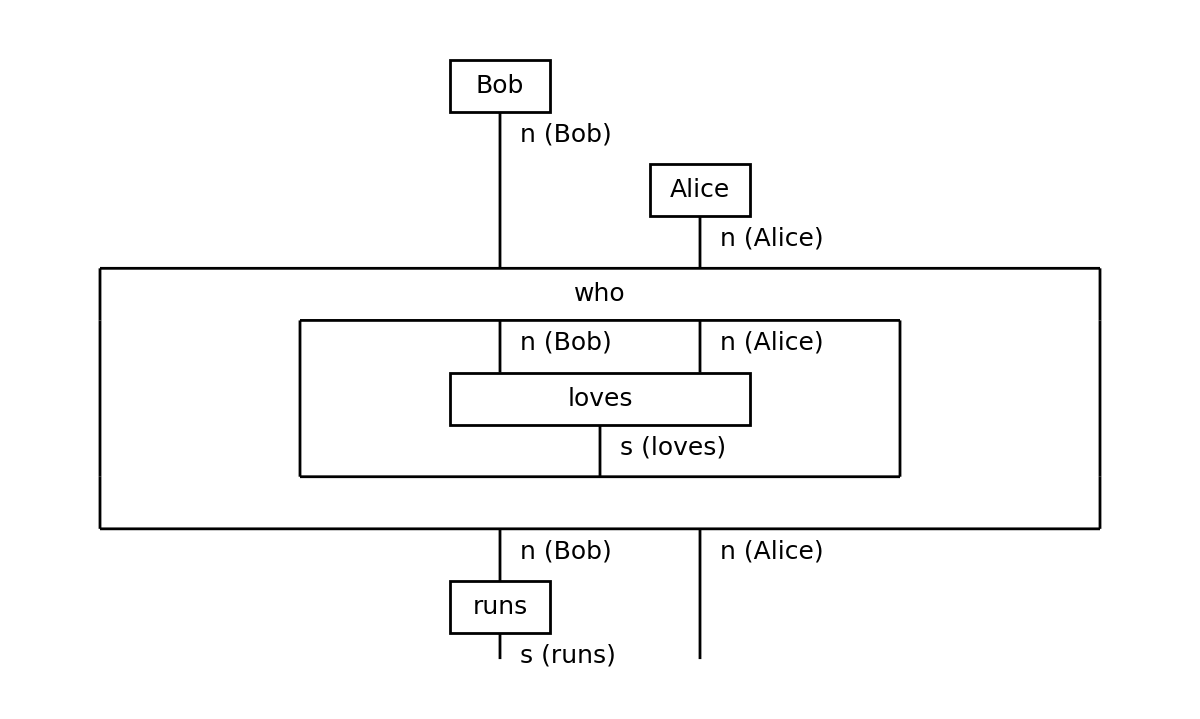}
\end{center}
Calling $s$-expansion on this term will result in the final text circuit.

All the cases where $n$-type expansion is triggered are similar in spirit to this example.
More precisely, the actual implementation of $n$-type expansion is essentially just a standard recursion on the $\lambda$-tree, except for the case where an $n$-type component is expanded and we have to modify the structure of the diagram in the manner described above.
This case is triggered when an $n$-type argument term is changed during recursion to something with type $n\times \ldots \times n$. 
If this happens, then the function term essentially needs to be parallel composed with an appropriate number of identities, before it and the $n$-expanded argument can be composed together.
Some swaps may also need to be applied to the $n$-expanded argument.

\subsection{Noun-coordination expansion}
\label{ssec:coord_expansion}

The final operation we discuss that recursively modifies the structure of the $\lambda$-tree deals with coordinating conjunctions (e.g. \texttt{and}, \texttt{or}) that combine two or more noun phrases into a single noun phrase.
This variety of coordinating conjunction was not considered in~\cite{wang2023distilling}, which mainly discussed coordinating conjunctions of entire sentences.
Note that in the pipeline, this step is performed before type expansion.

Like dragging out, noun-coordination expansion involves recursively doing surgery on the $\lambda$-tree.
That is, we need to slice the tree into parts and reattach the parts back together in a new way, while changing the type of some components.
However, in this case the surgery is not neatly encapsulated by a simple combinator, and we omit the technical details.

To see why coordinating conjunctions of nouns present a dilemma, consider
$$\texttt{Alice and Bob walk}.$$
The raw CCG parse gives the $\lambda$-term
\begin{center}
	\begin{minipage}{0.49\linewidth}
		\[\begin{tikzpicture}
	\begin{pgfonlayer}{nodelayer}
		\node [style=none] (0) at (3, 1.375) {\texttt{Alice}};
		\node [style=none] (1) at (3, 1) {$\scriptstyle n$};
		\node [style=none] (2) at (0, 1.25) {$@$};
		\node [style=none] (3) at (1.5, 2.375) {\texttt{Bob}};
		\node [style=none] (4) at (-1.125, 1.875) {};
		\node [style=none] (5) at (-0.375, 1.375) {};
		\node [style=none] (6) at (0.375, 1.375) {};
		\node [style=none] (7) at (1.125, 1.875) {};
		\node [style=none] (8) at (1.5, 2) {$\scriptstyle n$};
		\node [style=none] (9) at (0, 0.875) {$\scriptstyle n \to n$};
		\node [style=none] (10) at (1.5, 0.125) {$@$};
		\node [style=none] (11) at (-1.5, 2.375) {\texttt{and}};
		\node [style=none] (12) at (0.375, 0.75) {};
		\node [style=none] (13) at (1.125, 0.25) {};
		\node [style=none] (14) at (1.875, 0.25) {};
		\node [style=none] (15) at (2.625, 0.75) {};
		\node [style=none] (16) at (-1.5, 2) {$\scriptstyle n\to n\to n$};
		\node [style=none] (17) at (1.5, -0.25) {$\scriptstyle n$};
		\node [style=none] (18) at (-1.5, 0.125) {\texttt{walk}};
		\node [style=none] (19) at (-1.5, -0.25) {$\scriptstyle n\to s$};
		\node [style=none] (20) at (0, -1) {$@$};
		\node [style=none] (21) at (1.125, -0.375) {};
		\node [style=none] (22) at (0.375, -0.875) {};
		\node [style=none] (23) at (-0.375, -0.875) {};
		\node [style=none] (24) at (-1.125, -0.375) {};
		\node [style=none] (25) at (0, -1.375) {$\scriptstyle s$};
	\end{pgfonlayer}
	\begin{pgfonlayer}{edgelayer}
		\draw (4.center) to (5.center);
		\draw (7.center) to (6.center);
		\draw (12.center) to (13.center);
		\draw (15.center) to (14.center);
		\draw (21.center) to (22.center);
		\draw (24.center) to (23.center);
	\end{pgfonlayer}
\end{tikzpicture}\]
	\end{minipage}
	\begin{minipage}{0.49\linewidth}
	\includegraphics[width=\textwidth]{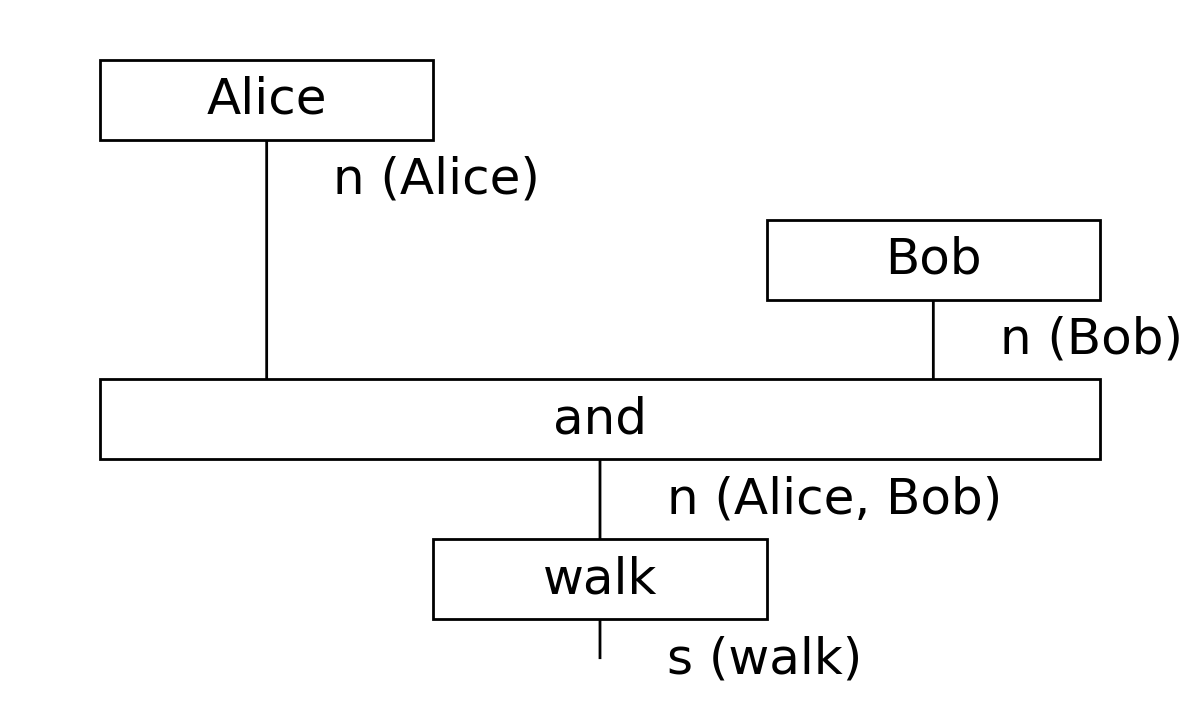}
	\end{minipage}
\end{center}
The noun phrase \texttt{Alice and Bob} has two semantic heads (\texttt{Alice}, \texttt{Bob}) corresponding to the two noun referents.
If we were to do our previously described $s$-expansion algorithm directly on this diagram, the $s$ would be replaced by a single $n$ wire co-indexed with both \texttt{Alice} and \texttt{Bob}, and we would be unable to refer to them separately.

The solution we implement involves rewriting the above diagram into the following diagram
\begin{center}
	\begin{minipage}{0.49\linewidth}
		\[\begin{tikzpicture}
	\begin{pgfonlayer}{nodelayer}
		\node [style=none] (0) at (5, 0.125) {\texttt{Alice}};
		\node [style=none] (1) at (5, -0.25) {$\scriptstyle n$};
		\node [style=none] (2) at (0.5, 1.25) {$@$};
		\node [style=none] (3) at (3.5, 1.25) {\texttt{Bob}};
		\node [style=none] (4) at (-0.625, 1.875) {};
		\node [style=none] (5) at (0.125, 1.375) {};
		\node [style=none] (6) at (0.875, 1.375) {};
		\node [style=none] (7) at (1.625, 1.875) {};
		\node [style=none] (8) at (3.5, 0.875) {$\scriptstyle n$};
		\node [style=none] (9) at (0.5, 0.875) {$\scriptstyle n \to n\to s$};
		\node [style=none] (10) at (2, 0.125) {$@$};
		\node [style=none] (11) at (-1, 2.375) {\texttt{and}};
		\node [style=none] (12) at (0.875, 0.75) {};
		\node [style=none] (13) at (1.625, 0.25) {};
		\node [style=none] (14) at (2.375, 0.25) {};
		\node [style=none] (15) at (3.125, 0.75) {};
		\node [style=none] (16) at (-1, 2) {$\scriptstyle (n\to s)\to n\to n\to s$};
		\node [style=none] (17) at (2, -0.25) {$\scriptstyle n\to s$};
		\node [style=none] (18) at (2, 2.375) {\texttt{walk}};
		\node [style=none] (19) at (2, 2) {$\scriptstyle n\to s$};
		\node [style=none] (20) at (3.5, -1) {$@$};
		\node [style=none] (21) at (2.375, -0.375) {};
		\node [style=none] (22) at (3.125, -0.875) {};
		\node [style=none] (23) at (3.875, -0.875) {};
		\node [style=none] (24) at (4.625, -0.375) {};
		\node [style=none] (25) at (3.5, -1.375) {$\scriptstyle s$};
	\end{pgfonlayer}
	\begin{pgfonlayer}{edgelayer}
		\draw (4.center) to (5.center);
		\draw (7.center) to (6.center);
		\draw (12.center) to (13.center);
		\draw (15.center) to (14.center);
		\draw (21.center) to (22.center);
		\draw (24.center) to (23.center);
	\end{pgfonlayer}
\end{tikzpicture}\]
	\end{minipage}
	\begin{minipage}{0.49\linewidth}
		\includegraphics[width=\textwidth]{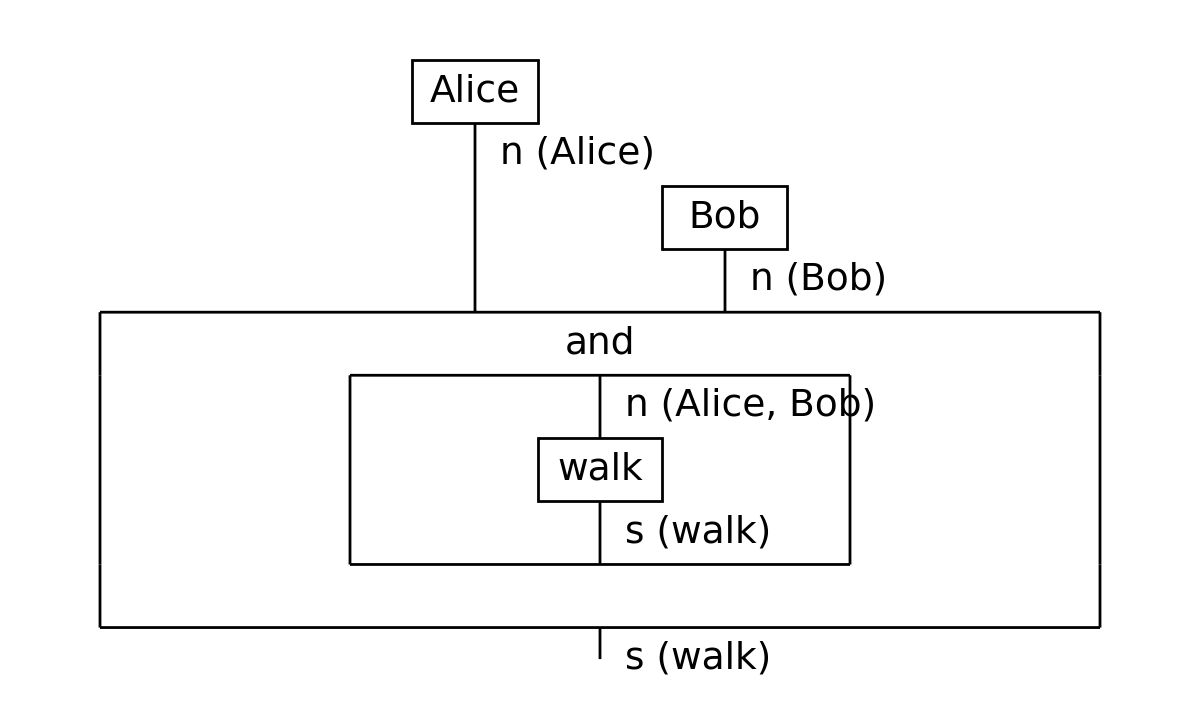}
	\end{minipage}
\end{center}
That is, we have modified the type of the noun-coordinating-conjunction \texttt{and}, promoting it to a frame that wraps around the relevant gate.
Thus the structure of the $\lambda$-tree has also changed.
The types of other constants are unchanged.
With this diagram, applying $s$-expansion would yield two $n$ wires at the bottom as desired: one for \texttt{Alice} and one for \texttt{Bob}.

In one regard, this noun-coordination expansion resembles $n$-type expansion, in that a triggering condition is the presence of an $n$-type wire carrying a noun phrase that contains multiple noun referents.
What distinguishes the noun phrases involved here and the noun phrases involved in $n$-type expansion is the number of head indices carried in the $n$-type wire.
When we see a noun phrase like \texttt{Bob who liked Alice}, the head of this noun phrase is still just \texttt{Bob}, in which case we do $n$-type expansion.
However, in a coordinating conjunction, the two conjuncts have equal weight, and so a noun phrase like \texttt{Alice and Bob} has two heads: \texttt{Alice} and \texttt{Bob}.

A $\lambda$-tree undergoes noun-coordination expansion if the $\lambda$-term corresponding to a noun phrase argument $N$ satisfies all four conditions: 
\begin{itemize}
	\item the argument term $N$ is $n$-type: in the previous example, $N\equiv \texttt{and}(\texttt{Bob})(\texttt{Charlie})$ is of $n$-type
	\item the argument term $N$ is not a constant or a variable. 
	This is not a redundant check, because sometimes variables may inherit multiple co-indices.
	\item the argument term $N$ has multiple head co-indices: in the example, the $n$-type of $N$ has co-indices \texttt{Alice} and \texttt{Bob}
	\item the inner most function of the argument term (i.e. the left-most leaf) should not be higher-order.
	This inner most function should correspond to the coordinating conjunction itself, and if it is already higher-order then it has already undergone expansion and should be left. 
	In the example, the inner most function would be \texttt{and}, which has the 1st-order type $n\to n\to n$
\end{itemize}




As with the previous operations, noun-coordination expansion must work recursively, due to the potential for arbitrary nesting in natural language.
A consequence of our choice of implementation is that we can obtain arbitrarily large noun-coordination frames due to nesting.
Consider
$$\texttt{Alice, Bob, Claire and Dave walk}.$$
Our CCG parser will treat each comma \texttt{,} as a separate token which functions like an \texttt{and}.
That is, the CCG parse returns the $\lambda$-term
\begin{center}
	\begin{minipage}{0.49\linewidth}
		\[\begin{tikzpicture}
	\begin{pgfonlayer}{nodelayer}
		\node [style=none] (0) at (0.5, 0.125) {\texttt{,}};
		\node [style=none] (1) at (0.5, -0.25) {$\scriptstyle n\to n\to n$};
		\node [style=none] (2) at (2, 1.25) {$@$};
		\node [style=none] (3) at (5, 1.25) {\texttt{Claire}};
		\node [style=none] (4) at (0.875, 1.875) {};
		\node [style=none] (5) at (1.625, 1.375) {};
		\node [style=none] (6) at (2.375, 1.375) {};
		\node [style=none] (7) at (3.125, 1.875) {};
		\node [style=none] (8) at (5, 0.875) {$\scriptstyle n$};
		\node [style=none] (9) at (2, 0.875) {$\scriptstyle n \to n$};
		\node [style=none] (10) at (3.5, 0.125) {$@$};
		\node [style=none] (11) at (0.5, 2.375) {\texttt{and}};
		\node [style=none] (12) at (2.375, 0.75) {};
		\node [style=none] (13) at (3.125, 0.25) {};
		\node [style=none] (14) at (3.875, 0.25) {};
		\node [style=none] (15) at (4.625, 0.75) {};
		\node [style=none] (16) at (0.5, 2) {$\scriptstyle n\to n\to n$};
		\node [style=none] (17) at (3.5, -0.25) {$\scriptstyle n$};
		\node [style=none] (18) at (3.5, 2.375) {\texttt{Dave}};
		\node [style=none] (19) at (3.5, 2) {$\scriptstyle n$};
		\node [style=none] (20) at (2, -1) {$@$};
		\node [style=none] (21) at (3.125, -0.375) {};
		\node [style=none] (22) at (2.375, -0.875) {};
		\node [style=none] (23) at (1.625, -0.875) {};
		\node [style=none] (24) at (0.875, -0.375) {};
		\node [style=none] (25) at (2, -1.375) {$\scriptstyle n\to n$};
		\node [style=none] (26) at (5, -1.125) {\texttt{Bob}};
		\node [style=none] (27) at (5, -1.5) {$\scriptstyle n$};
		\node [style=none] (28) at (3.5, -2.25) {$@$};
		\node [style=none] (29) at (2.375, -1.625) {};
		\node [style=none] (30) at (3.125, -2.125) {};
		\node [style=none] (31) at (3.875, -2.125) {};
		\node [style=none] (32) at (4.625, -1.625) {};
		\node [style=none] (33) at (3.5, -2.625) {$\scriptstyle n$};
		\node [style=none] (34) at (0.5, -2.375) {\texttt{,}};
		\node [style=none] (35) at (0.5, -2.75) {$\scriptstyle n\to n\to n$};
		\node [style=none] (36) at (2, -3.5) {$@$};
		\node [style=none] (37) at (3.125, -2.875) {};
		\node [style=none] (38) at (2.375, -3.375) {};
		\node [style=none] (39) at (1.625, -3.375) {};
		\node [style=none] (40) at (0.875, -2.875) {};
		\node [style=none] (41) at (2, -3.875) {$\scriptstyle n\to n$};
		\node [style=none] (42) at (5, -3.625) {\texttt{Alice}};
		\node [style=none] (43) at (5, -4) {$\scriptstyle n$};
		\node [style=none] (44) at (3.5, -4.75) {$@$};
		\node [style=none] (45) at (2.375, -4.125) {};
		\node [style=none] (46) at (3.125, -4.625) {};
		\node [style=none] (47) at (3.875, -4.625) {};
		\node [style=none] (48) at (4.625, -4.125) {};
		\node [style=none] (49) at (3.5, -5.125) {$\scriptstyle n$};
		\node [style=none] (50) at (0.5, -4.875) {\texttt{walk}};
		\node [style=none] (51) at (0.5, -5.25) {$\scriptstyle n\to s$};
		\node [style=none] (52) at (2, -6) {$@$};
		\node [style=none] (53) at (3.125, -5.375) {};
		\node [style=none] (54) at (2.375, -5.875) {};
		\node [style=none] (55) at (1.625, -5.875) {};
		\node [style=none] (56) at (0.875, -5.375) {};
		\node [style=none] (57) at (2, -6.375) {$\scriptstyle s$};
	\end{pgfonlayer}
	\begin{pgfonlayer}{edgelayer}
		\draw (4.center) to (5.center);
		\draw (7.center) to (6.center);
		\draw (12.center) to (13.center);
		\draw (15.center) to (14.center);
		\draw (21.center) to (22.center);
		\draw (24.center) to (23.center);
		\draw (29.center) to (30.center);
		\draw (32.center) to (31.center);
		\draw (37.center) to (38.center);
		\draw (40.center) to (39.center);
		\draw (45.center) to (46.center);
		\draw (48.center) to (47.center);
		\draw (53.center) to (54.center);
		\draw (56.center) to (55.center);
	\end{pgfonlayer}
\end{tikzpicture}\]
	\end{minipage}
	\begin{minipage}{0.49\linewidth}
	\includegraphics[width=\textwidth]{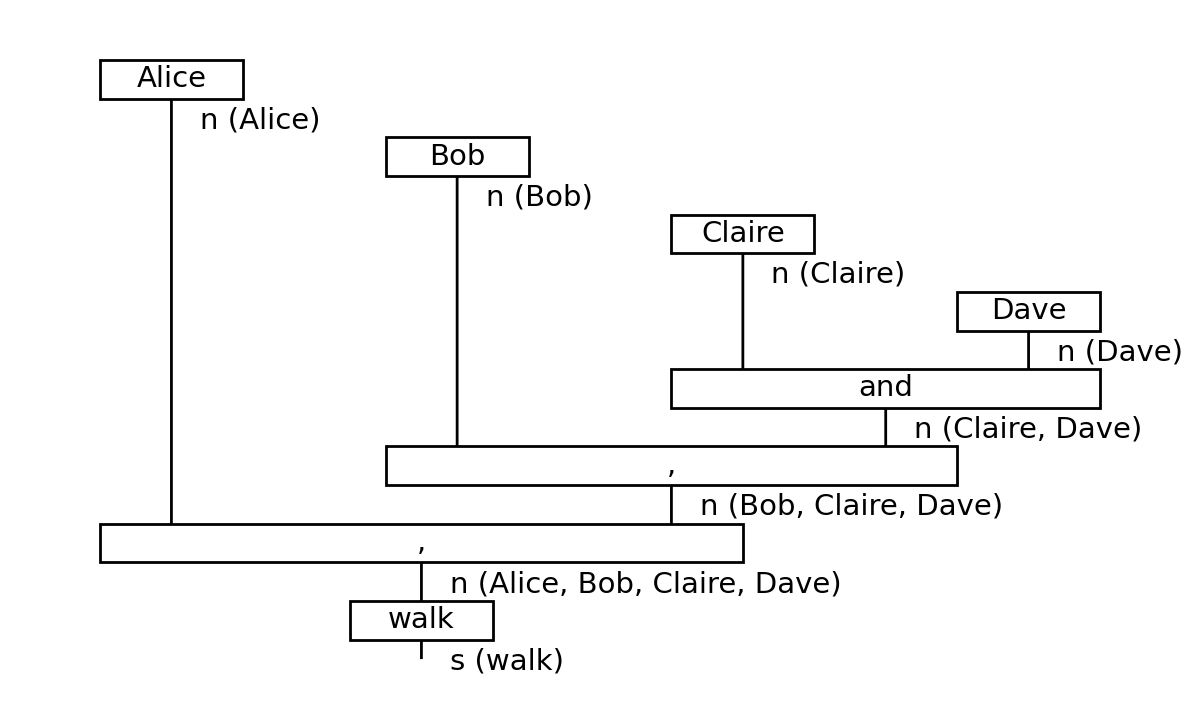}
	\end{minipage}
\end{center}
and the noun-coordination expansion acts on this term recursively to return
\begin{center}
	\includegraphics[width=0.8\textwidth]{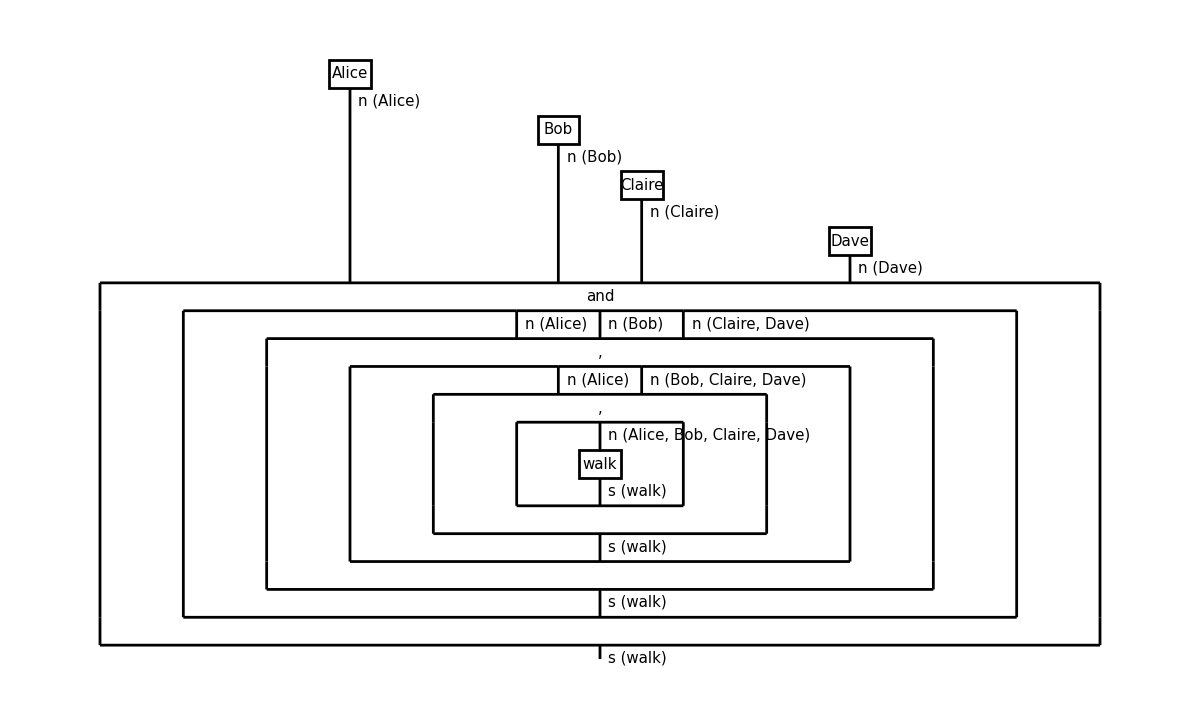}
\end{center}
Here we have a series of noun-coordinating-conjunction frames that each merge the two right-most input noun wires into one.
After applying $s$-expansion to this, the outputs of the frames will do the opposite of this, i.e. split the right-most wire into two wires.

\subsection{Sentence composition}
\label{sec:sentence-composition}
Having created the text circuits for individual sentences, we stitch them together to form the circuit of the entire text, following the procedure outlined in~\cite{coecke2021mathematics}. In this, wires referring to the same discourse referent are matched up, and the sentences are sequentially composed. For example, the two sentences \texttt{Alice likes Bob} and \texttt{He is funny} have the following circuits

\begin{center}
	  \includegraphics[width=0.45\linewidth]{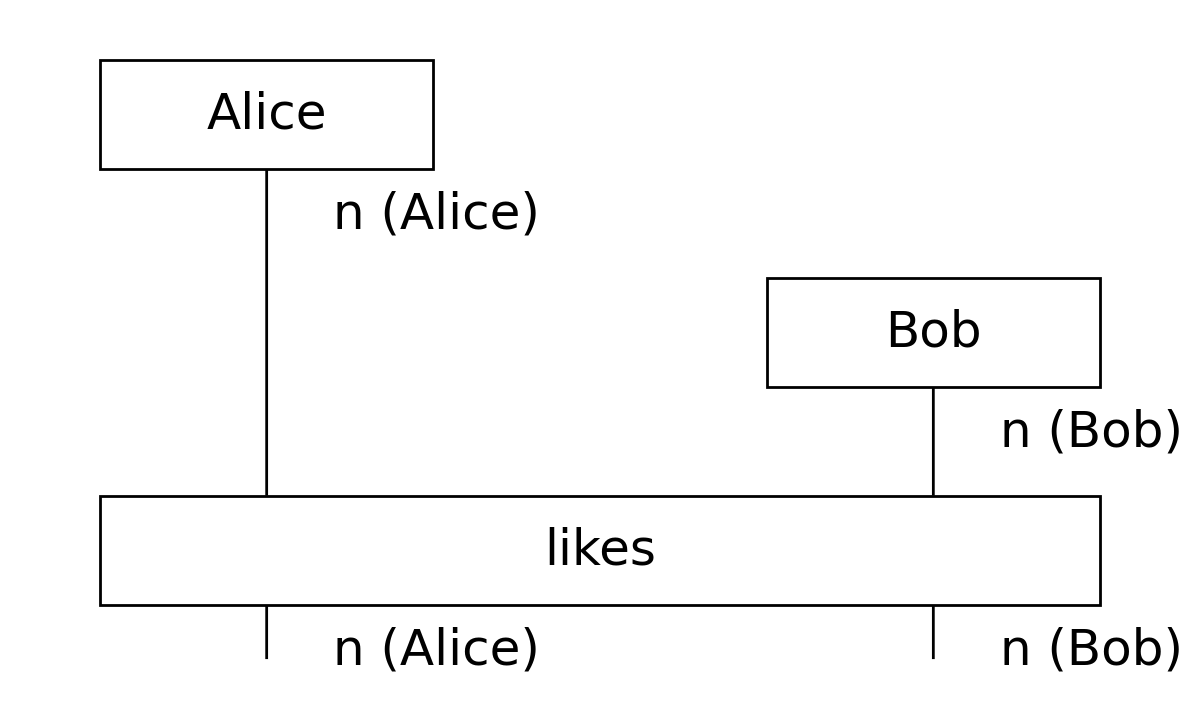}
	  \includegraphics[width=0.45\linewidth]{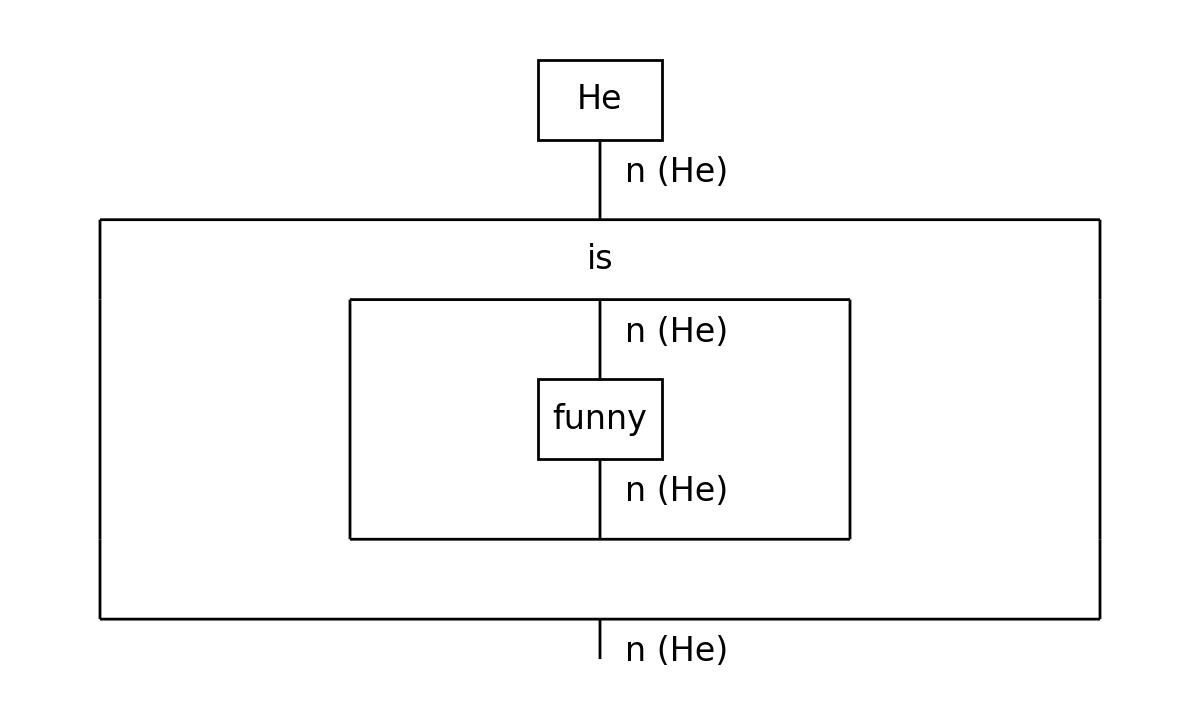}
\end{center}

\noindent When composing them, these two sentences form the circuit 

\begin{center}
	\includegraphics[width=0.55\linewidth]{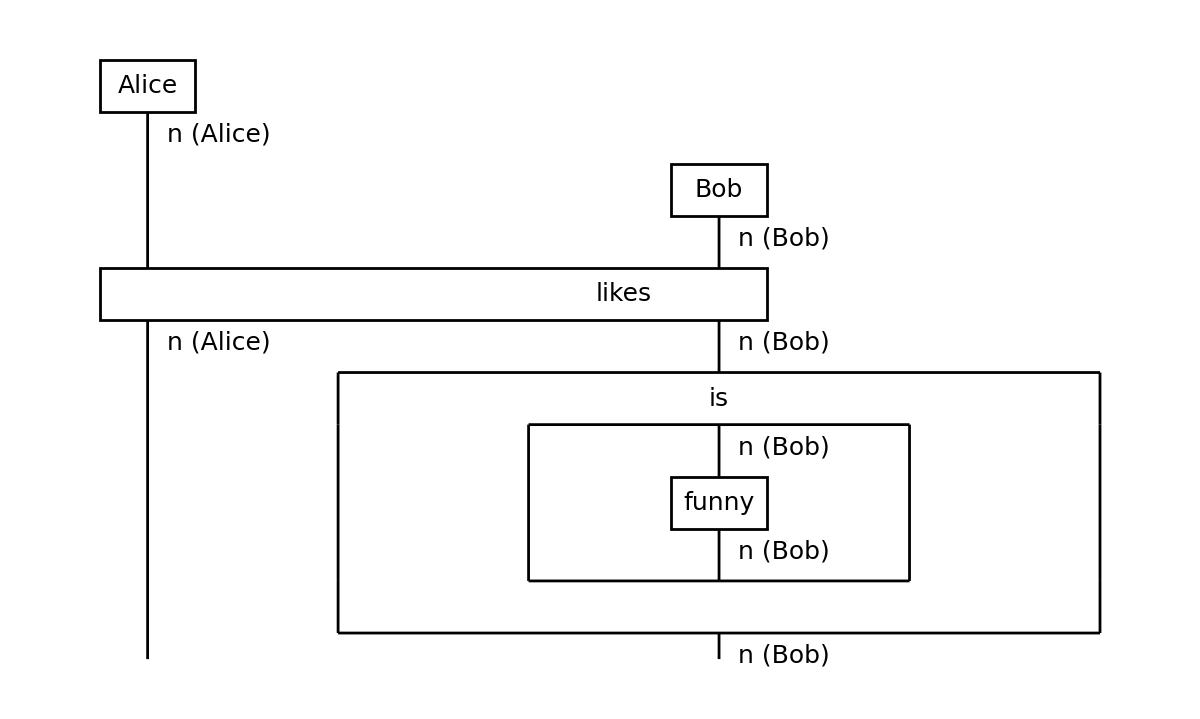}
\end{center}

\noindent For this, the pipeline uses the coreference parser provided by spacy~\cite{coreferee} to resolve coreferences. To match the wires correctly, the sentence composition may swap wires, which corresponds to the trivial shuffeling around of information. This can for example be seen in the example in the introduction.

During this composition, any sentence only updates the discourse referents it mentions. Thus, when a sentence does not mention a referent already present in the text, it does not update the referent. Consider, for example, \texttt{Alice}, whom the second sentence in the example in the example above --- \texttt{He is funny} ---  does not mention. For each discourse referent not affected by the sentence, we parallel compose the circuit of the sentence with an empty wire, i.e. the identity. Therefore, when composing the text circuit with this extended circuit, the composition still type-matches and only updates the mentioned referents. In essence, all wires not updated by a later sentence are simply made longer using the identity.

Using this method of composition, we can deal with discourse referents that are mentioned at most once. However, in some sentences, discourse referents can be mentioned multiple times, leading them to be represented by multiple wires. To compose these sentences, we introduce noun contraction.

\subsubsection{Noun contraction}
\label{sec:noun-contraction}
Some sentences introduce multiple wires for a single discourse referent. For example, this can occur through multiple mentions of an actor within a sentence, e.g. \texttt{Bob and Dave think Bob is funny} where there are two wires for \texttt{Bob}. Similarly, coreference can introduce multiple wires for one referent, e.g. \texttt{Bob thinks he is smart} where \texttt{Bob} and \texttt{he} each introduce one wire for the same referent
\begin{center}
	\includegraphics[width=0.55\linewidth]{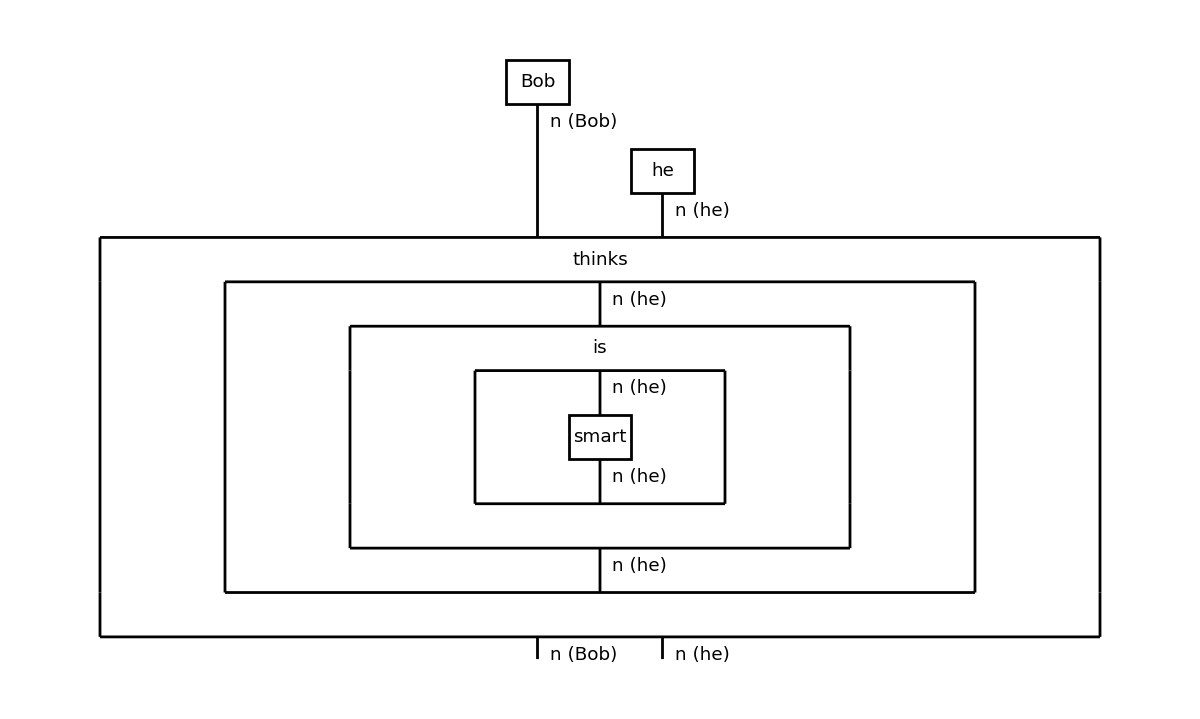}
\end{center}
However, DisCoCirc proposes one wire per discourse referent. Therefore, we introduce the noun-contraction. 

The noun contraction introduces a new frame that copies the wire in question, applies the original sentence and finally merges the two wires back into one. For example, for the sentence \texttt{Bob thinks he is smart}, the frame splits the \texttt{Bob} wire, feeds them into \texttt{Bob} and \texttt{he}, and, at the end of the sentence, merge the two wires again. Therefore, the sentence becomes a process that updates one wire --- namely \texttt{Bob}'s --- with the information that he thinks he is smart. We get
\begin{center}
	\includegraphics[width=0.55\linewidth]{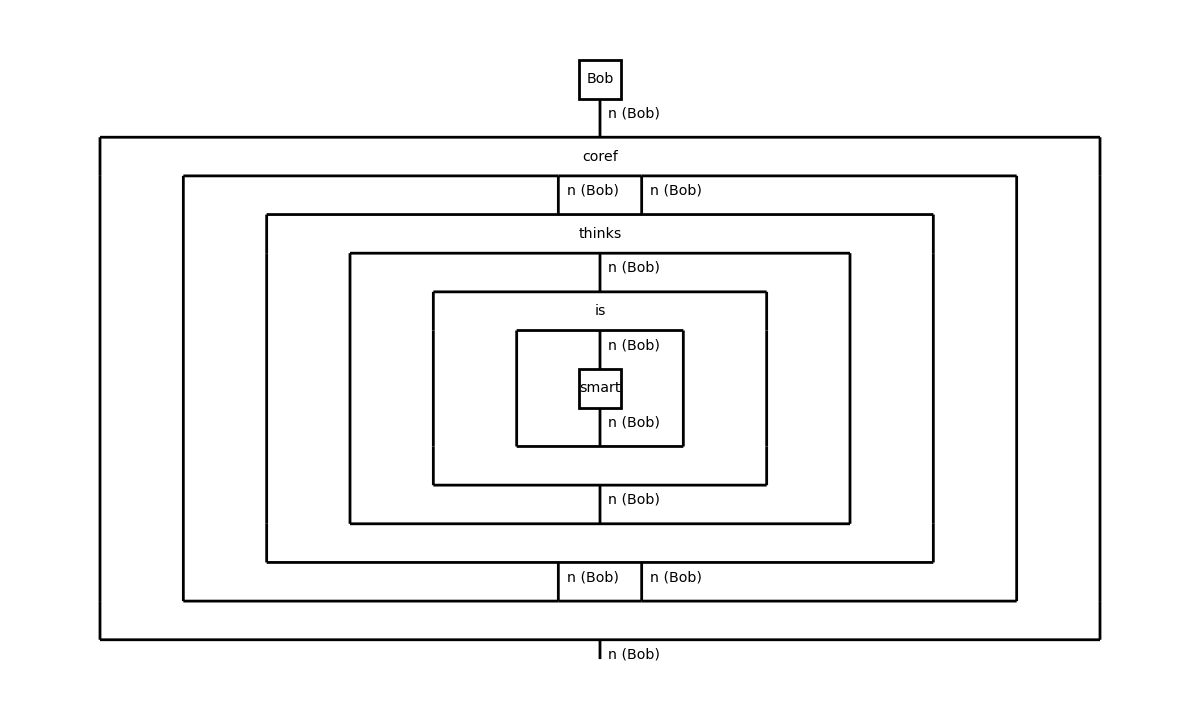}
\end{center}
We are able to identify which noun should appear at the top of the resulting diagram (in this case \texttt{Bob} rather than \texttt{he}), since the coreference resolution identifies a `most specific' mention of an entity.
This most specific mention will, for instance, be a proper names rather than a pronoun.

Where the noun contraction introduces a new \texttt{coref}-frame. 
This can be seen as somewhat dual to the noun-coordination expansion, which introduces a new frame to compress two wires into one to eventually split them again.

The procedure can be recursively applied to any arbitrary number of wires for a referent. Therefore, we guarantee that all discourse referents are represented by at most one wire. As such, we can apply the sentence composition as explained above.

\subsection{Semantic rewrites}
\label{sec:semantic_rewrites}
Text circuits already capture some degree of semantics.
However, it is easy to impose more semantics on text circuits, by specifying rewrite rules that replace some components in the circuit diagrams with some other component.
Which semantic rewrites we ultimately choose to apply may depend on the task or context at hand.
For instance, one of the rewrites we propose below effectively deletes all instances of the verb \texttt{is}.
If, for instance, we deleted this verb in its past tense form \texttt{was}, we would lose the tense information it conveys.
So whether we want to apply this rewrite on \texttt{was} depends on whether the tense information is important to us.
The rewrites we discuss below are related to the `equations between text' discussed in previous papers~\cite{coecke2021grammar}, and indeed several of them are taken from~\cite{wang2023distilling}.
\\

\noindent \textbf{Deleting determiners:}
Determiners like \texttt{a} or \texttt{the} are parsed in CCG like adjectives, i.e. in our diagrams they are a $n\to n$ gate.
In the case where we do not care about them, we can just rewrite them into identities.
For instance, \texttt{Alice pets the cat} will yield the circuit on the left,
\begin{center}
	\raisebox{-.4\height}{
	\includegraphics[width=0.4\textwidth]{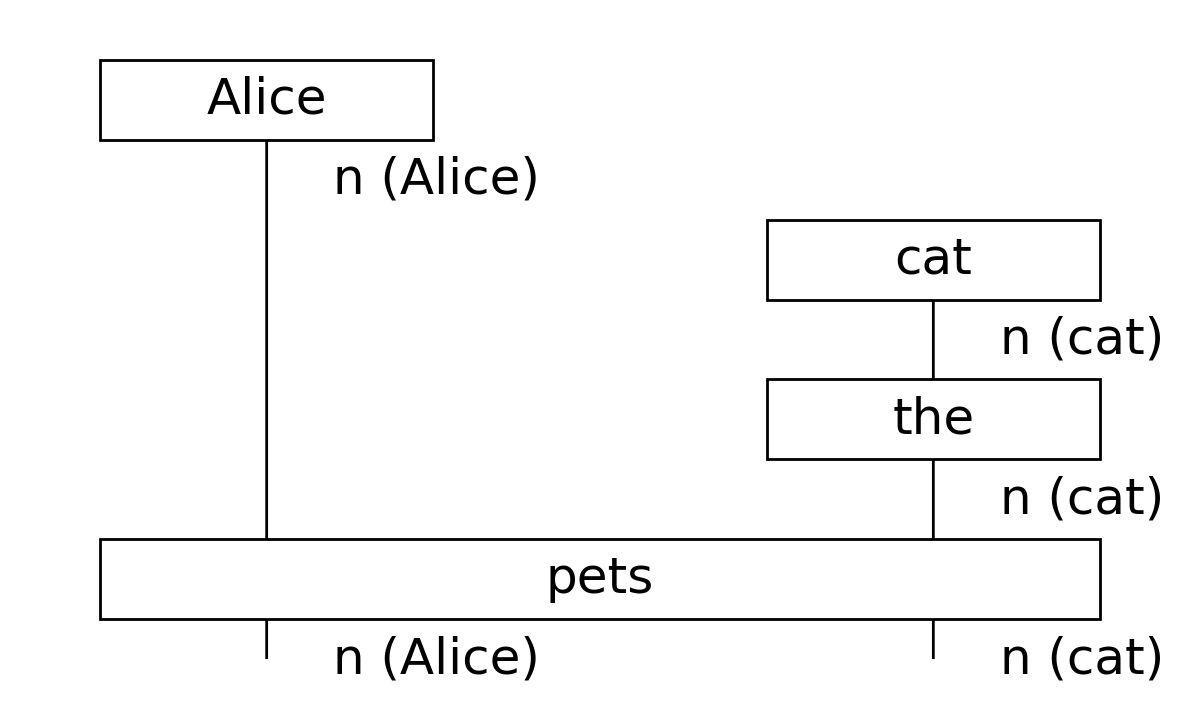}}
	$\rightsquigarrow$
	\raisebox{-.4\height}{
	\includegraphics[width=0.4\textwidth]{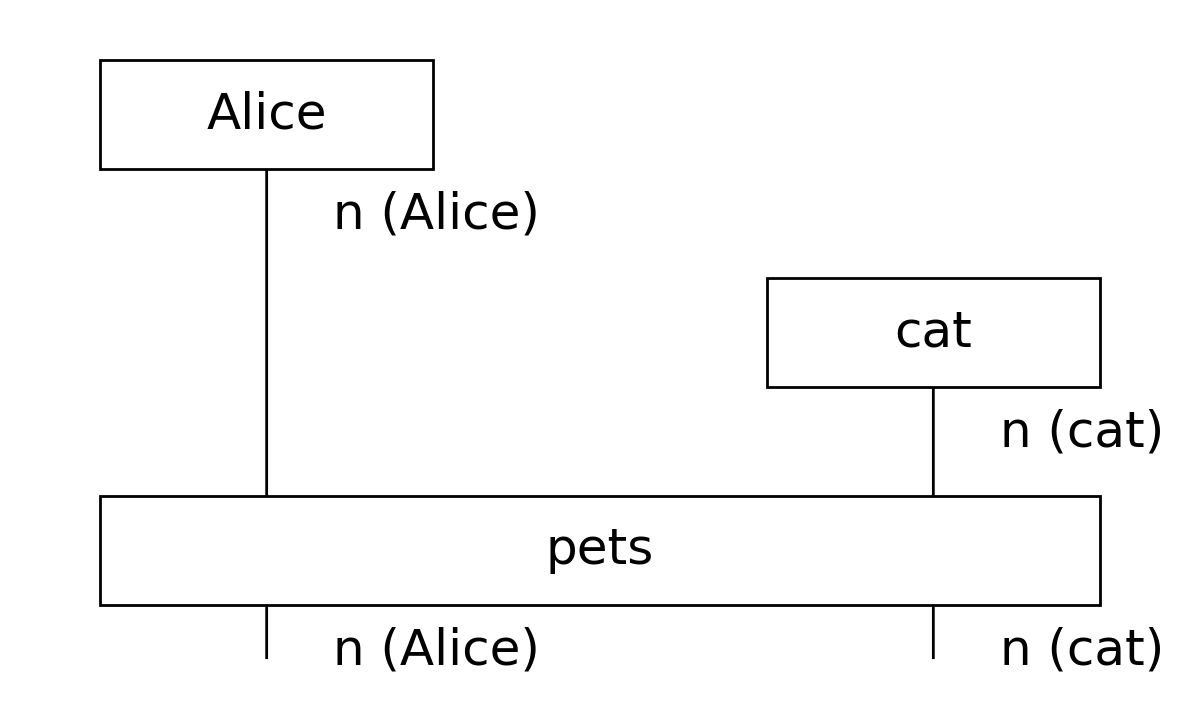}}
\end{center}
which becomes the circuit on the right if we decide that the \texttt{the} determiner here is semantically void.
The circuit on the right captures just as well that there is an \texttt{Alice} and there is a \texttt{cat}, and that the former pets the latter.
\\

\noindent \textbf{Is-deletion:}
The verb \texttt{is} (along with all its various forms) such as in \texttt{Alice is red}, is parsed as an $(n\to s)\to n\to s$ type in CCG.
That is, the CCG parse for a sentence like \texttt{Alice is red} will yield the diagram 
\begin{center}
	\includegraphics[width=0.4\textwidth]{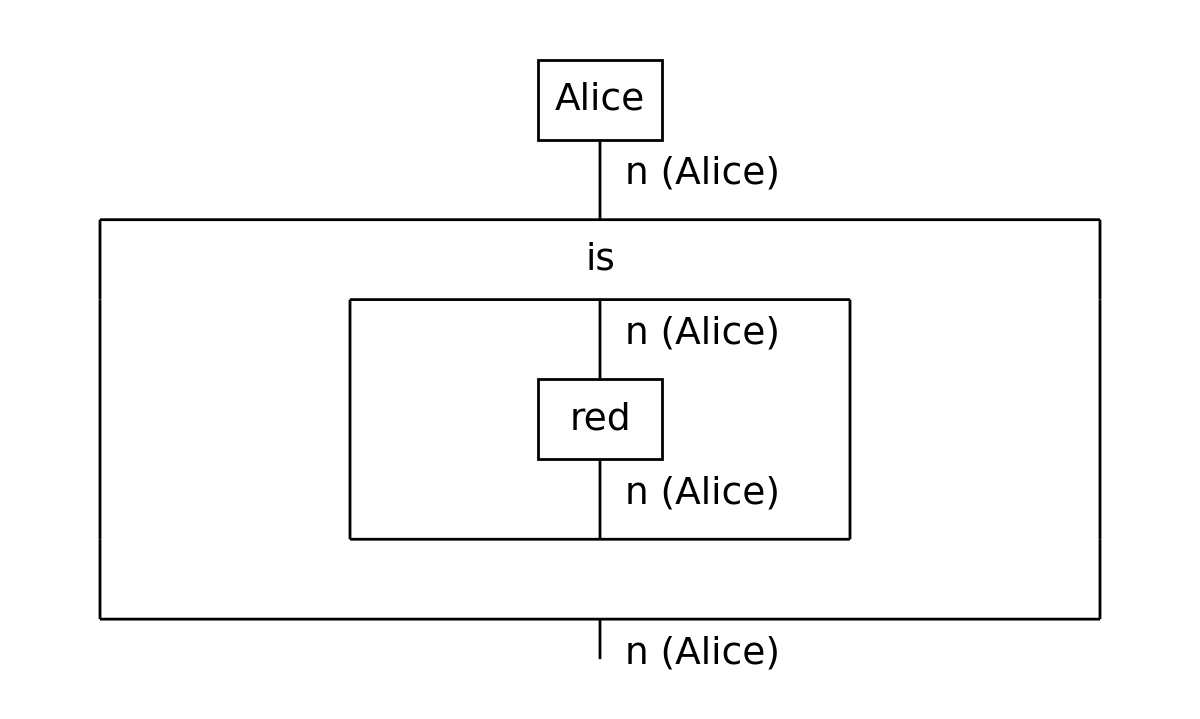}
\end{center}
which, after passing through the pipeline, will remain structurally the same, except with $s$ types replaced with $n$.
Rewriting the \texttt{is} frame into an identity would yield
\begin{center}
	\includegraphics[width=0.3\textwidth]{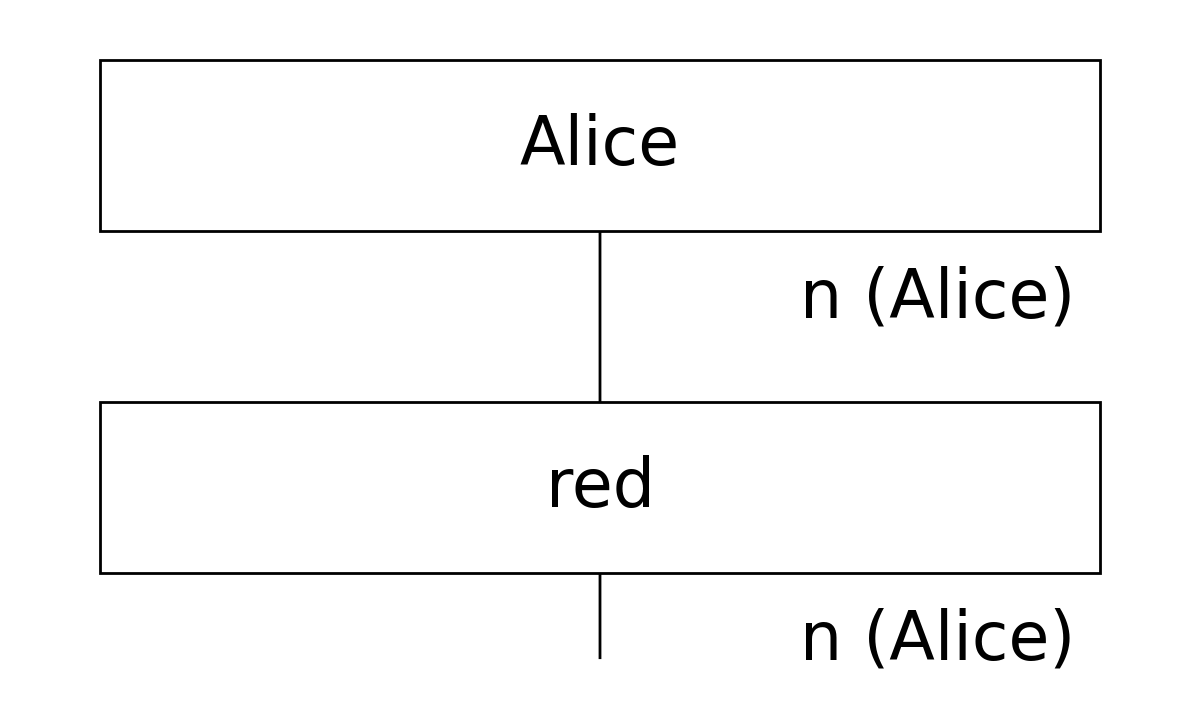}
\end{center}
which is the diagram that would be obtained from the noun phrase \texttt{red Alice}.
If one chooses to impose this equality, this rewrite embodies the assertion of a semantic equality between texts of the form \texttt{Alice is red} and those of the form \texttt{red Alice}.
\\

\noindent \textbf{Relative pronoun deletion:}
Another instance in which we may want to rewrite a higher-order frame into an identity, which also yields a kind of semantic equality between texts, is provided by relative pronouns (e.g. \texttt{who}, \texttt{that}).
Recall, for instance, the example \texttt{Bob who loves Alice runs} from Section~\ref{sssec:n_type_expansion}.
We may consider that, after $n$-type expansion, the relative pronoun has served its purpose, and contains no further semantic content.
Rewriting the \texttt{who} frame as an identity yields the circuit
\begin{center}
	\includegraphics[width=0.6\textwidth]{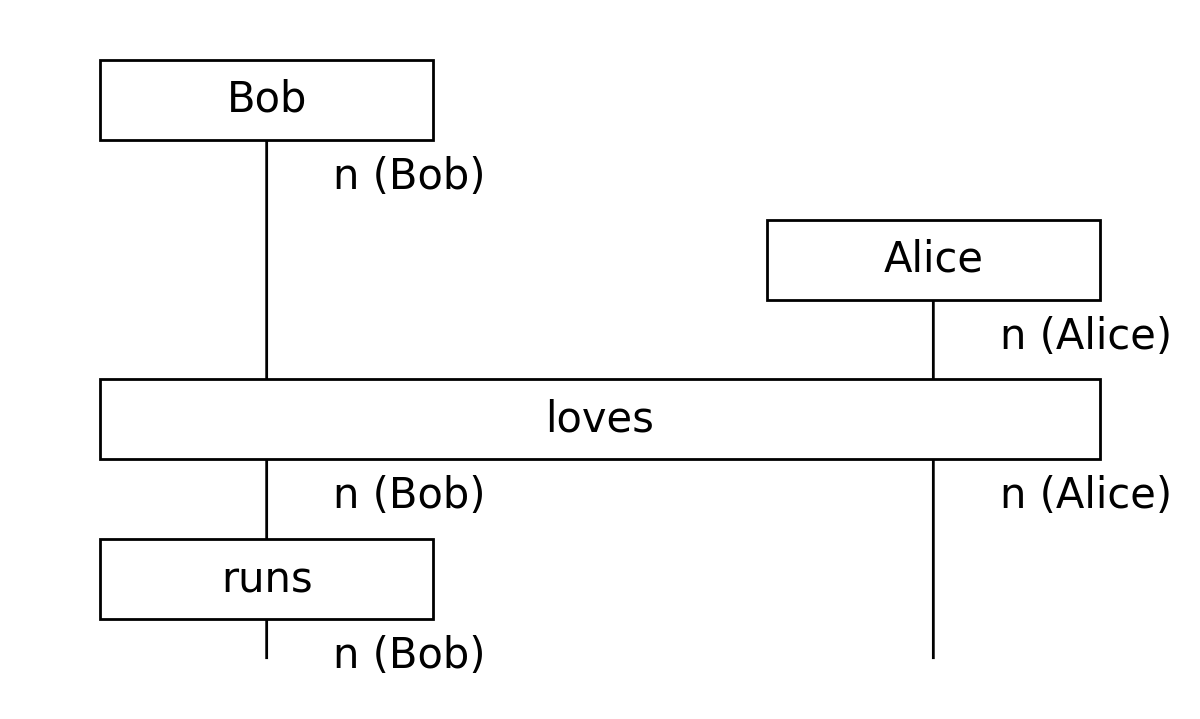}
\end{center}
This is equivalent to the circuit for the text 
$$\texttt{Bob loves Alice. Bob runs.}$$
which seems like a faithful representation of the semantic content of the original sentence.

It is worth mentioning that this rewrite was considered in the DisCoCat model~\cite{sadrzadeh2013frobenius}, and motivated the development of DisCoCirc.
\\

\noindent \textbf{Passive voice into active voice:}
The next semantic rewrite does not simply replace some component with an identity.
Consider a sentence with a passive voice construction:
\texttt{Alice is bored by the class}.
The diagram output by the pipeline for this would be
\begin{center}
	\includegraphics[width=0.6\textwidth]{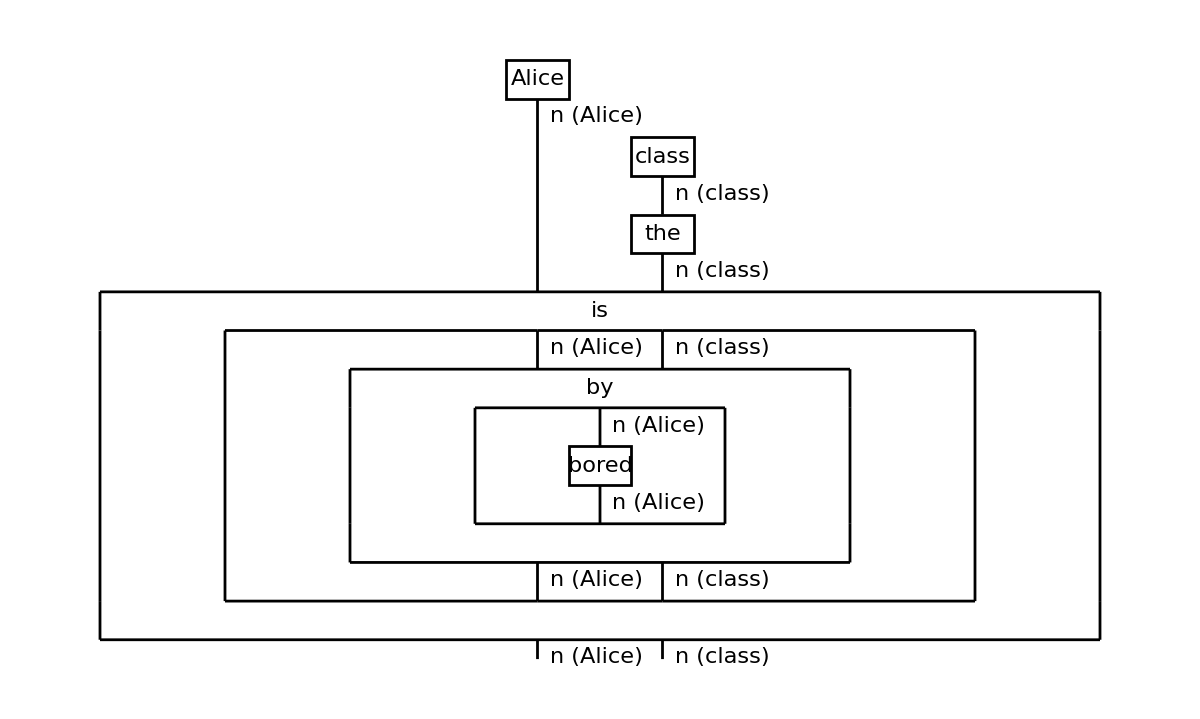}
\end{center}
which, after applying the \texttt{is} and \texttt{the} semantic rewrites described earlier, becomes
\begin{center}
	\includegraphics[width=0.6\textwidth]{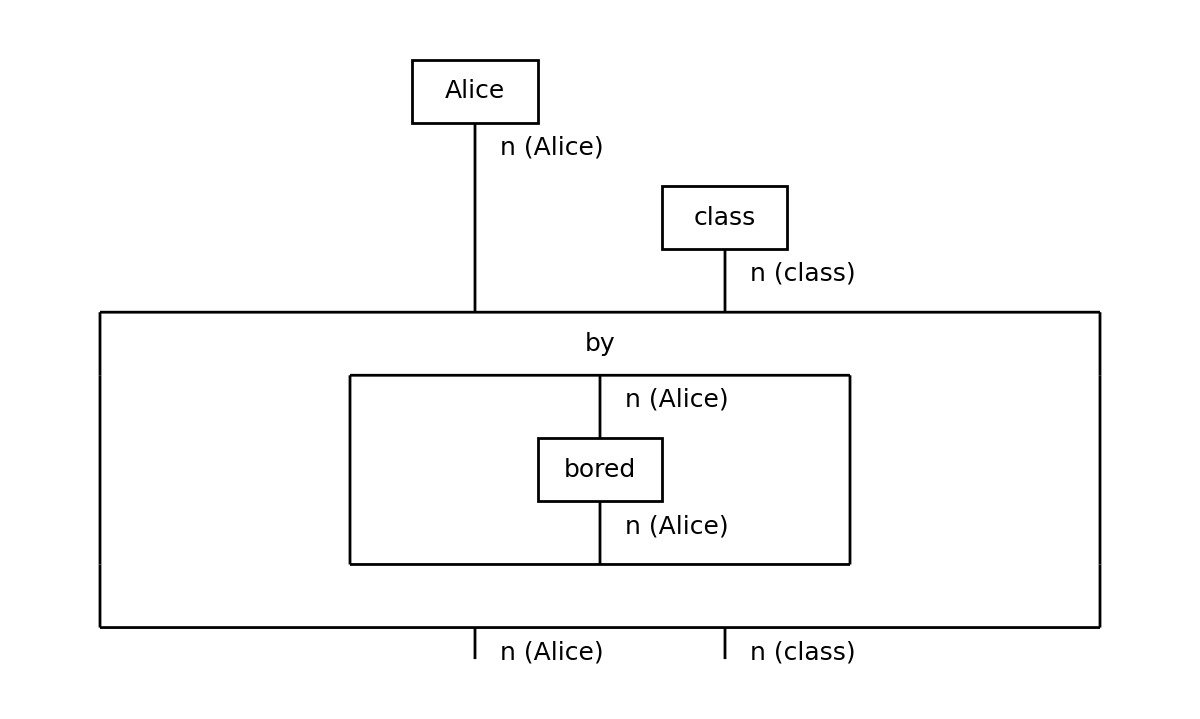}
\end{center}
Now, we assert \texttt{Alice is bored by the class} to be semantically equivalent to \texttt{The class bores Alice}.
To realize this, we rewrite the above \texttt{bored by} composite circuit component into a \texttt{bores} gate with swaps
\begin{center}
	\includegraphics[width=0.6\textwidth]{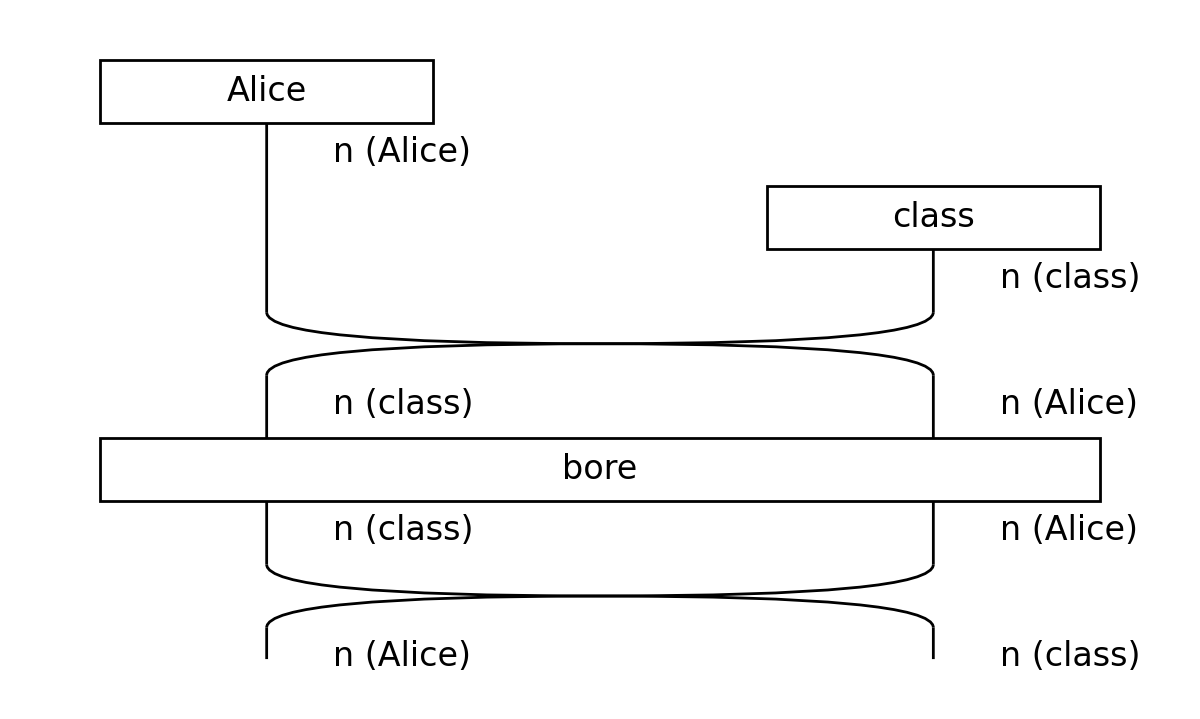}
\end{center}
The passive form of the verb can be changed into the active form via lemmatization.

\noindent \textbf{Possessive pronoun resolution:}
The final rewrite we consider incorporates information from the coreference resolver.
Besides the basic coreference taken into account in the sentence composition, the pipeline also deals with possessive pronouns. 
Take, for example, \texttt{Bob loves his dog}. 
\begin{center}
	\includegraphics[width=0.5\linewidth]{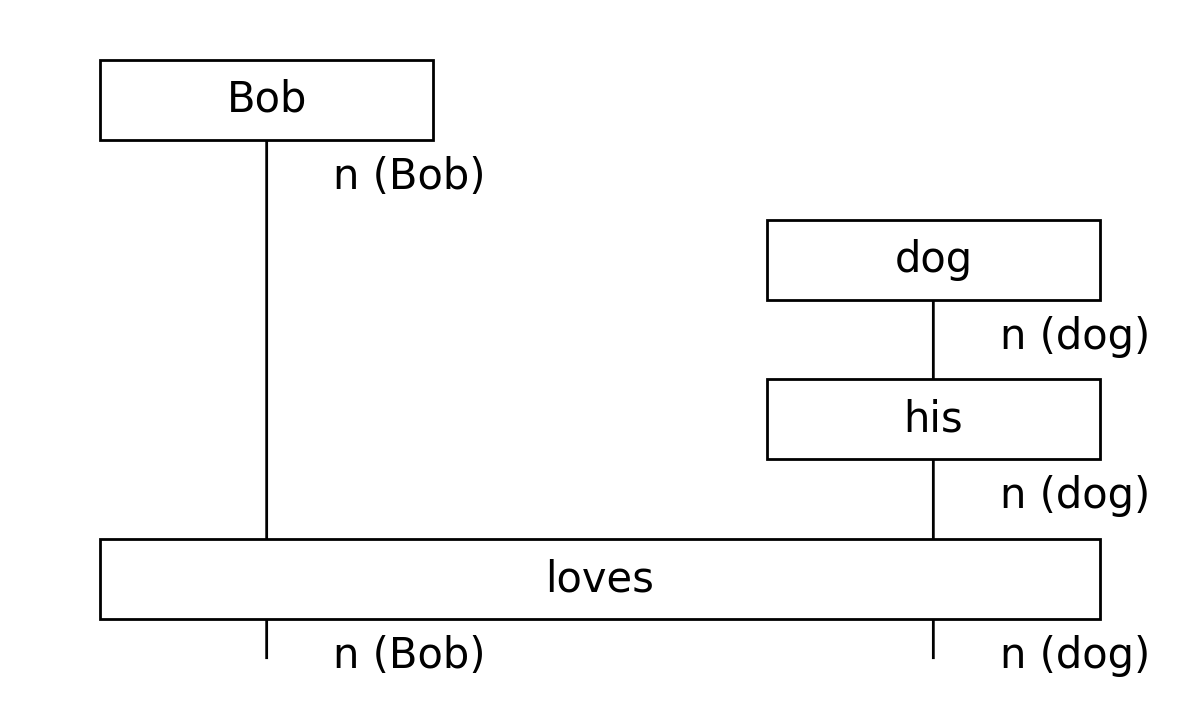}
\end{center}
Purely from the diagram, it is unclear who \texttt{his} refers to. Using a coreference resolver, we can add this additional information into the circuit. By changing the type of \texttt{his} to take in two nouns instead of one, we can feed both \texttt{Bob} and the \texttt{dog} into the same component, resulting in 
\begin{center}
	\includegraphics[width=0.5\linewidth]{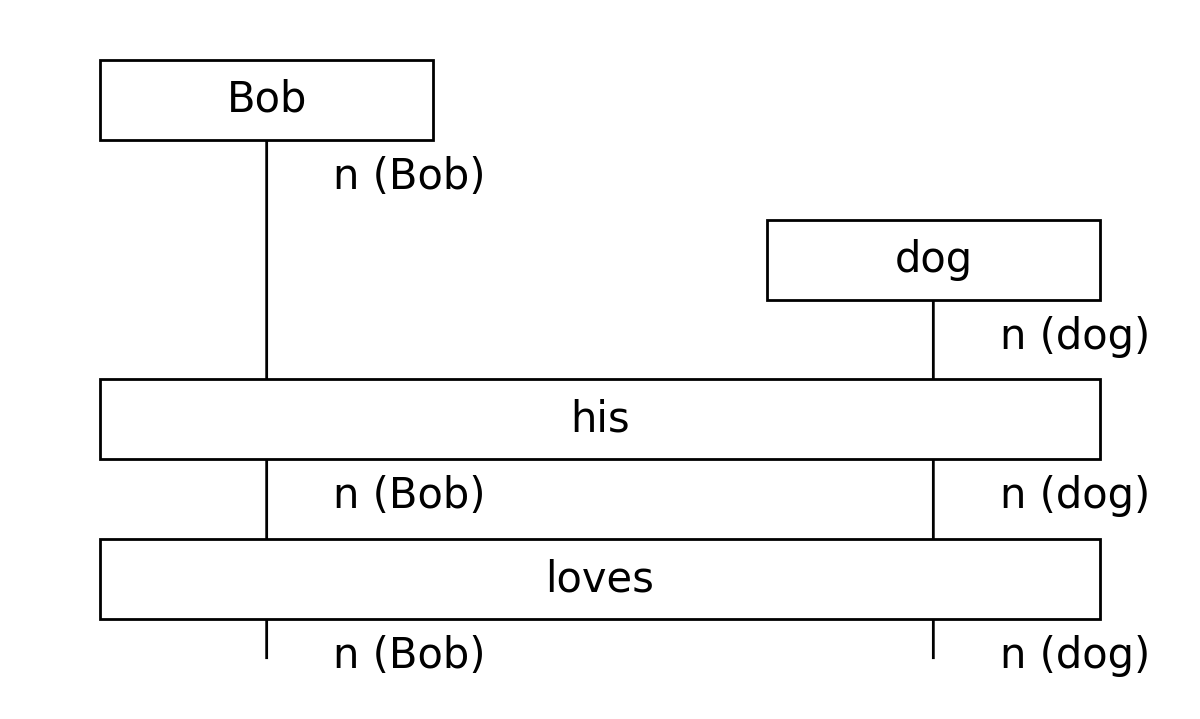}
\end{center}
This new circuit explicitly links \texttt{Bob} and his \texttt{dog} via the \texttt{his} box to indicate ownership. 
Given such possessive gates like \texttt{his}, the argument on the right is always the thing owned by the other arguments.

When multiple discourse referents are intended with the possessive pronouns, all of them are fed into the pronoun box. For example, the sentence \texttt{Alice, Bob and Dave like their dog} is rewritten to

\begin{center}
	\includegraphics[width=1\linewidth]{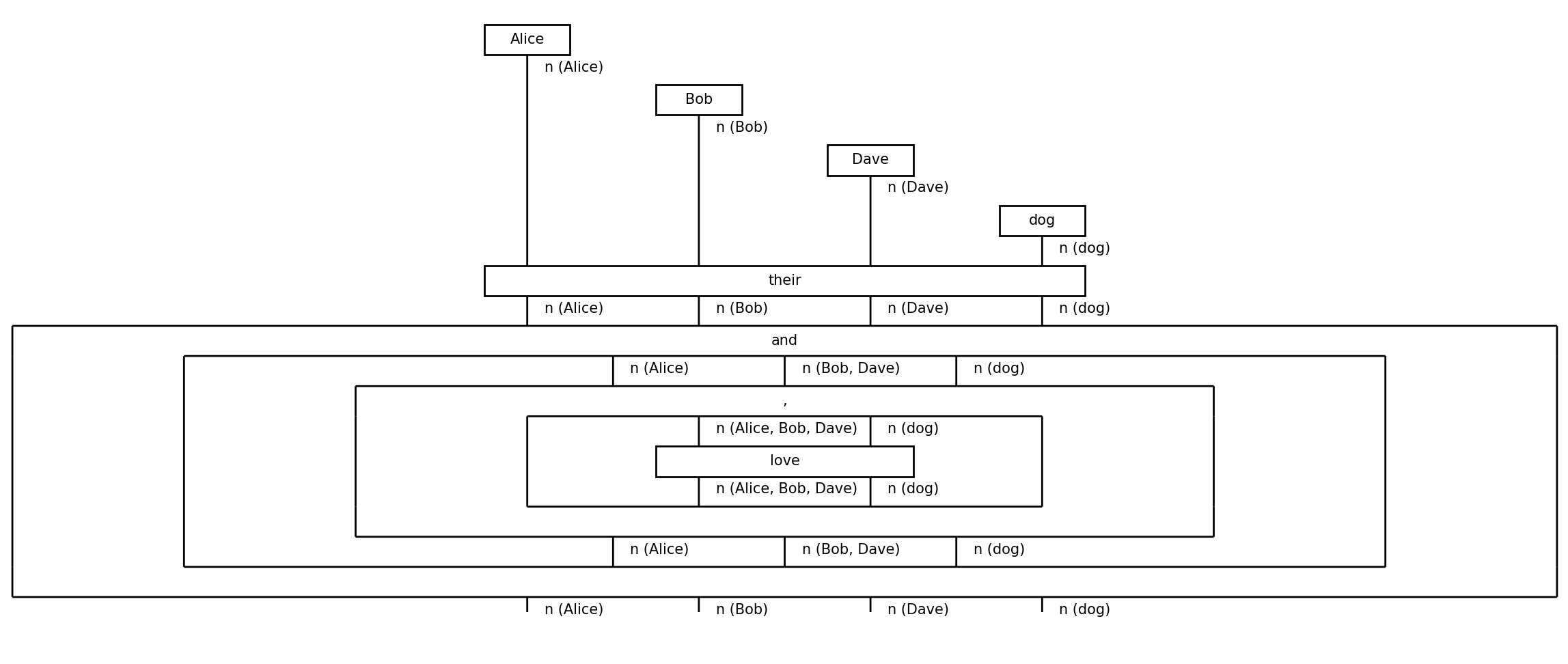}
\end{center}
Here the \texttt{their} box takes in three discourse referents --- \texttt{Alice}, \texttt{Bob} and \texttt{Dave} --- as well as the dog. 

Besides the implemented semantic rewrites, it is not difficult to imagine many more reasonable semantic rewrites -- for instance, making the conjunction \texttt{and} between two sentences an identity, or other semantic rewrites that may involve copying a component.

\section{Discussion and future work}
\label{sec:discussion}

We have detailed a software pipeline that converts natural language English text into its DisCoCirc representation.
This software enables for the first time the application of DisCoCirc to practical tasks in NLP.
One way text circuits can be used for NLP is to view them as a specification for dynamically generating a trainable model from smaller reusable components.
Specifically, the states in the diagram are viewed as trainable vector embeddings.
The wires propagating downwards out of the states should be thought of as carrying vectors.
The gates that act on the wires are parametrized processes that modify the vectors in the wires.
In this approach, a text circuit would itself be a large vector that encodes information about the referents in the text.
This vector can then be used for further processing as required by the task at hand.
A key point is that circuit components of the same shape and with same name would refer to the same trained component (i.e. they would share parameters).

The way in which we interpret circuit components as parametrized processes can be highly varied.
Given that DisCoCirc evolved from earlier models for quantum natural language processing, it should be unsurprising that DisCoCirc is particularly amenable to implementation on quantum computers.
Specifically, the representation of text as a circuit means it can be directly mapped onto a parametrized quantum circuit for the purposes of quantum machine learning.

Alternately, one can also interpret the diagrams as representing purely classical neural networks -- e.g. gates can be interpreted as generic feedforward networks.
This classical model can have a fundamentally different structure to the quantum version -- not only is there a difference in that the vectors are real-valued rather than complex-valued, but also the interpretation of two parallel wires can be entirely different (a tensor product in the quantum case, and a direct sum in the classical case).


Viewed this way as an approach to machine learning, DisCoCirc is notable for its interpretability.
An architecture based on text circuits would be more intrinsically interpretable (see \cite{molnar2022} for definition) as it exposes the flow of information.
Via this synthesis between structured representations and black-box machine learning approaches, this approach can help bridge the gulf between linguistic theory and the largely uninterpretable approaches of mainstream NLP.

At the level of DisCoCirc as a theory of semantics, the development of this pipeline points to many further questions. 
Many linguistic phenomena have not yet been modelled in DisCoCirc, and have not been considered in building this pipeline -- these include quantification, tense, and mood, to list a few.
Additionally, beyond the basic semantic relationships provided by CCG and coreference resolution, it would also be interesting to see if more general kinds of anaphora, such as ellipsis, can be incorporated into the model.
We may even think about incorporating even higher-level semantic relationships, like discourse relations.

Another major theoretical question for DisCoCirc is that of how to deal with discourse referents in general.
For one, the coreference resolution we have implemented in the pipeline cannot deal with more complicated cases of coreference, such as those involving compound nouns. 
Then there are even more complicated scenarios -- for instance, it is currently unclear in DisCoCirc how we can talk about \texttt{a group of people} and later on, say something about \texttt{one of the people}.
Many of the questions noted here can likely be answered by relating DisCoCirc with Discourse Representation Theory (DRT), which is a very similar formalism that has already addressed problems relating to tense, quantification, and more general referents.

The analogy between DRT and text circuits is significant, and would be interesting to explore.
Roughly speaking, the wires of text circuits correspond to the discourse referents of DRT, and the gates which act upon the wires
correspond to predicates and other conditions upon the referents that appear in DRT. 
There is also an analogy between the frames of text circuits (which are components that contain sub-circuits) and the nesting of discourse representation structures in DRT.

It is also interesting to explore the relationship between DisCoCirc and the previous framework of DisCoCat.
To this end, the work of~\cite{coecke2021grammar} suggests how the circuits of DisCoCirc could arise by looking at the \emph{internal wirings} of DisCoCat diagrams.
A concurrent work~\cite{ToumiDeFelice23} discusses how higher-order processes represented by frames could appear in DisCoCat by considering diagram-valued logical formulae.

Finally, another clear avenue to explore is to relate text circuits to dependency grammars~\cite{nivre2005dependency,de2021universal}, and investigate whether a text circuit parsing pipeline could be built starting from dependency grammar rather than CCG.
Dependency parsers have better coverage than CCG parsers in general -- for instance, being available for more textual domains -- and are available for more languages.

\section*{Acknowledgements}
We would like to thank Steve Clark for helpful comments and guidance involving CCG and co-indexing.
We would like to thank Vincent Wang-Maścianica for discussions and typesetting help, and Thomas Hoffman and Toumas Laakkonen for helpful comments.
We thank the Quantinuum Oxford office for reporting bugs in the software.
RS acknowledges support of the Clarendon Fund Scholarship. BR and RY thank Simon Harrison for his generous support for the Wolfson Harrison UK Research Council Quantum Foundation Scholarship.


\small
\bibliography{bibliography}


\clearpage
\normalsize



\appendix

\section{$\beta$-expansion}
\label{ssec:beta_expand_appendix}


Algorithm~\ref{alg:_drag_out} will suffice to perform the dragging out procedure on a $\lambda$-term, provided the term consists only of applications (and lists).
It fails in the presence of abstractions.
The remedy for this is straightforward. 
Before we do the main dragging out procedure, we first perform a recursive pass on the $\lambda$-tree going from the leaves to the root, which we call \textit{$\beta$-expansion} (Algorithm~\ref{alg:beta_expand}).
As the name suggests, this does a kind of inverse to the usual $\beta$-reduction
$$(\lambda x.M)N
\:\triangleright_{\beta}\:
M[N/x].
$$
Here $M[N/x]$ denotes the substitution operation that replaces instances of the variable $x$ in the body term $M$, with the term $N$. 

In our $\beta$-expansion procedure, we firstly go inside the body of an abstraction $\lambda$-term, and replace all instances of $n$-type constants as well as free $n$-type variables with newly introduced variables.
Then we take these constants and free variables that we removed outside the scope of the original abstraction.
That is, suppose we see a term
$$
\lambda x.M
$$
where $M$ is a subterm containing a single $n$-type free variable or constant $f$, and no more instances of abstraction.
Then the $\beta$-expansion step replaces the whole term with the $\beta$-equivalent term
$$
(\lambda y.\lambda x . M[y/f])f.
$$
Observe that $f$ has now been extracted from the scope of the original abstraction.

We illustrate the idea more concretely with a minimal example.
Consider the sentence
\begin{center}
	\texttt{Alice whom Bob likes walks}.
\end{center}
The CCG parse of this sentence involves an instance of the type-raising rule and (backward) composition
\begin{center}
	\small
	\cgex{5}{Alice & whom & Bob & likes & walkes \\
		\cgul & \cgul & \cgul & \cgul & \cgul \\
		\cat{n} & \cat{(n\bs n)\fs (s\bs n)} & \cat{n} & (\cat{s}\bs \cat{n})\fs \cat{n} & \cat{s \bs n}\\
		&& \cgline{1}{\cgfa T} \\[.4em]
		&& \cat{s\fs (s\bs n)} \\
		&& \cgline{2}{\cgfa B} \\
		&& \cgres{2}{s\fs n} \\
		& \cgline{3}{\cgfa} \\[.4em]
		& \cgres{3}{\cat{n\bs n}}\\
		\cgline{4}{\cgba}\\[.4em]
		\cgres{4}{\cat{n}}\\
		\cgline{5}{\cgba}\\[.4em]
		\cgres{5}{\cat{s}}\\
	}
\end{center}
which induces an irreducible abstraction in the $\lambda$-tree
\[\begin{tikzpicture}
	\begin{pgfonlayer}{nodelayer}
		\node [style=none] (2) at (2, 1.25) {$@$};
		\node [style=none] (3) at (5, 1.25) {\texttt{Bob}};
		\node [style=none] (4) at (0.875, 1.875) {};
		\node [style=none] (5) at (1.625, 1.375) {};
		\node [style=none] (6) at (2.375, 1.375) {};
		\node [style=none] (7) at (3.125, 1.875) {};
		\node [style=none] (8) at (5, 0.875) {$\scriptstyle n$};
		\node [style=none] (9) at (2, 0.875) {$\scriptstyle n \to s$};
		\node [style=none] (10) at (3.5, 0.125) {$@$};
		\node [style=none] (11) at (0.5, 2.375) {\texttt{likes}};
		\node [style=none] (12) at (2.375, 0.75) {};
		\node [style=none] (13) at (3.125, 0.25) {};
		\node [style=none] (14) at (3.875, 0.25) {};
		\node [style=none] (15) at (4.625, 0.75) {};
		\node [style=none] (16) at (0.5, 2) {$\scriptstyle n\to n\to s$};
		\node [style=none] (17) at (3.5, -0.25) {$\scriptstyle s$};
		\node [style=none] (18) at (3.5, 2.375) {$x$};
		\node [style=none] (19) at (3.5, 2) {$\scriptstyle n$};
		\node [style=none] (20) at (3.5, -1.375) {$\lambda x$};
		\node [style=none] (21) at (3.5, -0.5) {};
		\node [style=none] (22) at (3.5, -1.125) {};
		\node [style=none] (25) at (3.5, -1.75) {$\scriptstyle n\to s$};
		\node [style=none] (26) at (0.5, -1.5) {\texttt{whom}};
		\node [style=none] (27) at (0.5, -1.875) {$\scriptstyle (n\to s)\to n\to n$};
		\node [style=none] (28) at (2, -2.625) {$@$};
		\node [style=none] (29) at (3.125, -2) {};
		\node [style=none] (30) at (2.375, -2.5) {};
		\node [style=none] (31) at (1.625, -2.5) {};
		\node [style=none] (32) at (0.875, -2) {};
		\node [style=none] (33) at (2, -3) {$\scriptstyle n\to n$};
		\node [style=none] (34) at (5, -2.75) {\texttt{Alice}};
		\node [style=none] (35) at (5, -3.125) {$\scriptstyle n$};
		\node [style=none] (36) at (3.5, -3.875) {$@$};
		\node [style=none] (37) at (2.375, -3.25) {};
		\node [style=none] (38) at (3.125, -3.75) {};
		\node [style=none] (39) at (3.875, -3.75) {};
		\node [style=none] (40) at (4.625, -3.25) {};
		\node [style=none] (41) at (3.5, -4.25) {$\scriptstyle n$};
		\node [style=none] (42) at (0.5, -4) {\texttt{walks}};
		\node [style=none] (43) at (0.5, -4.375) {$\scriptstyle n\to s$};
		\node [style=none] (44) at (2, -5.125) {$@$};
		\node [style=none] (45) at (3.125, -4.5) {};
		\node [style=none] (46) at (2.375, -5) {};
		\node [style=none] (47) at (1.625, -5) {};
		\node [style=none] (48) at (0.875, -4.5) {};
		\node [style=none] (49) at (2, -5.5) {$\scriptstyle s$};
	\end{pgfonlayer}
	\begin{pgfonlayer}{edgelayer}
		\draw (4.center) to (5.center);
		\draw (7.center) to (6.center);
		\draw (12.center) to (13.center);
		\draw (15.center) to (14.center);
		\draw (21.center) to (22.center);
		\draw (29.center) to (30.center);
		\draw (32.center) to (31.center);
		\draw (37.center) to (38.center);
		\draw (40.center) to (39.center);
		\draw (45.center) to (46.center);
		\draw (48.center) to (47.center);
	\end{pgfonlayer}
\end{tikzpicture}\]
To illustrate the affect of abstractions, for this example we can diagrammatically depict an abstraction as a frame which bounds a certain subterm.
The drawing for this $\lambda$-term is then:
\begin{center}
	\includegraphics[width=0.7\textwidth]{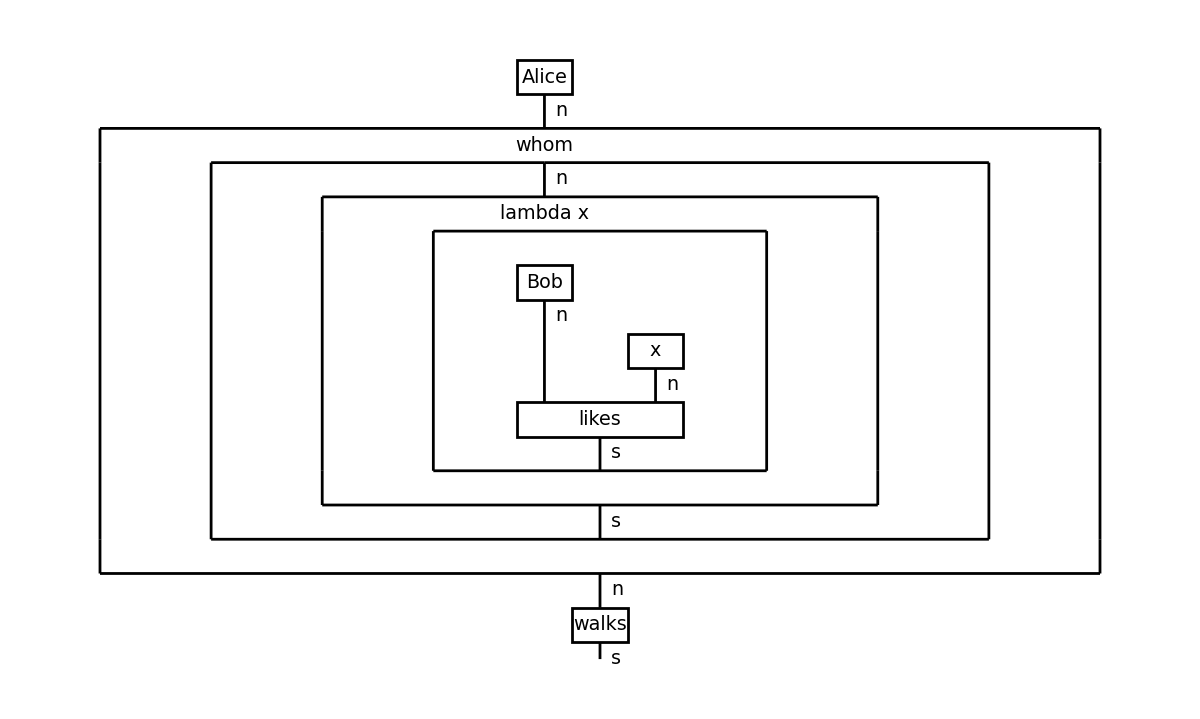}
\end{center}
Evidently this simplifies to
\begin{center}
	\includegraphics[width=0.7\textwidth]{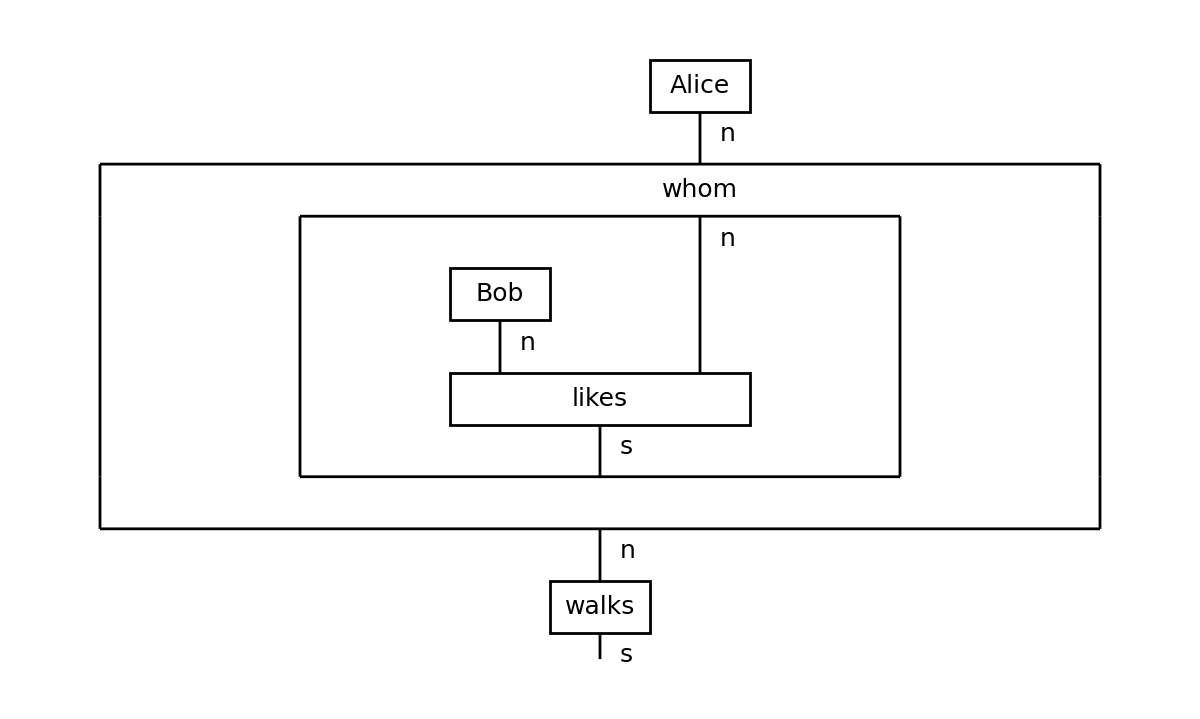}
\end{center}
This diagram makes it clear that \texttt{Bob} needs to be dragged out of the \texttt{who} box.
However, the dragging out procedure given in the previous section is insufficient to extract states that are inside the body of a abstraction.

The solution is to rewrite the $\lambda$-tree to the $\beta$-equivalent version below (where we have replaced \texttt{Bob} by a bound variable \texttt{y}):
\[\begin{tikzpicture}
	\begin{pgfonlayer}{nodelayer}
		\node [style=none] (0) at (1.875, 4.5) {$@$};
		\node [style=none] (1) at (4.875, 4.5) {$y$};
		\node [style=none] (2) at (0.75, 5.125) {};
		\node [style=none] (3) at (1.5, 4.625) {};
		\node [style=none] (4) at (2.25, 4.625) {};
		\node [style=none] (5) at (3, 5.125) {};
		\node [style=none] (6) at (4.875, 4.125) {$\scriptstyle n$};
		\node [style=none] (7) at (1.875, 4.125) {$\scriptstyle n \to s$};
		\node [style=none] (8) at (3.375, 3.375) {$@$};
		\node [style=none] (9) at (0.375, 5.625) {\texttt{likes}};
		\node [style=none] (10) at (2.25, 4) {};
		\node [style=none] (11) at (3, 3.5) {};
		\node [style=none] (12) at (3.75, 3.5) {};
		\node [style=none] (13) at (4.5, 4) {};
		\node [style=none] (14) at (0.375, 5.25) {$\scriptstyle n\to n\to s$};
		\node [style=none] (15) at (3.375, 3) {$\scriptstyle s$};
		\node [style=none] (16) at (3.375, 5.625) {$x$};
		\node [style=none] (17) at (3.375, 5.25) {$\scriptstyle n$};
		\node [style=none] (18) at (3.375, 2) {$\lambda x$};
		\node [style=none] (19) at (3.375, 2.75) {};
		\node [style=none] (20) at (3.375, 2.25) {};
		\node [style=none] (21) at (3.375, 1.625) {$\scriptstyle n\to s$};
		\node [style=none] (22) at (6.25, 0.5) {\texttt{Bob}};
		\node [style=none] (23) at (6.25, 0.125) {$\scriptstyle n$};
		\node [style=none] (24) at (4.75, -0.625) {$@$};
		\node [style=none] (25) at (3.625, 0) {};
		\node [style=none] (26) at (4.375, -0.5) {};
		\node [style=none] (27) at (5.125, -0.5) {};
		\node [style=none] (28) at (5.875, 0) {};
		\node [style=none] (29) at (4.75, -1) {$\scriptstyle n\to s$};
		\node [style=none] (30) at (6.25, -1.75) {\texttt{Alice}};
		\node [style=none] (31) at (6.25, -2.125) {$\scriptstyle n$};
		\node [style=none] (32) at (4.75, -2.875) {$@$};
		\node [style=none] (33) at (3.625, -2.25) {};
		\node [style=none] (34) at (4.375, -2.75) {};
		\node [style=none] (35) at (5.125, -2.75) {};
		\node [style=none] (36) at (5.875, -2.25) {};
		\node [style=none] (37) at (4.75, -3.25) {$\scriptstyle n$};
		\node [style=none] (38) at (1.75, -3) {\texttt{walks}};
		\node [style=none] (39) at (1.75, -3.375) {$\scriptstyle n\to s$};
		\node [style=none] (40) at (3.25, -4.125) {$@$};
		\node [style=none] (41) at (4.375, -3.5) {};
		\node [style=none] (42) at (3.625, -4) {};
		\node [style=none] (43) at (2.875, -4) {};
		\node [style=none] (44) at (2.125, -3.5) {};
		\node [style=none] (45) at (3.25, -4.5) {$\scriptstyle s$};
		\node [style=none] (50) at (3.375, 0.625) {$\lambda y$};
		\node [style=none] (51) at (3.375, 1.375) {};
		\node [style=none] (52) at (3.375, 0.875) {};
		\node [style=none] (53) at (3.375, 0.25) {$\scriptstyle n\to n\to s$};
		\node [style=none] (56) at (3.25, -1.75) {$@$};
		\node [style=none] (57) at (2.125, -1.125) {};
		\node [style=none] (58) at (2.875, -1.625) {};
		\node [style=none] (59) at (3.625, -1.625) {};
		\node [style=none] (60) at (4.375, -1.125) {};
		\node [style=none] (61) at (1.75, -0.5) {\texttt{whom`}};
		\node [style=none] (62) at (1.75, -0.875) {$\scriptstyle (n\to s)\to n\to n$};
		\node [style=none] (63) at (3.25, -2.125) {$\scriptstyle n\to n$};
	\end{pgfonlayer}
	\begin{pgfonlayer}{edgelayer}
		\draw (2.center) to (3.center);
		\draw (5.center) to (4.center);
		\draw (10.center) to (11.center);
		\draw (13.center) to (12.center);
		\draw (19.center) to (20.center);
		\draw (25.center) to (26.center);
		\draw (28.center) to (27.center);
		\draw (33.center) to (34.center);
		\draw (36.center) to (35.center);
		\draw (41.center) to (42.center);
		\draw (44.center) to (43.center);
		\draw (51.center) to (52.center);
		\draw (57.center) to (58.center);
		\draw (60.center) to (59.center);
	\end{pgfonlayer}
\end{tikzpicture}\]
In circuit notation, we get
\begin{center}
	\raisebox{-.4\height}{
	\includegraphics[width=0.45\textwidth]{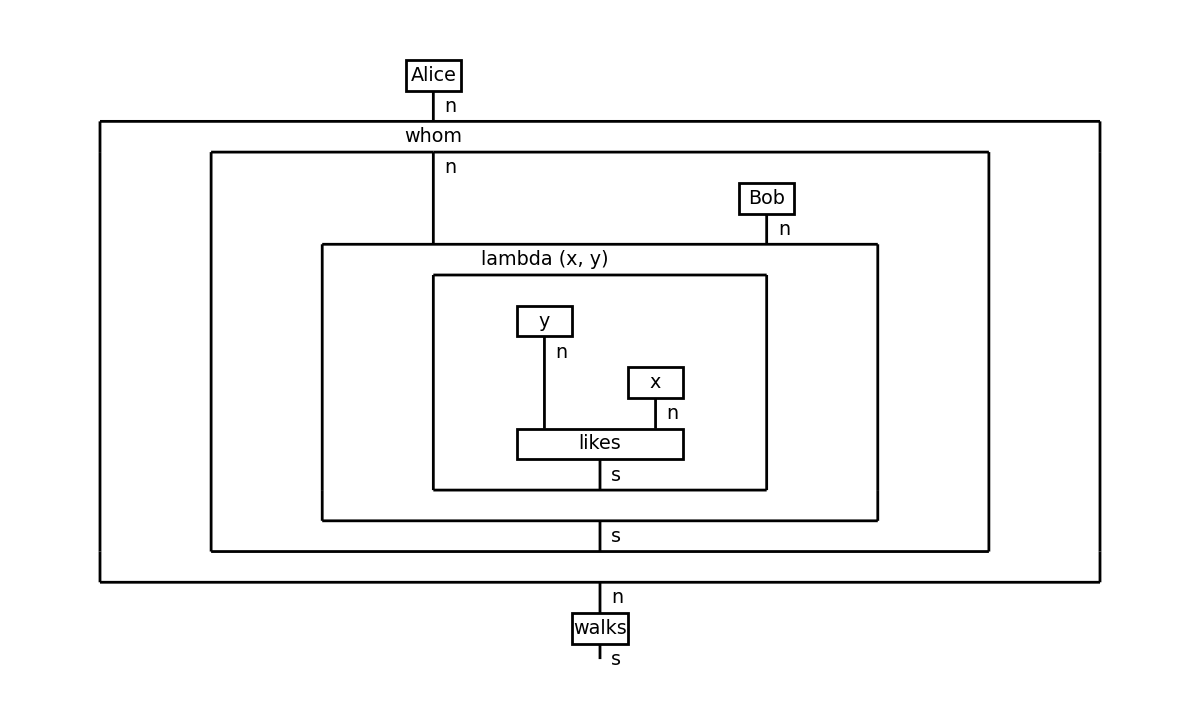}}
	$\rightsquigarrow$
	\raisebox{-.4\height}{
	\includegraphics[width=0.45\textwidth]{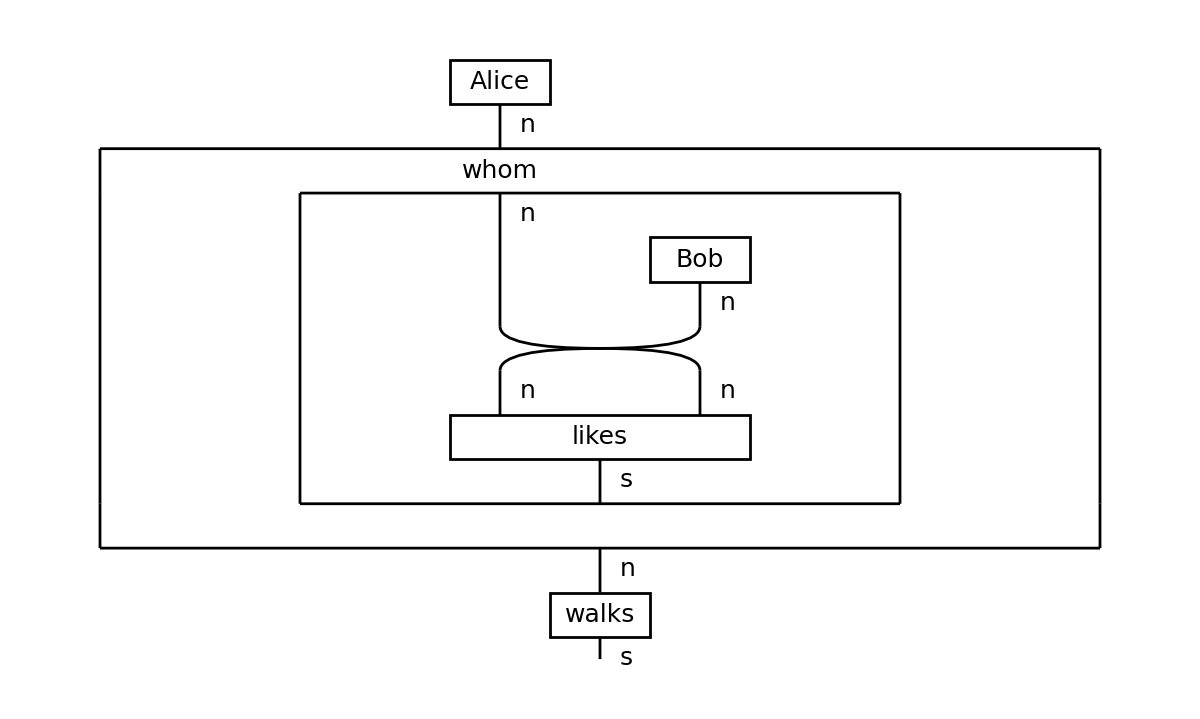}}
\end{center}
In the $\lambda$-tree, \texttt{Bob} has been freed from the scope of the abstraction.
This is best seen in the left-hand diagram, where \texttt{Bob} is outside of the abstraction frame.
Now we can apply our usual dragging out routine to obtain the properly dragged out tree and diagram:
\begin{center}
	\raisebox{-.4\height}{
	\includegraphics[width=0.45\textwidth]{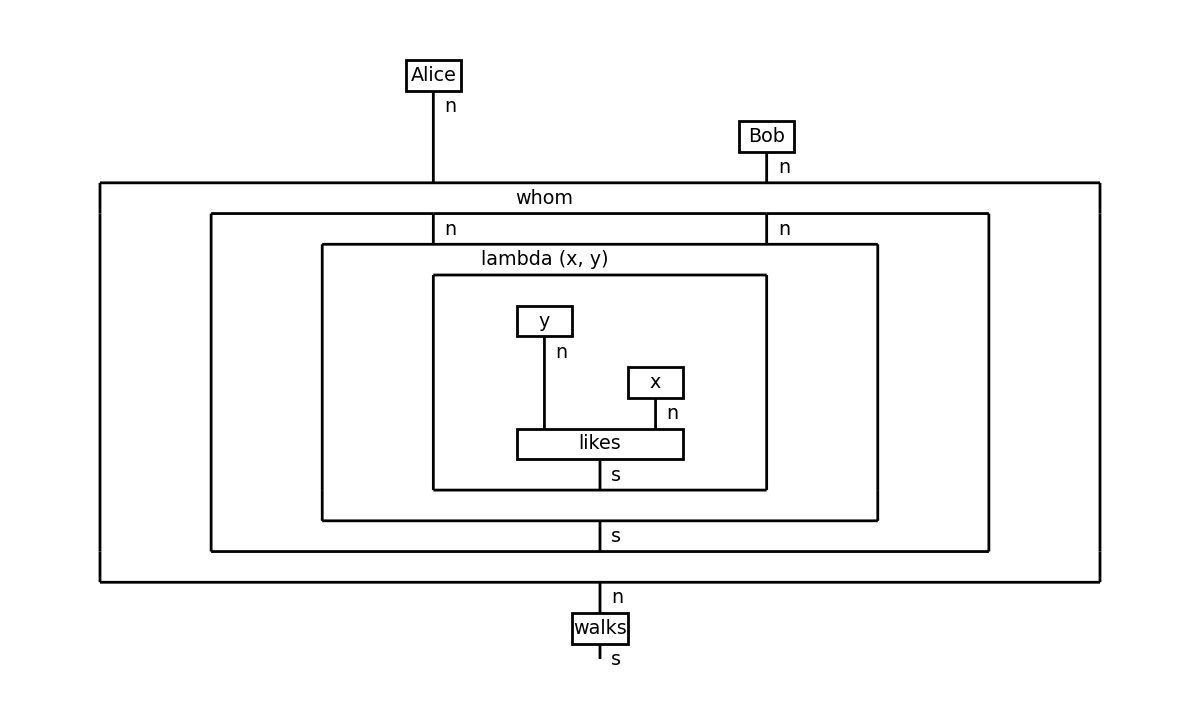}}
	$\rightsquigarrow$
	\raisebox{-.4\height}{
	\includegraphics[width=0.45\textwidth]{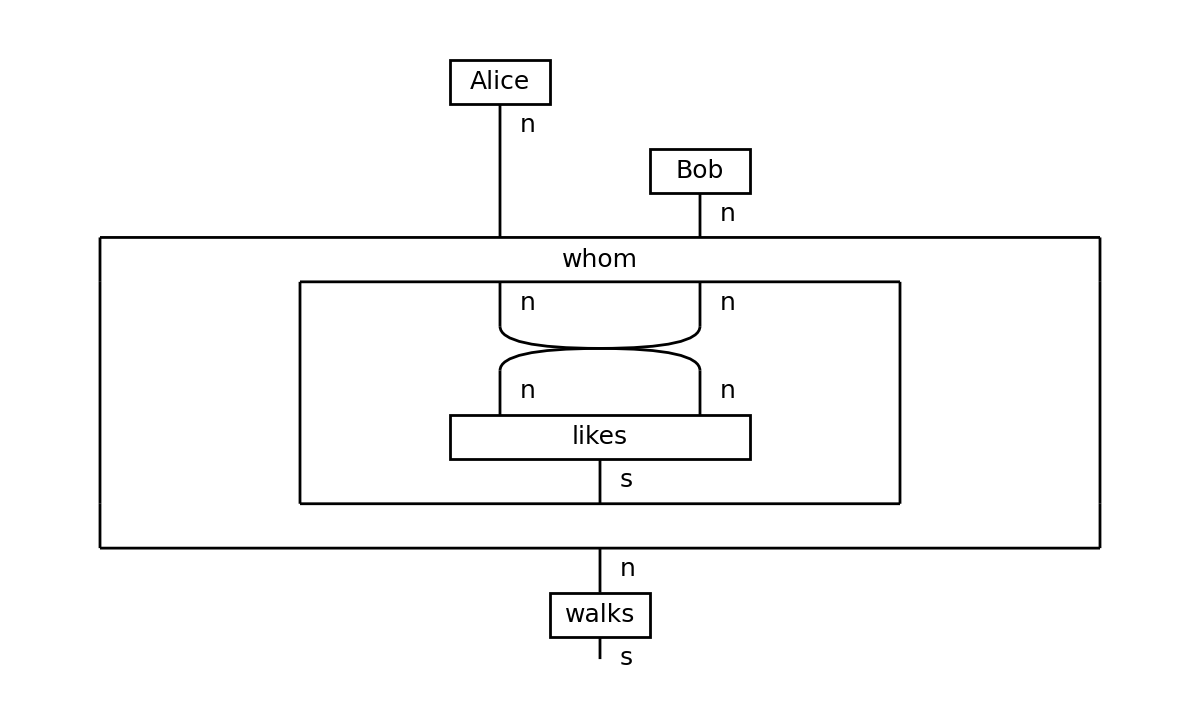}}
\end{center}
Indeed, in the general case, after performing the $\beta$-expansion pass that recursively extracts all free nouns from abstraction scopes, we are able to just drag out independently within each abstraction scope.
The complete dragging out routine is then described by:
\begin{center}
	DragOut(BetaExpand(term))
\end{center}

\begin{algorithm}[H]
	\caption{BetaExpand}\label{alg:beta_expand}
	\begin{algorithmic}
		\STATE Input: a $\lambda$-term
		\STATE Output: a $\beta$-expanded $\lambda$-term
		\IF{term is a constant or variable}
			\STATE pass
		\ELSIF{term is an application}
			\STATE term $\leftarrow$ BetaExpand(term.function)(BetaExpand(term.argument))
		\ELSIF{term is a list}
			\STATE term $\leftarrow$ [BetaExpand(t) for t in term]
		\ELSIF{term is an abstraction}
			\STATE term $\leftarrow$ $\lambda$ (term.variable).(BetaExpand(term.body))
			\STATE vars $\leftarrow$ find all constants and free variables in term
			\STATE dummyVars $\leftarrow$ initialize a list of dummy variables for each variable and constant in vars
			\FOR{i in length(vars)}
				\STATE term $\leftarrow$ term[dummyVars[i]/vars[i]]
			\ENDFOR
			\FOR{var in dummyVars}
				\STATE term $\leftarrow$ $\lambda$var.term
			\ENDFOR
			\FOR{var in reversed(vars)}
				\STATE \# reattach the freed vars to the term
				\STATE term $\leftarrow$ term(var)
			\ENDFOR
		\ENDIF
		\RETURN term
	\end{algorithmic}
\end{algorithm}


%
%
%

%
%

\end{document}